%% file: main.tex
\documentclass[10pt]{article} 
\usepackage[accepted]{tmlr}

\input{math_commands.tex}

\usepackage{hyperref}
\usepackage{url}
\usepackage{subcaption}
\usepackage{graphicx}
\usepackage{colortbl}
\usepackage{adjustbox}
\usepackage{booktabs, multirow} 
\usepackage{soul}
\usepackage{changepage,threeparttable} 
\usepackage{xspace}
\usepackage{amsmath}
\usepackage{amssymb}
\usepackage{booktabs}       
\usepackage{amsfonts}       

\title{From Spikes to Heavy Tails: Unveiling the Spectral \\Evolution of Neural Networks}

\author{\name Vignesh Kothapalli \email vk2115@nyu.edu \\
      \addr Courant Institute of Mathematical Sciences\\
      New York University
      \AND
      \name Tianyu Pang \email tianyupang628@gmail.com \\
      \addr Department of Computer Science \\
      Dartmouth College
      \AND
      \name Shenyang Deng \email  shenyang.deng.gr@dartmouth.edu\\
      \addr Department of Computer Science \\
      Dartmouth College
      \AND
      \name Zongmin Liu \email zongminl@stanford.edu \\
      \addr Department of Neurology and Neurological Sciences
      \\Stanford University
      \AND
      \name Yaoqing Yang \email yaoqing.yang@dartmouth.edu\\
      \addr Department of Computer Science \\
      Dartmouth College
      }

\def\month{08}  
\def\year{2025} 
\def\openreview{\url{https://openreview.net/forum?id=DJHB8eBUnt}} 

\begin{document}

\maketitle

\begin{abstract}
Training strategies for modern deep neural networks (NNs) tend to induce a heavy-tailed (HT) empirical spectral density (ESD) in the layer weights. While previous efforts have shown that the HT phenomenon correlates with good generalization in large NNs, a theoretical explanation of its occurrence is still lacking. Especially, understanding the conditions which lead to this phenomenon can shed light on the interplay between generalization and weight spectra. Our work aims to bridge this gap by presenting a simple, rich setting to model the emergence of HT ESD. In particular, we present a theory-informed setup for `crafting' heavy tails in the ESD of two-layer NNs and present a systematic analysis of the HT ESD emergence without any gradient noise. This is the first work to analyze a noise-free setting, and we also incorporate optimizer (\texttt{GD/Adam}) dependent (large) learning rates into the HT ESD analysis. Our results highlight the role of learning rates on the Bulk+Spike and HT shape of the ESDs in the early phase of training, which can facilitate generalization in the two-layer NN. These observations shed light on the behavior of large-scale NNs, albeit in a much simpler setting. The code and documentation is available at: \href{https://github.com/kvignesh1420/single-index-ht}{\texttt{https://github.com/kvignesh1420/single-index-ht}}
\end{abstract}

\section{Introduction}
\label{sec:introduction}

Training NNs on real-world data, which often has low intrinsic dimensionality~\citep{Wright-Ma-2022, hastie2005elements}, can commonly lead to spiked covariance structures in the ESD of weight matrices~\citep{ba2023learning, lee2024neural}. From a classical statistics viewpoint, spiked models capture low-rank signal directions emerging from noisy high-dimensional data~\citep{anderson2010introduction, arous2008spectrum, ding2021spiked}, offering a natural framework to study spectral evolution during training. Recent work by~\citet{martin2021implicit} showed that the ESD of weight matrices evolves through a sequence of phases to exhibit a heavy-tailed distribution—even without explicit regularization. While the Bulk+Spike shaped covariance matrices have been widely analyzed~\citep{baik2005phase, anderson2010introduction,ba2023learning}, the emergence of heavy-tailed ESDs has drawn increasing attention due to their strong correlation with generalization performance in modern deep learning tasks across vision and language domains. The shape of HT ESDs has also been successfully leveraged to assess the quality of pre-trained NNs \citep{martin2020heavy, martin2021predicting, martin2021post} (including Large Language Models \citep{yang2023test}), and design layer-wise learning rate schedulers \citep{zhou2023temperature, liu2024model}.


\paragraph{Towards a practical systematic analysis of HT ESDs.} Owing to the wide range of applications described above, many efforts \citep{simsekli2019tail, simsekli2020hausdorff, simsekli2020fractional, gurbuzbalaban2021heavy, hodgkinson2021multiplicative, hodgkinson2022generalization, dandi2024random} have tried to understand the underlying mechanisms of the HT phenomenon. In particular, they have focused on the limiting distributions of the weight values by modeling the stochastic gradient noise as stochastic processes. The underlying idea is that when the weight values tend to an HT distribution, then the resulting ESD also exhibits a similar transition towards heavy-tailedness (see \cite{arous2008spectrum}). Although fundamentally connected to the HT ESDs, these insights apply to the limiting/convergence phases of training and do not directly explain the HT ESD emergence after finite training steps as observed by the empirical studies of \cite{martin2021implicit,martin2021predicting}. Furthermore, it is currently unclear if such emergence is limited to large-scale NNs trained until convergence and if it is even possible to simulate such behavior in small-scale NNs during the early phases of training.

\begin{figure}[t!]
         \centering
         \begin{subfigure}[b]{0.24\textwidth}
         \centering
         \includegraphics[width=\textwidth]{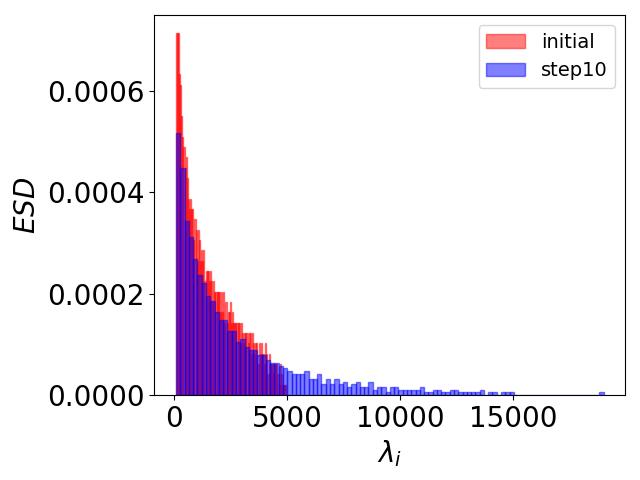}
          \caption{\texttt{GD} $\eta=2000$}
     \end{subfigure}
     \hfill
     \begin{subfigure}[b]{0.24\textwidth}
         \centering
         \includegraphics[width=\textwidth]{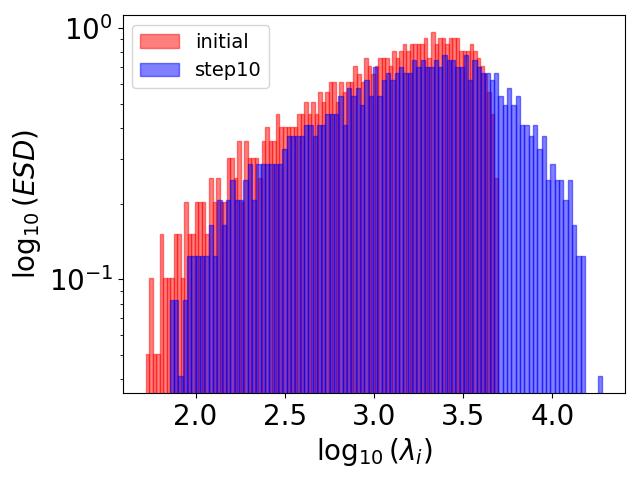}
         \caption{\texttt{GD} $\eta=2000$}
     \end{subfigure}
     \hfill
     \begin{subfigure}[b]{0.24\textwidth}
         \centering
         \includegraphics[width=\textwidth]{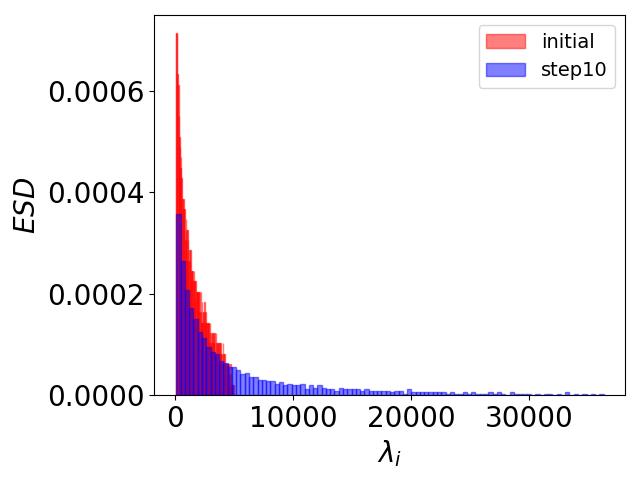}
         \caption{\texttt{FB-Adam} $\eta=0.5$}
     \end{subfigure}
     \hfill
     \begin{subfigure}[b]{0.24\textwidth}
         \centering
         \includegraphics[width=\textwidth]{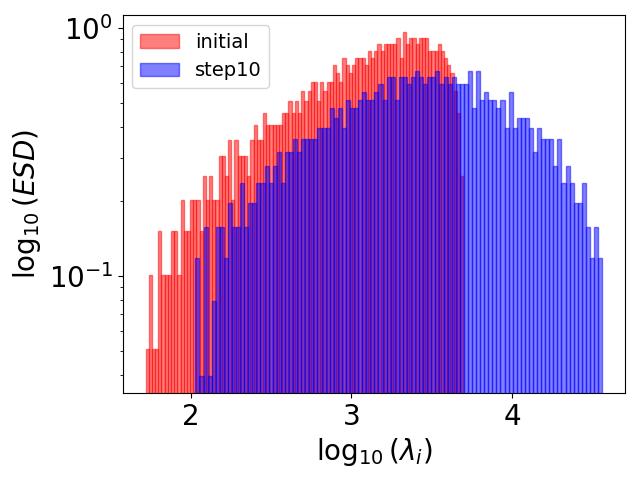}
         \caption{\texttt{FB-Adam} $\eta=0.5$}
     \end{subfigure}
        \caption{Emergence of HT spectra after $10$ \texttt{GD}/\texttt{FB-Adam} steps. Both linear and log scales are shown. For the purpose of illustrating clear HT shapes, we did not choose a highly HT spectrum. Highly HT spectra can also be generated without gradient noise and are discussed in Appendix~\ref{app:heaviest}.}
        \label{fig:intro_teaser}
        \vspace{-4mm}
\end{figure}

\paragraph{A Tangible Setup.} In this paper, we present an alternative setup that is amenable to discrete-step analysis and is rich enough to study the evolution of ESDs in two-layer NNs without any gradient noise. Our setup does not rely on continuous time approximations to SGD or limiting distribution analysis but instead employs a \emph{Teacher-Student} setting \citep{arous2021online, bietti2022learning, ba2022high, dandi2023two, ba2023learning, mousavi2023gradient} to study the effects of finite optimizer steps on the weight matrix ESD (Figure \ref{fig:intro_teaser}). By training a two-layer feed-forward NN (\emph{Student}) to learn a single-index model (\emph{Teacher}) using vanilla Gradient Descent (\texttt{GD}) and Full-batch Adam (\texttt{FB-Adam}) optimizers, we show that: 

\emph{``The ESD of the hidden layer weight matrix exhibits heavy tails after multiple steps of \texttt{GD}/\texttt{FB-Adam} with (sufficiently) large learning rates''.}

\paragraph{Intuition.} Before diving into the details, we present the reader with a mental model of the ESD evolution in this setup. At initialization, the hidden layer weight matrix ESD of the student NN can be characterized by the Marchencko-Pastur distribution (i.e. random-like). The first step with a \emph{large} $\eta$ has been shown to result in an outlier singular value (i.e. a `spike') in the ESD and results in a `Bulk+Spike' shape. The singular vector corresponding to this spike tends to align with the target direction of the teacher model and results in improved generalization \citep{ba2022high, dandi2023two}. By continuing to train the student NN with such \emph{large} $\eta$, our work shows that the interactions between the spike and the bulk gradually lead to a `Bulk-Decay' and finally lead to HT ESD.




Our theory is primarily aimed at identifying the \emph{large} $\eta$ for \texttt{FB-Adam}, which can result in a spike after the one-step update. Especially, we formulate the scale of $\eta$ (depending on hidden layer width and data dimension) which can lead to a Bulk+Spike ESD after the one-step update. Such a result has been formulated and studied for one-step \texttt{GD} by recent works \citep{ba2022high, dandi2023two} but none of the previous works have extended the analysis to $\texttt{FB-Adam}$ (which is closer to the practical scenarios where \texttt{Adam} is widely used). 

Toward understanding the shape transitions of ESD from the Bulk+Spike to an HT, we vary $\eta$ from small to large values and observe that the distance between the spike and the bulk (i.e spectral gap) determines the HT emergence during the early phases of training. For instance, $\eta=1$ leads to HT ESDs just after $t=10$ steps, whereas $\eta=0.1$ requires very long training up to $t=10000$ steps (see also Table~\ref{tab:GD_transition}, Table~\ref{tab:ADAM_transition}). Thus highlighting the necessity to study critical learning rates for one-step optimizer updates. We also analyze the correlations between the heaviness of tails (as measured by power-law fits) in the ESD and the generalization of the student NN after multiple steps. We showcase empirical results on the existence of a range of $\eta$ (depending on the optimizer) that can result in an HT ESD as well as improve the generalization of the student NN. In particular, $\eta$ in such a suitable range results in HT ESDs whose power-law fit (as measured by the hill-climbing approach \citep{yang2023test}) lie in the range of $(2, 2.5)$ and lead to strong generalization. This result supports the observations of \citet{martin2021implicit, martin2021predicting} for well-trained deep NNs where the truncated power-law fit of the ESDs lies within a range of $(2,4)$. 

To summarize, our main contributions are as follows:
\begin{itemize}
    \item We present a gradient-noise-free setting to study the emergence of HT ESD in the hidden layer weight matrix of two-layer NNs during the initial phases of training. To the best of our knowledge, this is the first work to study the early evolution phases of the ESD across discrete training steps with large learning rates. In this setup, we also present deeper connections between feature learning after the first step (via the spike in the weight matrix ESD) and the HT emergence after finite steps.
    \item We theoretically establish the scale of the learning rate at which one step of \texttt{FB-Adam} results in a spike in the hidden layer weight matrix ESD. This result can be of independent interest to the community since prior results have been established only for \texttt{GD}.
    \item We empirically analyze the evolution of the hidden layer weight matrix ESD into an HT distribution during training. Interestingly, we show that for a certain range of optimizer-dependent learning rates, the two-layer NN can exhibit HT ESDs and generalize well. 
\end{itemize}

Overall, our paper makes multiple novel theoretical and empirical contributions centered around the emergence of HT ESDs and their correlations with the good generalization of NNs.






\section{Related Work}
\label{sec:related_work}

\paragraph{The heavy-tailed phenomenon.} Heavy tails in machine learning have been observed and studied in various forms. The most prominent empirical results are from a series of works by \citet{martin2020heavy, martin2021implicit, martin2021post, martin2021predicting}, which propose a heavy-tailed self regularization (HT-SR) theory of deep NNs. In particular, \citet{martin2021implicit} proposed a `$5+1$' phase model corresponding to the ESD evolution, and aims to model the HT-SR effect during training. The phases are as follows: (1) Random-like, (2) Bleeding-out, (3) Bulk+Spikes, (4) Bulk-Decay, (5) Heavy-Tailed (HT), and the extreme case of (6) Rank-Collapse. Their observations from extensive empirical analysis showcased a correlation between the power law fits of weight matrix ESDs and generalization. Although it is still unclear if HT ESDs are necessary for generalization, \citet{martin2021predicting, yang2023test} have shown that the shape metrics of the ESD can be effectively leveraged to identify well-trained NNs (and can be extended to large-scale models such as LLMs). From a theoretical perspective, previous efforts have primarily focused on stochastic optimization settings and proposed generalization bounds based on the tail indices of the stochastic noise \citep{simsekli2019tail, simsekli2020fractional, simsekli2020hausdorff, gurbuzbalaban2021heavy, hodgkinson2021multiplicative, hodgkinson2022generalization, raj2023algorithmic, nguyen2019first, barsbey2021heavy}. In particular, the earlier work by \citet{simsekli2020hausdorff} employed continuous-time approximation of \texttt{SGD} via Feller-processes and studied the role of the Hausdorff Dimension on generalization. More recently, \citet{hodgkinson2022generalization} extended this analysis to discrete-time settings. In concurrent works, \citet{gurbuzbalaban2021heavy} studied the heavy-tailed stationary distributions of discrete Markov processes as an approximation to SGD iterates in the infinite data regime, while \citet{hodgkinson2021multiplicative} analyzed the role of multiplicative noise in such settings. While these efforts have emphasized the role of the learning rate/batch size ratio in determining the tail index of the iterates, a formal study on the role of these hyper-parameters is still lacking. More importantly, none of these efforts have focused on the evolution of the ESD itself. Recent work by \cite{wang2023spectral} empirically studied the emergence of HT-ESDs when training shallow neural networks to learn multi-index models with Adam. They showcased that Adam can facilitate the emergence of HT-ESDs, but do not theoretically characterize the learning rates required to observe such behavior. Furthermore, they do not present a systematic study of the correlation between learning rates and tail indices for HT ESD. 
Our work aims to bridge these gaps and analyzes the fine-grained ESD evolution from a Marchenko-Pastur (MP) fit (i.e. random initialization) $\to$ ``Bulk+Spike'' $\to$ ``Bulk-Decay'' $\to$ HT distributions and the correlations with generalization.

\paragraph{Feature learning and large learning rates.} Learning \emph{single-index} models using two-layer NNs under the \emph{Teacher-Student} setup has provided rich insights into the sample complexity \citep{damian2023smoothing, hosseini2023neural, zweig2023on, damian2024computational, damian2022neural, abbe2023sgd} and training dynamics \citep{bietti2022learning, wang2023spectral, cui2024asymptotics, moniri2023theory} of NNs. In particular, the study of \emph{feature learning} in such two-layer NNs focuses on the factors (such as optimizers \citep{abbe2023sgd}, loss landscapes \citep{damian2023smoothing}, representations \citep{nichani2023provable} and learning rates \citep{dandi2023two}) that facilitate sample efficient learning beyond the kernel regime \citep{louart2018random,gerace2020generalisation,mei2022generalization,hu2022universality,goldt2022gaussian, liu2021random}. Recently, \citet{ba2022high} analyzed the first step of \texttt{GD} update in the high dimensional setting and formalized the scale of $\eta$ required to go beyond the random feature regime. In particular, such a \emph{large} $\eta$ is necessary for a two-layer NN to learn the hidden direction of a single-index model after one step of \texttt{GD} (see also \citep{dandi2023two}). This first \emph{large} update was shown to result in an outlier singular value (i.e. a `spike') in the ESD of the hidden layer weight matrix. Our work presents the first result for such \emph{large} $\eta$ in the case of \texttt{FB-Adam} and can be of wider interest in the feature learning context. Such a characterization is important since \cite{dandirandom} have shown in the asymptotic limit of input dimension $d\to \infty$ (and the proportional scaling regime) that the spikes in the weight matrix ESD tend to affect the ESD shape of feature matrix after a single step with large $\eta$. Beyond the one-step analysis, \citet{dandi2024benefits} analyzed the role of two-pass \texttt{GD} in learning single-index models with large information exponents \citep{arous2021online}. However, the effects of multiple passes over the data on the ESD and generalization are yet to be fully understood. Our multi-step analysis empirically highlights the importance of such feature learning after the first step for the emergence of HT ESD. Thus, showcasing the underexplored connections between these two areas of research.

\section{Preliminaries and Setup}
\label{sec:setup}

\paragraph{Notation.} For $n \in \sN$, we denote $[n] = \{1, \cdots, n\}$. We use $O(\cdot)$ to denote the standard big-O notation and the subscript $O_d(\cdot)$ to denote the asymptotic limit of $d \to \infty$. Formally, for two sequences of real numbers $x_d$ and $y_d$, $x_d = O_d(y_d)$ represents $\lim_{d \to \infty} |x_d| \leq C_1 |y_d|$ for some constant $C_1$. Similarly, $x_d = O_{d, \mathbb{P}}(y_d)$ denotes that the asymptotic inequality almost surely holds under a probability measure $\mathbb{P}$. The definitions can be extended to the standard $\Omega(\cdot), \Theta(\cdot)$ or $\asymp$ notations analogously \citep{graham19890}. For two sequences of real numbers $x_d$ and $y_d$, $x_d\asymp y_d$ represents $ |y_d| C_2 \le |x_d| \le C_1 |y_d|$, for constants $C_1, C_2 > 0$ \citep{wang2021convergence, moniri2023theory}. For a real matrix $\mB=(B_{ij})_{n\times m} \in \sR^{n \times m}$, $\mB^{\circ p}$ represents an element-wise $p$-power transformation such that $\mB^{\circ p}=(B_{ij}^{p})_{n\times m}$. $\odot$ is the matrix Hadamard product, $\operatorname{sign}(.)$ denotes the element-wise sign function. $\norm{\cdot}_{2}$ denotes the $\ell_2$ norm for vectors and the operator norm for matrices. $\norm{\cdot}_{F}$ denotes the Frobenius norm. $\vzero_{h \times d}, \vone_{h \times d} \in \sR^{h \times d}$ represent the all-zero and all-ones matrices.

\paragraph{Dataset.} We sample $n$ data points $\{\vx_1, \cdots, \vx_n\}$ from the isotropic Gaussian $\vx_i \sim \gN(\vzero_d, \mI_d), \forall i \in [n]$ as our input data. For a given $\vx_i \in \sR^d$, we use a single-index \textit{teacher} model $F^*:\sR^d \to \sR$ to generate the corresponding scalar label $y_i \in \sR$ as follows:
\begin{equation}
    y_i = F^*(\vx_i) + \xi_i = \sigma_*( \vbeta^{*\top} \vx_i) + \xi_i.
\end{equation}
Here $\vbeta^* \in \sS^{d-1}$ (the $d-1$-dimensional sphere in $\sR^d$) is the \textit{target direction}, $\sigma_* : \sR \to \sR$ is the \textit{target non-linear link function}, and $\xi_i \sim \gN(0, \rho_e^2)$ is the independent additive label noise. We represent $\mX \in \sR^{n \times d}, \vy \in \sR^n$ as the input matrix and the label vector, respectively.

\paragraph{Learning.} We consider a two-layer fully-connected NN with activation $\sigma: \sR \to \sR$ as our \textit{student} model $f(\cdot) : \sR^d \to \sR$. For an input $\vx_i \in \sR^d$, its prediction is formulated as:
\begin{align}
    \label{update_formula}
    f(\vx_i) &= \frac{1}{\sqrt{h}} \va^\top \sigma\left( \frac{1}{\sqrt{d}}  \mW \vx_i \right).
\end{align}
Here $\mW \in \sR^{h \times d}, \va \in \sR^h$ are the first and second layer weights, respectively, with entries sampled i.i.d as follows $\left[\mW_0\right]_{i, j} \sim \mathcal{N}(0,1)$, $\left[a\right]_{i} \sim \mathcal{N}(0,1), \forall i \in [h], j \in [d]$.

\subsection{Training Procedure}

We employ the following \emph{Two-stage training} procedure \citep{ba2022high,moniri2023theory,cui2024asymptotics,dandi2023two,wang2023spectral} on the student network. In the first stage, we fix the last layer weights $\va \in \sR^h$ and apply optimizer update(s) (\texttt{GD}/\texttt{FB-Adam}) only for the first layer $\mW$. In the second stage, we perform ridge regression on the last layer using a hold-out dataset of the same size to calculate the ideal value of $\va$ \citep{ba2022high}.

\paragraph{Optimizer updates for the first layer.} In this phase, we fix the last layer weights $\va$ to its value at initialization and perform \texttt{GD}/\texttt{FB-Adam} update(s) on $\mW$ to minimize the mean-squared error $R(f, \mX, \vy) = \frac{1}{2n} \sum_{i=1}^{n}( y_i - f(\vx_i) )^2$. The update to $\mW$ using \texttt{GD} is given by:
\begin{align}
\label{eq:gd_update}
     \mW_{t+1} = \mW_{t} - \eta \mG_t,
\end{align}
where $\mW_t$ denotes the weights $\mW$ at step $t$ and $\mG_t = \nabla_{\mW_t} R(f, \mX, \vy)$ represents the full-batch gradient. 
Next, to formulate the updates using \texttt{FB-Adam},  let:
\begin{align}
\label{eq:adam_m_v}
    \widetilde{\mM}_{t+1} = \beta_1\widetilde{\mM}_t  + (1-\beta_1)\mG_t, \hspace{20pt} \widetilde{\mV}_{t+1} = \beta_2\widetilde{\mV}_t  + (1-\beta_2)\mG_t^{\circ 2}.
\end{align}
Here $\widetilde{\mM}_t, \widetilde{\mV}_t \in \sR^{h \times d}$ represent the first and second order moving averages of the gradient respectively, with base values $\widetilde{\mM}_0 = \vzero_{h \times d}, \widetilde{\mV}_0 = \vzero_{h \times d}$ \citep{kingma2014adam}. $(\beta_1, \beta_2) \in \sR$ are the decay factors. Considering $\widetilde{\mG}_{t} = (\widetilde{\mV}_{t+1}^{\circ 1/2} + \epsilon \vone_{h \times d} )^{\circ -1}\odot \widetilde{\mM}_{t+1}$, we formulate the \texttt{FB-Adam} update\footnote{We choose the subscript $t$ in $\widetilde{\mG}_{t}$ for notational consistency between \texttt{FB-Adam} and \texttt{GD} updates.}:
\begin{align}
\label{eq:fb_adam_update}
    \mW_{t+1} = \mW_{t} - \eta \widetilde{\mG}_{t}.
\end{align}
For the remainder of this paper, we use the overloaded term \textit{`optimizer update'} to represent either \texttt{FB-Adam} or \texttt{GD} update. Specific choices of the optimizer will be mentioned explicitly.

\paragraph{Ridge-regression on final layer.} Similar to the setup of \citet{ba2022high, moniri2023theory,wang2023spectral, ba2023learning}, we consider a hold-out training dataset $\overline{\mX} \in \sR^{n \times d}, \overline{\vy} \in \sR^n$ sampled in the same fashion as $\mX, \vy$ to learn the last layer weights. Formally, after $t$ optimizer updates to the first layer to obtain $\mW_t$, we calculate the post-activation features $\overline{\mZ}_t = \frac{1}{\sqrt{h}}\sigma\left(\frac{1}{\sqrt{d}}\mW_t\overline{\mX}^\top\right)$ and solve the following ridge-regression problem:
\begin{align}
\label{eq:second_layer_regression}
    \hat{\va} = \arg \min_{\va \in \sR^h} \frac{1}{n}\norm{\overline{\vy} - \overline{\mZ}_t^\top\va}_2^2 + \frac{\lambda}{h}\norm{\va}_2^2.
\end{align}
Here $\lambda > 0$ is the regularization constant. The solution $\hat{\va}$ is now used as the last-layer weight vector for our student network $f(\cdot)$ and we consider the resulting regression loss as our \textit{training loss}. Formally, this setup allows us to measure the impact of updates to $\mW$ after $t$ steps on the hold-out dataset's regression loss. Finally, for a test sample $\vx \in \sR^d$, the student network prediction is given as: $\hat{y} = \frac{1}{\sqrt{h}}{\hat{\va}}^{\top}\sigma\left(\frac{1}{\sqrt{d}}\mW_t\vx\right)$. These predictions on the test data are used for computing the \textit{test loss} using the mean squared error.

\subsection{Alignment and HT Metrics}
\label{sec:align_ht_metrics}

\paragraph{Alignment between $\mW, \vbeta^*$.} To quantify the ``extent'' of feature learning in our student network during training, we measure the alignment between the first principal component of $\mW$ (denoted as $\vu_1$) and the target direction $\vbeta^*$ \citep{ba2022high,wang2023spectral} as: \texttt{sim}$(\mW, \vbeta^*) = |\vu_1^\top \vbeta^*|$.
\paragraph{Kernel Target Alignment (\texttt{KTA}).} In addition to analyzing $\mW$, we also consider the alignment between the Conjugate Kernel (\textit{CK}) \citep{wang2023spectral, lee2018deep, matthews2018gaussian, fan2020spectra} based on hidden layer activations and the target outputs. Formally, consider the hidden layer activations of the holdout data as $\overline{\mZ} = \frac{1}{\sqrt{h}}\sigma\left(\frac{1}{\sqrt{d}}\mW\overline{\mX}^\top \right)$ and define \textit{CK} as: $\mK = \overline{\mZ}^\top\overline{\mZ} \in \sR^{n \times n}$. The \texttt{KTA} \citep{cristianini2001kernel} between $\mK$ and $\vy\vy^\top \in \sR^{n \times n}$ is given as:
\begin{align}
    \texttt{KTA} = \frac{\langle \mK, \vy\vy^\top \rangle}{\norm{\mK}_F\norm{\vy\vy^\top}_F}, \hspace{10pt} \langle \mK, \vy\vy^\top \rangle = \sum_{i,j}^n K_{i,j}(yy^\top)_{i,j}.
\end{align}
\paragraph{Power-law fits (\texttt{PL\_Alpha\_Hill}, \texttt{PL\_Alpha\_KS})} To quantify the heaviness of the tails in the ESD, we measure \texttt{PL\_Alpha\_Hill} \citep{zhou2023temperature}, and \texttt{PL\_Alpha\_KS}, which refer to the power-law exponents that are fit to the ESD of $\mW_t^\top\mW_t$ using the Hill estimator \citep{hill1975simple} and based on the Kolmogorov–Smirnoff statistic respectively \citep{martin2021predicting, clauset2009power}.

\section{Bulk+Spike Phenomenon after One Optimizer Step}
\label{sec:one_step_fb_adam_update}

In this section, we present empirical and theoretical results for the scale of $\eta$ for \texttt{FB-Adam} that results in the ``Bulk+Spike'' ESD and facilitates feature learning in the student network after the first update. The discussion on the first optimizer update and $\eta$ is essential, as we show in Section~\ref{sec:HT_phenomenon} that the Bulk+Spike ESD after the first update facilitates the emergence of HT ESDs.

\vspace{-2mm}
\subsection{The Bulk+Spike ESD by Scaling $\eta$}

\paragraph{Setup.} We consider the two-layer NN $f(\cdot)$ of width $h=1500, \sigma=\texttt{tanh}$ and train it on a dataset of size $n=2000$, with input dimension $d=1000$. We choose $\sigma_* = \texttt{softplus}$ as the target link function and set the label noise to $\rho_e = 0.3$, with $\lambda=0.01$. The test data consists of $200$ samples.

\begin{figure}
     \centering
     \begin{subfigure}[b]{0.24\textwidth}
         \centering
         \includegraphics[width=\textwidth]{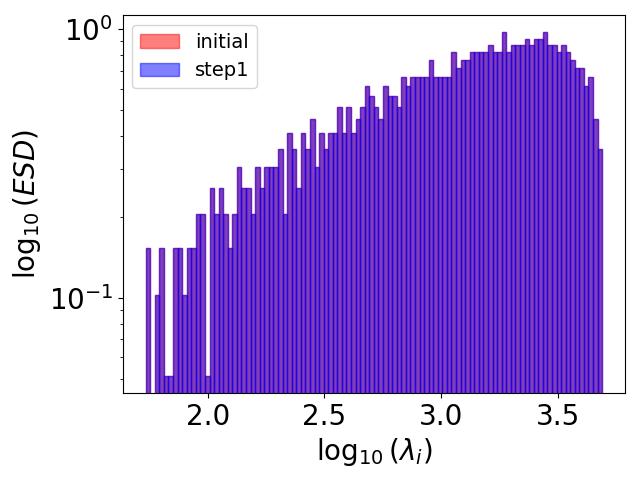}
         \caption{\texttt{GD} $\eta=0.1$}
         \label{fig:main:gd_fb_adam_W_esd_gd_lr0.1}
     \end{subfigure}
     \hfill
     \begin{subfigure}[b]{0.24\textwidth}
         \centering
         \includegraphics[width=\textwidth]{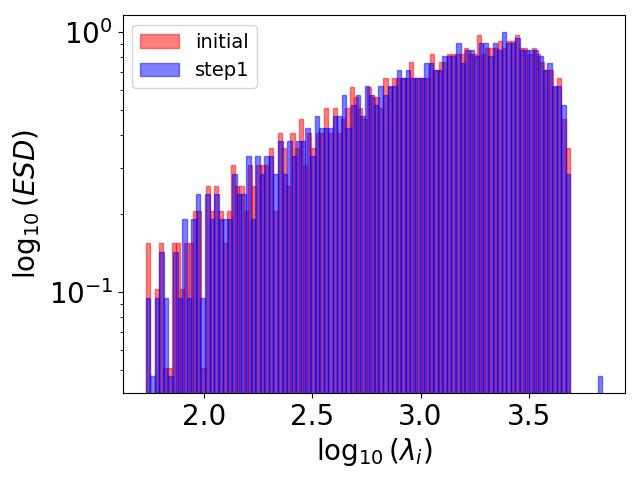}
         \caption{\texttt{GD} $\eta=2000$}
         \label{fig:main:gd_fb_adam_W_esd_gd_lr2000}
     \end{subfigure}
     \begin{subfigure}[b]{0.24\textwidth}
         \centering
         \includegraphics[width=\textwidth]{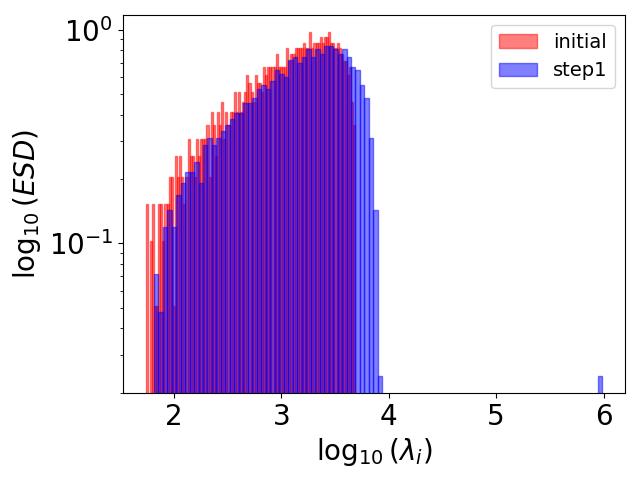}
         \caption{\texttt{FB-Adam} $\eta=1$}
         \label{fig:main:gd_fb_adam_W_esd_adam_lr0.1}
     \end{subfigure}
     \hfill
     \begin{subfigure}[b]{0.24\textwidth}
         \centering
         \includegraphics[width=\textwidth]{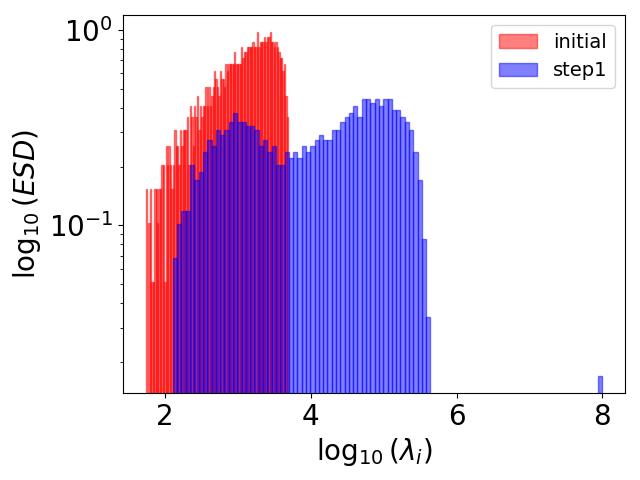}
         \caption{\texttt{FB-Adam} $\eta=10$}
         \label{fig:main:gd_fb_adam_W_esd_adam_lr10}
     \end{subfigure}
        \caption{ESD of $\mW_1^\top\mW_1$ for \texttt{GD/FB-Adam} with varying $\eta$, and $n=2000,d=1000,h=1500$,$\sigma_* = \texttt{softplus}$, $\sigma = \texttt{tanh}, \rho_e = 0.3, \lambda=0.01$.
}
        \label{fig:main:gd_fb_adam_W_esd}
        \vspace{-2mm}
\end{figure}

\paragraph{\texttt{FB-Adam} needs a much smaller $\eta$ than \texttt{GD} to exhibit a Bulk+Spike ESD.} By varying $\eta$, we train the student network $f(\cdot)$ for one step using \texttt{GD}/\texttt{FB-Adam}. Observe from Figure \ref{fig:main:gd_fb_adam_W_esd_gd_lr0.1} that after one \texttt{GD} update with $\eta=0.1$, the ESD of $\mW_1^\top\mW_1$ remains largely unchanged from that of $\mW_0^\top\mW_0$ (i.e ESD at random initialization). On the other hand, Figure \ref{fig:main:gd_fb_adam_W_esd_gd_lr2000} illustrates that for $\eta=2000$, the ESD of $\mW_1^\top\mW_1$ exhibits a spike. However, \texttt{FB-Adam} exhibits a spike in the ESD of $\mW_1^\top\mW_1$ after the first step even with $\eta=1$ (see Figure \ref{fig:main:gd_fb_adam_W_esd_adam_lr0.1}). Finally, for $\eta=10$, the ESD tends towards a seemingly bimodal distribution (see Figure \ref{fig:main:gd_fb_adam_W_esd_adam_lr10}).

\begin{figure}
     \centering
     \begin{subfigure}[b]{0.24\textwidth}
         \centering
         \includegraphics[width=\textwidth]{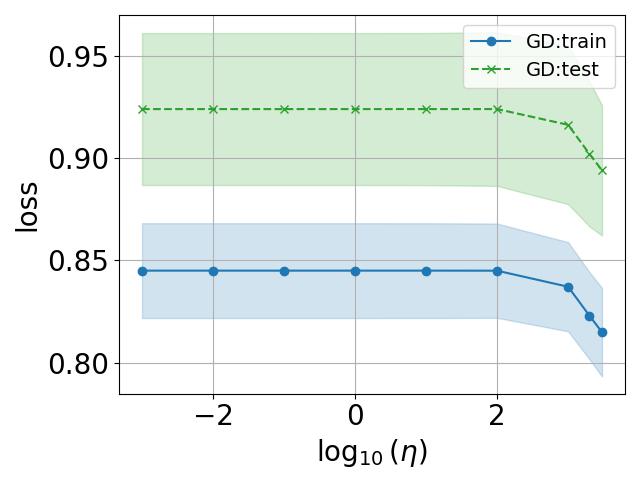}
         \caption{\texttt{GD} loss}
         \label{fig:gd_fb_adam_bulk_loss_alignments_1_step_n_2000_gd_loss}
     \end{subfigure}
     \hfill
     \begin{subfigure}[b]{0.24\textwidth}
         \centering
         \includegraphics[width=\textwidth]{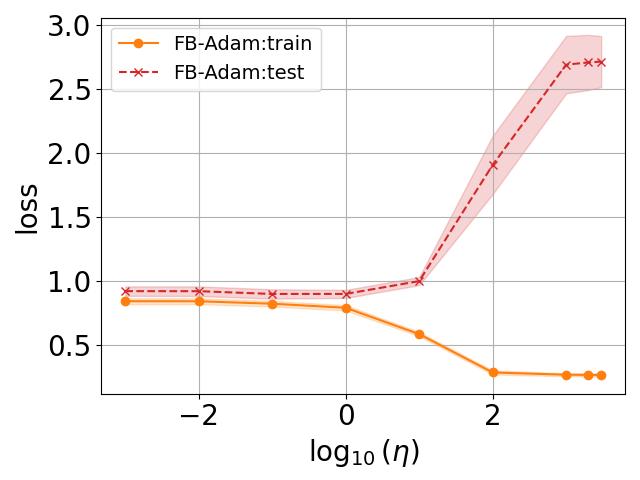}
         \caption{\texttt{FB-Adam} loss}
         \label{fig:gd_fb_adam_bulk_loss_alignments_1_step_n_2000_fb_adam_loss}
     \end{subfigure}
     \hfill
     \begin{subfigure}[b]{0.24\textwidth}
         \centering
         \includegraphics[width=\textwidth]{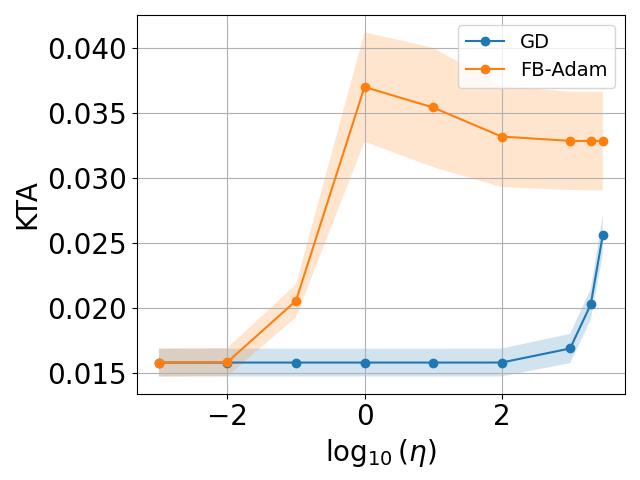}
         \caption{\texttt{KTA}}
         \label{fig:gd_fb_adam_bulk_loss_alignments_1_step_n_2000_kta}
     \end{subfigure}
     \hfill
     \begin{subfigure}[b]{0.24\textwidth}
         \centering
         \includegraphics[width=\textwidth]{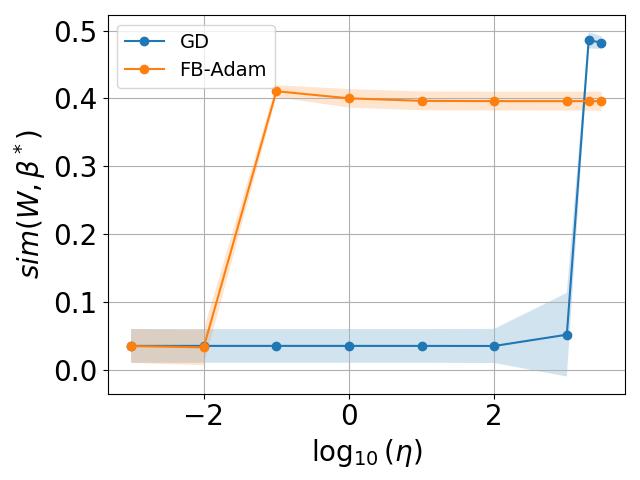}
         \caption{\texttt{sim}$(\mW, \vbeta^*)$}
         \label{fig:gd_fb_adam_bulk_loss_alignments_1_step_n_2000_sim}
     \end{subfigure}
        \caption{Losses, \texttt{KTA}, \texttt{sim}$(\mW, \vbeta^*)$  for $f(\cdot)$ trained with one-step of  \texttt{GD}, \texttt{FB-Adam}. Here $n=2000$, $d=1000$, $h=1500$, $\sigma_* = \texttt{softplus}, \sigma = \texttt{tanh}, \rho_e = 0.3, \lambda=0.01$.}
        \label{fig:gd_fb_adam_bulk_loss_alignments_1_step_n_2000}
    \vspace{-2mm}
\end{figure}

\paragraph{Impact on losses, \texttt{KTA}, and \texttt{sim}$(\mW, \vbeta^*)$.} As the choice of optimizer affects the scale of $\eta$ leading to a Bulk+Spike ESD of $\mW_1^\top\mW_1$, we vary $\eta$ across $\{ 0.001, 0.01, 0.1, 1, 10, 100, 1000, 2000, 3000\}$ and plot the means and standard deviations of losses, \texttt{KTA} and \texttt{sim}$(\mW, \vbeta^*)$ across $5$ runs in Figure \ref{fig:gd_fb_adam_bulk_loss_alignments_1_step_n_2000}. In the case of \texttt{GD}, observe that $\eta = 2000$ is the threshold for the: reduction of train and test losses (Figure \ref{fig:gd_fb_adam_bulk_loss_alignments_1_step_n_2000_gd_loss}), an increase in \texttt{KTA} (Figure \ref{fig:gd_fb_adam_bulk_loss_alignments_1_step_n_2000_kta}), and an increase in \texttt{sim}$(\mW, \vbeta^*)$ (Figure \ref{fig:gd_fb_adam_bulk_loss_alignments_1_step_n_2000_sim}). Thus, implying that the occurrence of a spike leads to better generalization after one step (as also verified by \citet{ba2022high}). In the case of \texttt{FB-Adam}, $\eta < 0.01$ does not improve generalization and we do not see an increase in \texttt{KTA}/\texttt{sim}$(\mW, \vbeta^*)$. For $0.01 \le \eta \le 1$, generalization improves, and \texttt{KTA}/\texttt{sim}$(\mW, \vbeta^*)$ values increase. For $\eta > 1$, generalization degrades and there is a slight reversal in the trend for \texttt{KTA}. Note that due to the two-phase training strategy, the large magnitude of the first step update with $\eta>1$ can lead to lower training loss (since the last layer is computed based on a closed-form solution), but lead to higher test loss.
These observations hint at a sweet spot for $\eta$ beyond which \texttt{FB-Adam} leads to poor test performance (see also Appendix~\ref{app:subsec:one_step}). Towards understanding these observations, we focus on the following pressing question: \textit{why does $\eta=0.1$ suffice for one-step of \texttt{FB-Adam} to exhibit a spike in the ESD, whereas one-step \texttt{GD} requires $\eta=2000$?} Specifically, how large should $\eta$ be for \texttt{FB-Adam} to exhibit a spike in the ESD?

\vspace{-2mm}
\subsection{Theoretical Results for Scaling $\eta$ with  \texttt{FB-Adam}}
\vspace{-2mm}

To theoretically answer the above questions, we start by formulating the first step gradient $\mG_0$ as:
\begin{align}
    \mG_0 &=  \frac{1}{n\sqrt{d}} \left[  \frac{1}{\sqrt{h}}\left(\va\vy^\top -\frac{1}{\sqrt{h}} \va\va^\top\sigma\left(\frac{1}{\sqrt{d}}\mW_0\mX^\top\right) \right) \odot \sigma'\left(\frac{1}{\sqrt{d}}\mW_0\mX^\top\right) \right]\mX.
\end{align}
Here  $\sigma'(\cdot): \sR \to \sR$ is the derivative of the activation function $\sigma$ acting element-wise on $\mX\mW_0^\top/\sqrt{d}$. Based on equation \eqref{eq:adam_m_v}, let $\widetilde{\mP}_1 = \widetilde{\mV}_{1}^{\circ 1/2} + \epsilon \vone_{h \times d}$. Considering the \texttt{FB-Adam} epsilon hyper-parameter $\epsilon \approx 0$ \citep{kunstner2022noise}, the scaled sign update \citep{bernstein2018signsgd} can be given as:
\begin{equation}
    \label{adam_one_step_update}
    \widetilde{\mG}_0 = \widetilde{\mP}_1^{\circ -1} \odot \widetilde{\mM}_1= \frac{1-\beta_1}{\sqrt{1-\beta_2}}\operatorname{sign}(\mG_0).
\end{equation}

\begin{theorem}
 \textit{Given the two-stage training procedure with large $n,d$, such that $n \asymp d$,  and large (fixed) $h$,   assume the teacher $F^*$ is $\lambda_\sigma$-Lipschitz with $\left\|F^*\right\|_{L^2}=\Theta_d(1)$, and a normalized `student' activation $\sigma$, which has $\lambda_\sigma$-bounded first three derivatives almost surely and satisfies $\mathbb{E}[\sigma(z)]=0, \mathbb{E}[z \sigma(z)] \neq 0$, for $z \sim \mathcal{N}(0,1)$; then the matrix norm bounds for the one-step \texttt{FB-Adam} update can be given as: }
\label{thm:fb_adam_norms}
\begin{equation}
     \norm{\widetilde{\mG}_0}_2=\Theta_{d, \mathbb{P}}(\sqrt{hd}),\hspace{10pt} \norm{\widetilde{\mG}_0}_F=\Theta_{d, \mathbb{P}}(\sqrt{hd}) .
\end{equation}
\end{theorem}

Appendix~\ref{app:thm:fb_adam_norms:proof} presents the proof. The sketch of the proof is as follows: First, we show that $\mG_0$ does not contain values that are exactly equal to $0$ almost surely. This allows us to obtain $\norm{\widetilde{\mG}_0}_F=\Theta_{d, \mathbb{P}}(\sqrt{hd})$. Next, we leverage a rank $1$ approximation $\mA$ of $\mG_0$ (in the operator norm \citep{ba2022high}) to show that $\operatorname{sign}(\mA) = \operatorname{sign}(\mG_0)$. Finally, we obtain the lower-bound of $\norm{\widetilde{\mG}_0}_2=\Omega_{d, \mathbb{P}}(\sqrt{hd})$ to prove the theorem \footnote{The normalized activation is not a practical limitation and is solely required for the proof \citep{ba2022high}.}. 
\begin{corollary}
\label{cor:lr_scale}
    Under the assumptions of Theorem \ref{thm:fb_adam_norms}, we have the following learning rate scales
    \begin{align}
    \begin{split}
        \eta = \Theta(1) \implies \norm{\mW_1 - \mW_0}_F \asymp \norm{\mW_0}_F \\
        \eta = \Theta\left(1/\sqrt{h}\right) \implies \norm{\mW_1 - \mW_0}_2 \asymp \norm{\mW_0}_2,
    \end{split}
    \end{align}
    where $\norm{\mW_0}_2 = \Theta_{d, \sP}(\sqrt{d}), \norm{\mW_0}_F = \Theta_{d, \sP}(\sqrt{hd})$.
\end{corollary}
Theorem \ref{thm:fb_adam_norms} shows that the spectral and Frobenius norms of $\widetilde{\mG}_0$ scale similarly and that the top singular value contributes the most to the Frobenius norm as $d,h$ increase. As a consequence, Corollary \ref{cor:lr_scale} indicates that $\eta = \Theta(1)$ (which is independent of $h,d$) is sufficient for $\eta\widetilde{\mG}_0$ to result in a `spike' in the ESD of $\mW_1^\top\mW_1$. This explains our empirical observations above where even $\eta=1$ was sufficient for the ESD to transition into a Bulk+Spike shape.
\paragraph{Remark.} We note that a similar result for $\eta$ in the case of \texttt{GD} was previously established by \citet{ba2022high}. However, they employ a mean-field initialization of the two-layer NN $f(\cdot)$ and differs from our NTK-based initialization \citep{wang2023spectral}. Nonetheless, we show that our results can be extended to such a setup as well and the adjusted $\eta$ can indeed explain the Bulk+Spike ESD after one step update (Appendix~\ref{app:mean_field_discussion} presents a comprehensive discussion).

\section{HT Phenomenon after Multiple Optimizer Steps}
\label{sec:HT_phenomenon}
In this section, we empirically analyze the ESD evolution from the ``Bulk+Spike'' shape into an HT distribution. This is followed by a correlation analysis with generalization of the two-layer NN.
\subsection{Transitioning from Bulk+Spike to HT ESD} 
\paragraph{Spike after one step facilitates the emergence of HT ESD.} We begin by employing the same setup as Section \ref{sec:one_step_fb_adam_update} and show in Table \ref{main:tab:fb_adam_esd_evolution} that the spike in the ESD after one step facilitates the emergence of HT ESD. We observe that after just $10$ \texttt{FB-Adam} steps with $\eta=1$, the HT ESD emerges. However, $\eta=0.1$ requires $t=10000$. This indicates that the relative position of the spike from the bulk (which is captured by the spectral gap in this setup) plays a key role in determining the step complexity of the HT ESD emergence \footnote{Theoretical analysis of the spectral gap and its role on the step complexity for HT ESD emergence is out of scope of this paper and can be a valuable direction for future research.}.Finally, for $\eta=0.01$, which does not exhibit an HT ESD even after $t=10000$, we observed that a significantly long amount of training for $t=10^6$ steps can lead to HT. Note that $\eta=0.01$ tends to fall within the $(\Theta(1/\sqrt{h}), \Theta(1))$ range, considering the scaling constants for $\Theta$. This leads to a scenario where a spike does not emerge after one step but the magnitude of gradient updates over $t=10^6$ steps can lead to HT ESD (see Table~\ref{tab:ADAM_transition_t_1000000}). Given this observation, extremely small $\eta$ might not be able to exhibit such changes to the ESD even after $t > 10^6$ steps (see Corollary 5.3 in \cite{wang2023spectral} for a related formal result). We extend the analysis to a wider range of $\eta$ in Appendix~\ref{app:sec:esd_evolution} (see Table \ref{tab:ADAM_transition}).

\begin{figure}[t!]
     \centering
     \begin{subfigure}[b]{0.24\textwidth}
         \centering
         \includegraphics[width=\textwidth]{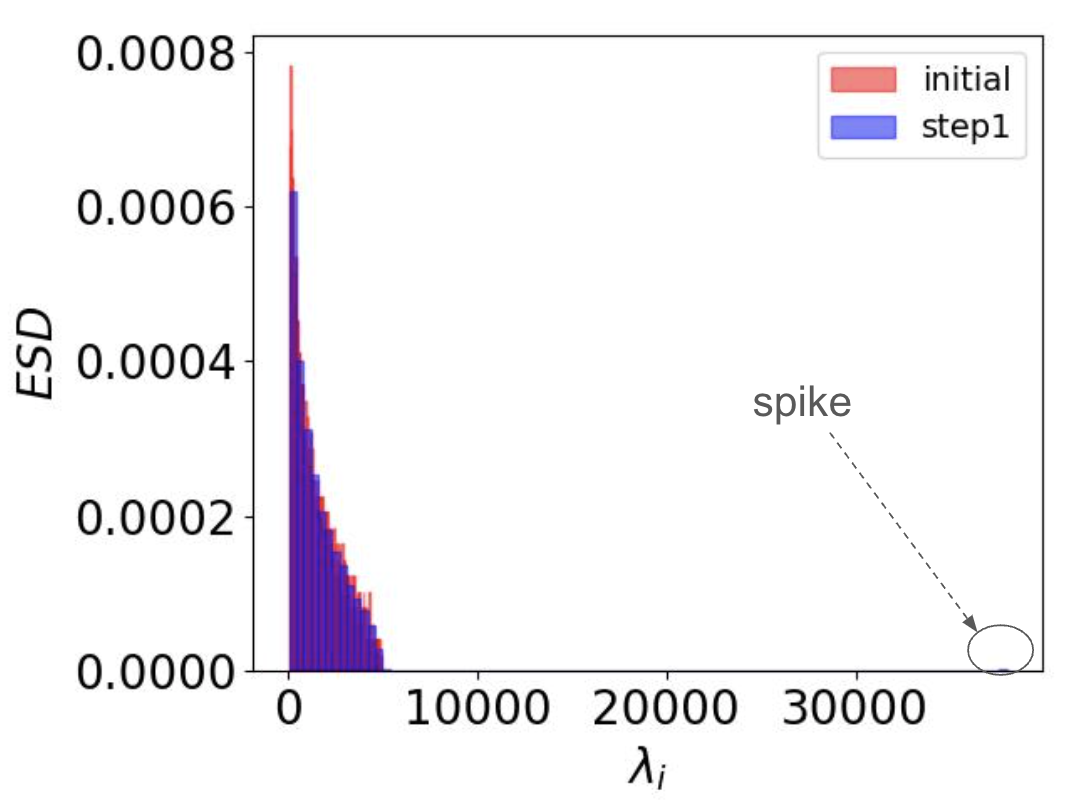}
         \caption{Bulk+Spike}
         \label{fig:bs_bd_ht_phases_bs}
     \end{subfigure}
     \hfill
     \begin{subfigure}[b]{0.24\textwidth}
         \centering
         \includegraphics[width=\textwidth]{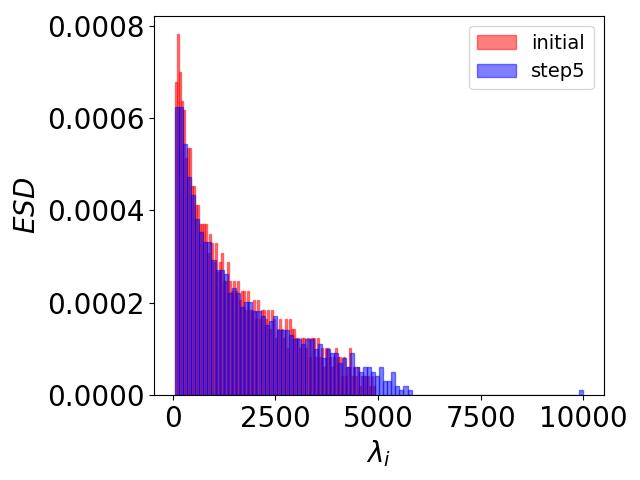}
         \caption{Bulk-Decay (onset)}
         \label{fig:bs_bd_ht_phases_bd1}
     \end{subfigure}
     \hfill
     \begin{subfigure}[b]{0.24\textwidth}
         \centering
         \includegraphics[width=\textwidth]{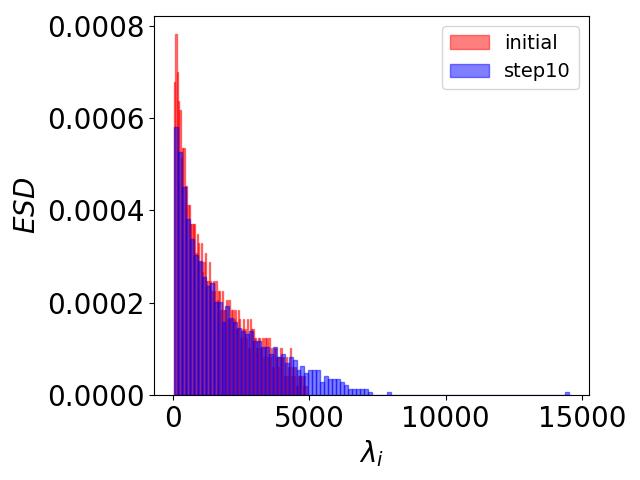}
         \caption{Bulk-Decay (progress)}
         \label{fig:bs_bd_ht_phases_bd2}
     \end{subfigure}
     \begin{subfigure}[b]{0.24\textwidth}
         \centering
         \includegraphics[width=\textwidth]{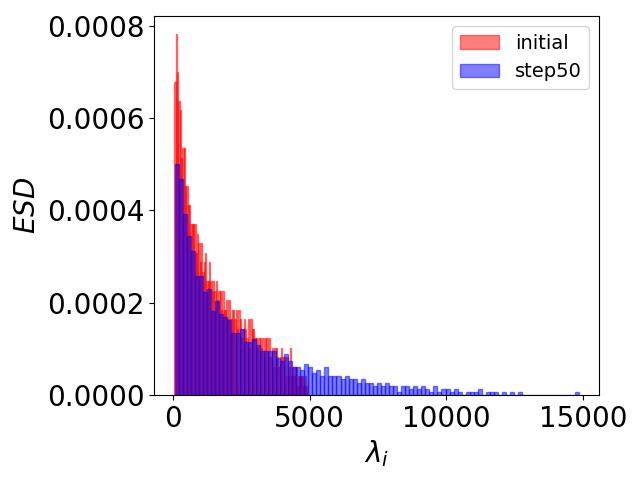}
         \caption{Heavy Tails (onset)}
         \label{fig:bs_bd_ht_phases_ht}
     \end{subfigure}
     \caption{Evolution of ESD (linear-linear scale) from a Marchenko-Pastur fit (at initialization $t=0$: red bulk) to (a) Bulk+Spike ($t=1$), (b) the onset of Bulk-Decay ($t=5$), (c) Continued decay of the bulk ($t=10$), (d) onset of heavy tails ($t=50$), with \texttt{FB-Adam} optimizer and $\eta=0.2$. }
     \label{fig:bs_bd_ht_phases}
     \vspace{-4mm}
\end{figure}

\paragraph{Bulk-Decay as an intermediate stage.} 
By considering \texttt{FB-Adam} with $\eta=0.2$ and the same setup as Section \ref{sec:one_step_fb_adam_update}, we illustrate in Figure \ref{fig:bs_bd_ht_phases} that the spike emerges after one-step (Figure \ref{fig:bs_bd_ht_phases_bs}) and gradually decays the Bulk (Figure \ref{fig:bs_bd_ht_phases_bd1}, Figure \ref{fig:bs_bd_ht_phases_bd2}) toward a HT distribution (Figure \ref{fig:bs_bd_ht_phases_ht}). Since practical settings employ deep NNs and train with stochastic gradient noise on complex datasets, the ESD evolution is much more nuanced. Especially the Bleeding-out phase and the presence of multiple spikes before the onset of Bulk-Decay. Nonetheless, we extended our results from the student-teacher setup to more practical settings, such as VGG, and ResNet models, and validated our HT ESD findings in those settings as well (see Section~\ref{subsec:applicability_deeper_nn}). Overall, we emphasize that our setup with two-layer NNs and single-index models opens up the possibility of approximating such dynamics while being subject to theoretical treatment.

\begin{table}[t!]
\centering
\setlength{\tabcolsep}{4pt} 
\renewcommand{\arraystretch}{1.5}

\begin{tabular}{|c|ccc|}
    \hline
    & $t = 1$ & $t = 100$ & $t = 10000$ \\ \hline
    \multirow{2}{*}{$\eta = 0.01$} & \adjustbox{valign=c}{\includegraphics[width=0.24\textwidth]{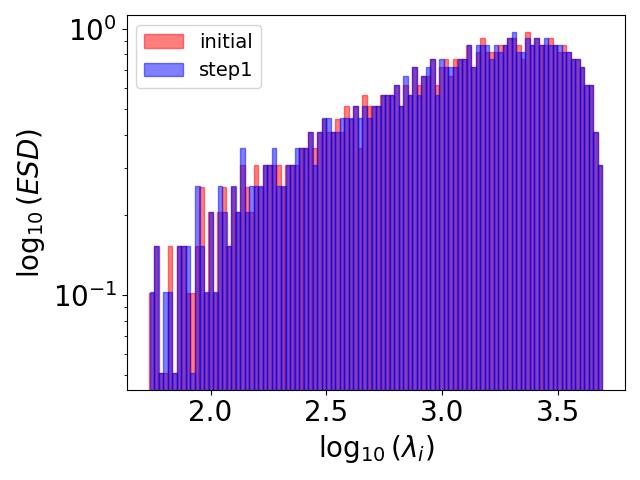}} & 
    \adjustbox{valign=c}{\includegraphics[width=0.24\textwidth]{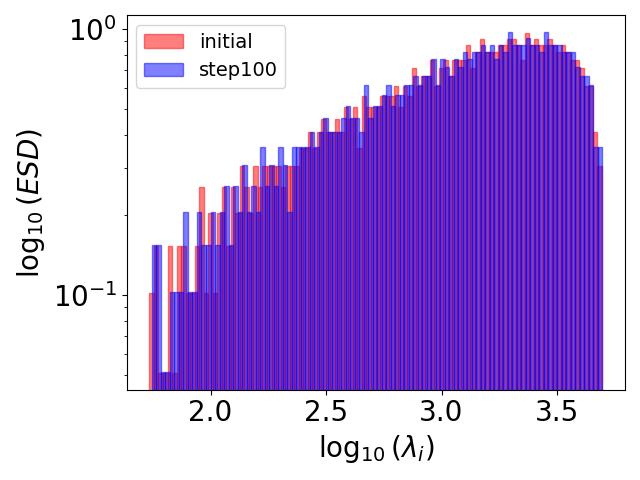}} & 
    \adjustbox{valign=c}{\includegraphics[width=0.24\textwidth]{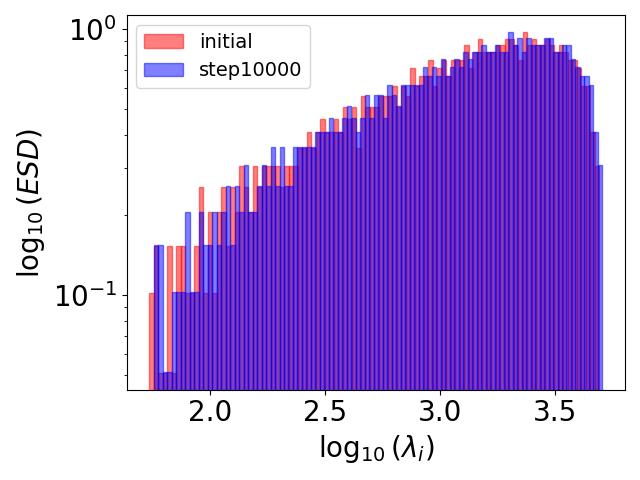}} \\ 
    \multirow{2}{*}{$\eta = 0.1$} & \adjustbox{valign=c}{\includegraphics[width=0.24\textwidth]{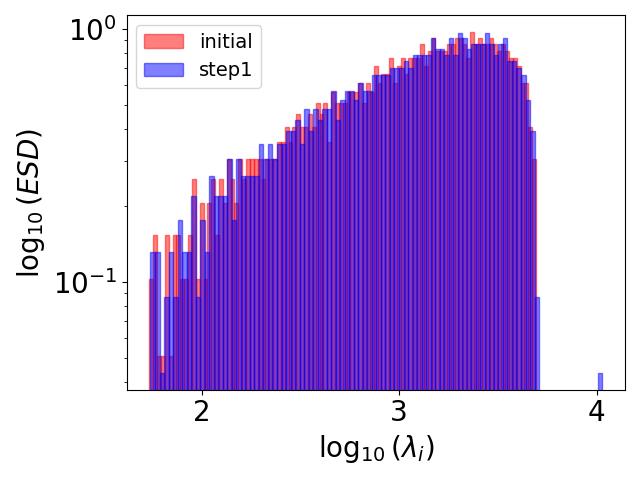}} & 
    \adjustbox{valign=c}{\includegraphics[width=0.24\textwidth]{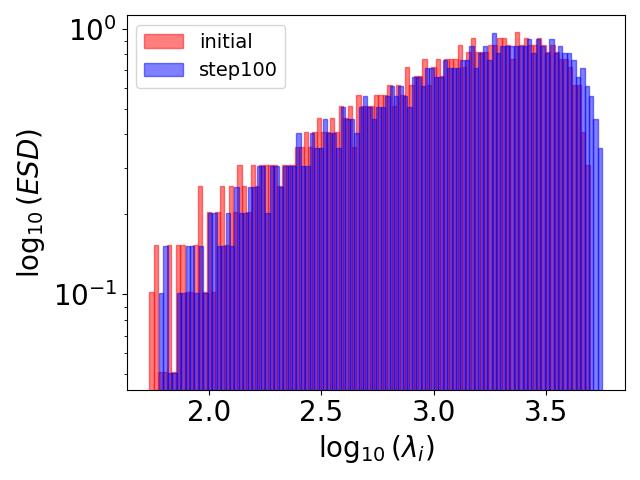}} & 
    \adjustbox{valign=c}{\includegraphics[width=0.24\textwidth]{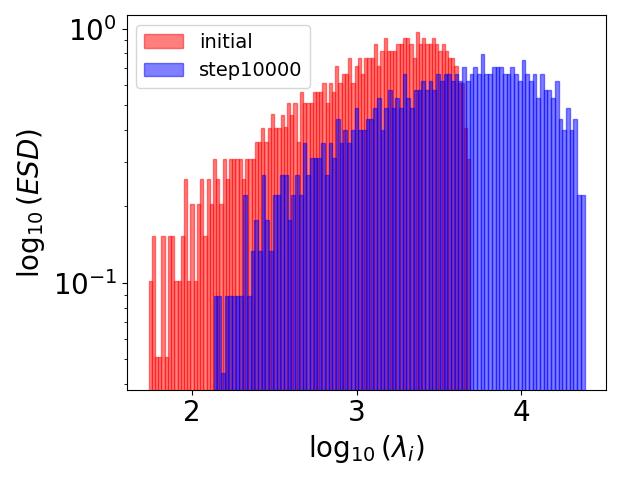}} \\ 
    \multirow{2}{*}{$\eta = 1$} & \adjustbox{valign=c}{\includegraphics[width=0.24\textwidth]{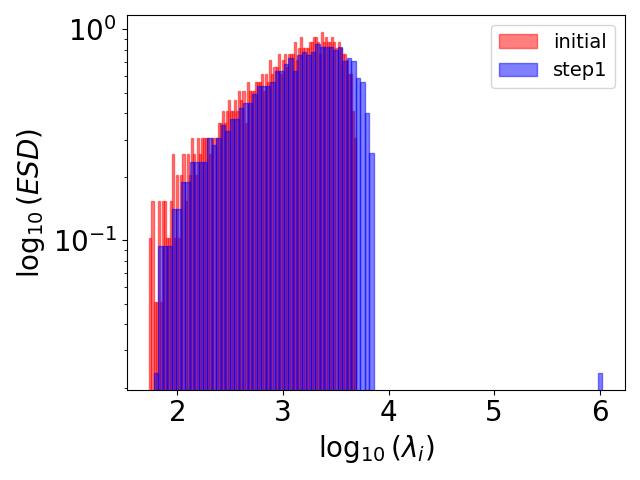}} & 
    \adjustbox{valign=c}{\includegraphics[width=0.24\textwidth]{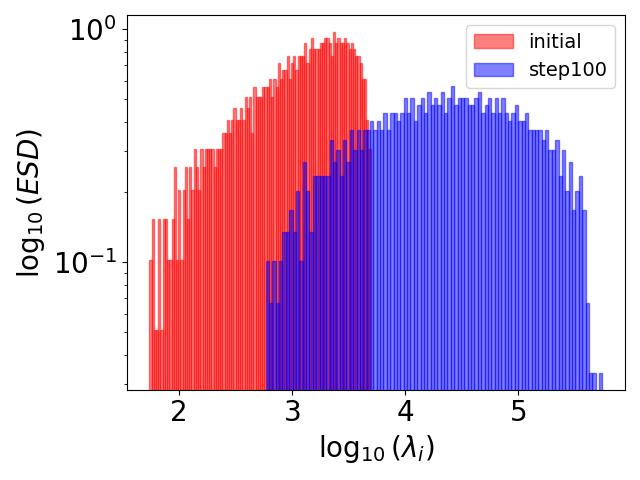}} & 
    \adjustbox{valign=c}{\includegraphics[width=0.24\textwidth]{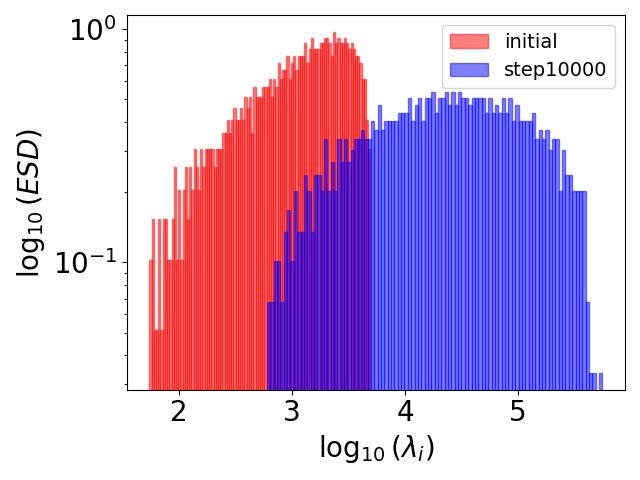}} \\ 
    \hline
\end{tabular}
  \caption{Evolution of ESD over steps $t \in \{1,100,10000\}$ with $\eta \in \{0.01,0.1,1\}$ for \texttt{FB-Adam}.}
\label{main:tab:fb_adam_esd_evolution}
\end{table}


\subsection{Impact on Generalization}
\label{Impact on Generalization}

\begin{figure}[t!]
     \centering
     \begin{subfigure}[b]{0.24\textwidth}
         \centering
         \includegraphics[width=\textwidth]{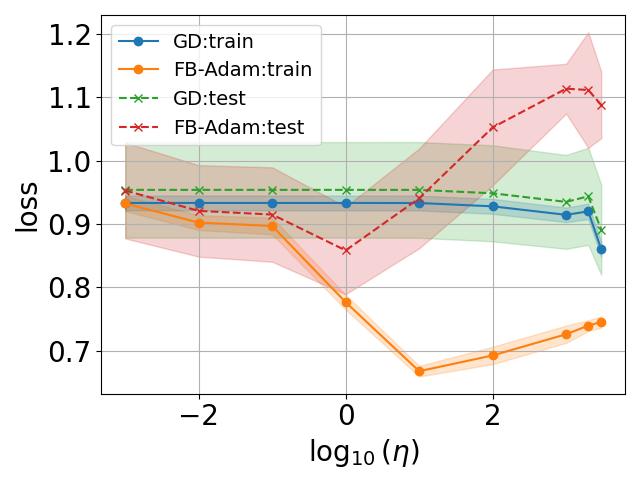}
         \caption{loss}
         \label{fig:gd_fb_adam_bulk_loss_alignments_10_steps_n_8000_loss}
     \end{subfigure}
     \hfill
     \begin{subfigure}[b]{0.24\textwidth}
         \centering
         \includegraphics[width=\textwidth]{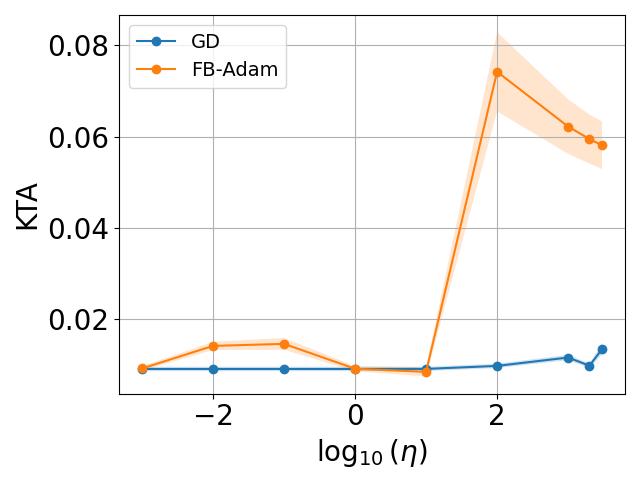}
         \caption{\texttt{KTA}}
         \label{fig:gd_fb_adam_bulk_loss_alignments_10_steps_n_8000_kta}
     \end{subfigure}
     \hfill
     \begin{subfigure}[b]{0.24\textwidth}
         \centering
         \includegraphics[width=\textwidth]{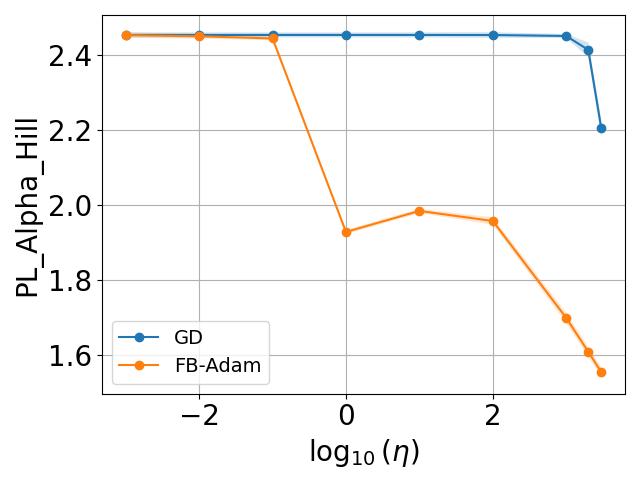}
         \caption{\texttt{PL\_Alpha\_Hill}}\label{fig:gd_fb_adam_bulk_loss_alignments_10_steps_n_8000_alpha_hill}
     \end{subfigure}
     \begin{subfigure}[b]{0.24\textwidth}
         \centering
         \includegraphics[width=\textwidth]{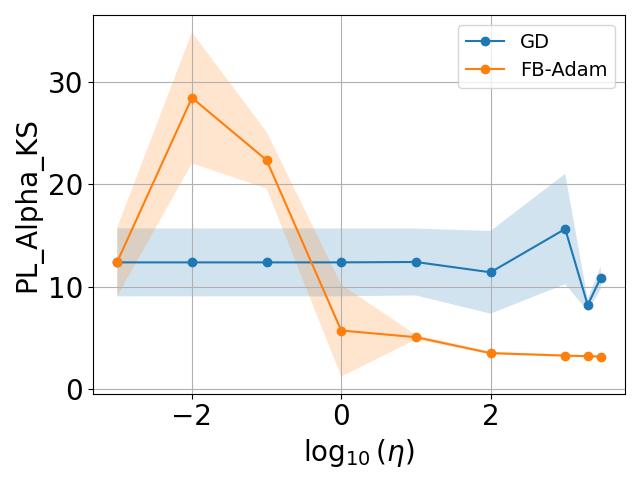}
         \caption{\texttt{PL\_Alpha\_KS}}
         \label{fig:gd_fb_adam_bulk_loss_alignments_10_steps_n_8000_alpha_ks}
     \end{subfigure}
        \caption{Losses, \texttt{KTA}, \texttt{PL\_Alpha\_Hill}, \texttt{PL\_Alpha\_KS} after $10$ steps of \texttt{GD}, \texttt{FB-Adam}, with $n=8000$, $d=1000$, $h=1500$, $\sigma_* = \texttt{softplus}, \sigma = \texttt{tanh}, \rho_e = 0.3, \lambda=0.01$.}
\label{fig:gd_fb_adam_bulk_loss_alignments_10_steps_n_8000}
\vspace{-3mm}
\end{figure}

\paragraph{Setup.} We consider the two-layer student NN $f(\cdot)$ of width $h=1500, \sigma=\texttt{tanh}$ and train it on a dataset of size $n=8000$ for $10$ steps using \texttt{GD}/\texttt{FB-Adam}. We choose a sample dimension $d=1000$, $\sigma_* = \texttt{softplus}$ and $\rho_e = 0.3$. The test dataset has $200$ samples.
\paragraph{Correlations between ESD and losses.} We plot the means and standard deviations across $5$ runs for the train/test losses, \texttt{KTA}, \texttt{PL\_Alpha\_Hill} and \texttt{PL\_Alpha\_KS} (see Section~\ref{sec:align_ht_metrics}) of $\mW_{10}^\top\mW_{10}$ after $10$ \texttt{GD}/\texttt{FB-Adam} updates by varying $\eta$ across $\{ 0.001, 0.01, 0.1, 1, 10, 100, 1000, 2000, 3000\}$ in Figure \ref{fig:gd_fb_adam_bulk_loss_alignments_10_steps_n_8000}. A lower value of \texttt{PL\_Alpha\_Hill} / \texttt{PL\_Alpha\_KS} indicates a heavier-tailed spectrum\footnote{The effects of the power-law exponent estimation approaches are discussed in Appendix~\ref{app:sec:pl_estimator_note}.}. Observe that for baseline \texttt{GD} experiments with $\eta \ge 1000$, the reduction in train and test losses are correlated with an increase in \texttt{KTA} and a decrease in \texttt{PL\_Alpha\_Hill} and \texttt{PL\_Alpha\_KS}. Additional experiments are presented in Appendix~\ref{app:subsec:10_steps}.

In the case of \texttt{FB-Adam}, a much clearer correlation between the training loss, \texttt{KTA}, \texttt{PL\_Alpha\_Hill} and \texttt{PL\_Alpha\_KS} can be observed. Especially, there seems to be a region of benign learning rates ($0.01 \le \eta \le 1$) for which, the \texttt{PL\_Alpha\_Hill} estimates lie in the range of $(2, 2.5)$ and a decrease in the estimate (resulting in a `heavier' tailed ESD) improves generalization.   For $1 \le \eta \le 100$, although we observe similar values of \texttt{PL\_Alpha\_Hill}, the ESDs of $\mW_{10}^\top\mW_{10}$ differ in the scale of the singular values, and the spike seems to have a large influence on the estimation of \texttt{PL\_Alpha\_Hill} (see Figure \ref{fig:app:10_step_fb_adam_W_esd_n_8000_eta_1_to_1000} in Appendix~\ref{app:add_exp}). However, the \texttt{PL\_Alpha\_KS} captures the monotonically decreasing trend for this range of $\eta$. Finally, for extremely large $\eta > 100$, we observe much smaller estimates of \texttt{PL\_Alpha\_Hill} ($< 1.8)$ but these extremely heavier tails do not correlate with better generalization. We also note that these benign $\eta$ ranges vary based on the choice of the activation functions. In particular, when $\sigma=\sigma_*=\texttt{tanh}$, we observed that the range can be reduced by an order of magnitude (Appendix~\ref{app:sec:same_act}). Overall, these observations support the conclusions of \citet{martin2021implicit, martin2021predicting} which state that well-trained deep NNs do not exhibit extreme HT ESDs but rather whose power-law estimates lie within a suitable range. For instance, a range of $(2,4)$ for the truncated power-law fit estimates. 

\paragraph{A note on the suitable range of tail index values}
Previous works such as \cite{simsekli2020hausdorff, hodgkinson2022generalization} studied the limiting distribution of the weight matrix values by modeling the stochastic gradient noise using Levy/Feller processes. In particular, they showed that smaller tail-index values lead to smaller generalization errors. The underlying idea is that when the weight values tend to an HT distribution, then the resulting ESD is also an HT (see \cite{arous2008spectrum}). Given this explanation, we note that the theoretical results by \cite{simsekli2020hausdorff, hodgkinson2022generalization} were shown to hold in practice by their numerical experiments, where the tail index tends to lie in a suitable range (depending on the PL alpha fit approaches) and represents heavy-tailed noise for good generalization.
Even though our setting differs significantly from these previous works and avoids gradient noise, the suitability of such a range seems to hold. In particular, for our setup the \texttt{PL\_Alpha\_Hill} estimates lie in the range of $(2, 2.5)$ where a lower value implies a `heavier' tailed ESD, and improves generalization of student network.

\paragraph{Remark.} Since practical training approaches employ techniques such as weight normalization (WN) and learning rate schedules, we present a preliminary analysis of their role in the ESD evolution and generalization in Appendix~\ref{app:sec:wn_lr}. We employ a WN technique \citep{huang2023normalization} after each update
and observed that it leads to relatively heavier-tailed spectra (i.e. lower \texttt{PL\_Alpha\_Hill}) while exhibiting similar correlations with generalization as discussed above. On the other hand, by employing schedulers such as \texttt{torch.optim.StepLR}, we showcase a fine-grained manipulation of the ESD evolution depending on the decay rate $\gamma$.


\begin{figure}[t]
     \centering
     \begin{subfigure}[b]{0.24\textwidth}
         \centering
         \includegraphics[width=\textwidth]{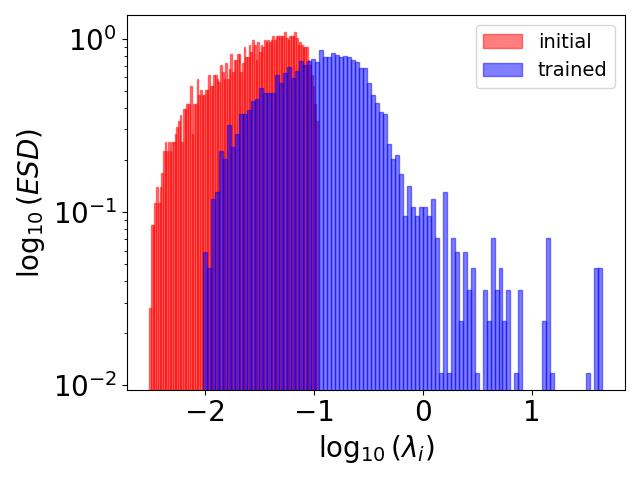}
         \caption{ VGG11-L4 (CIFAR10) } 
     \end{subfigure}
     \hfill
     \begin{subfigure}[b]{0.24\textwidth}
         \centering
         \includegraphics[width=\textwidth]{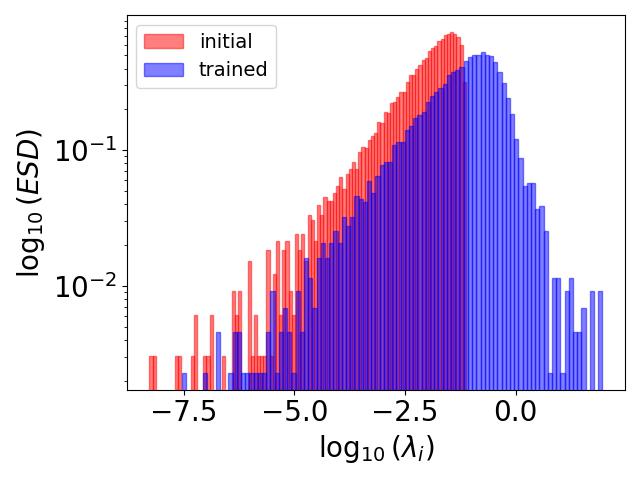}
         \caption{VGG11-L7 (MNIST) } 
     \end{subfigure}
     \begin{subfigure}[b]{0.24\textwidth}
         \centering
         \includegraphics[width=\textwidth]{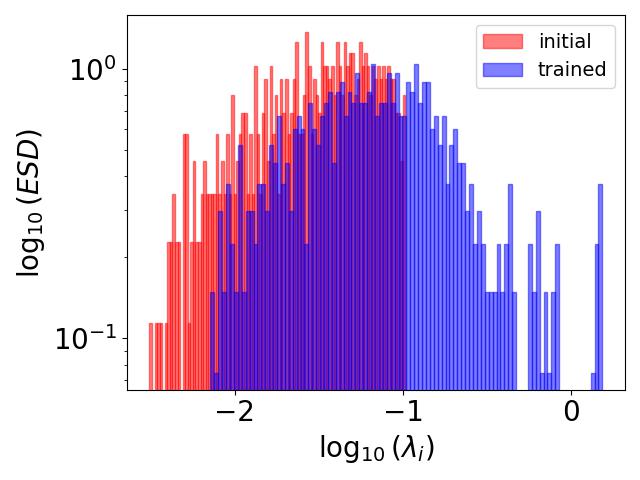}
         \caption{ ResNet18-L5 (CIFAR10) } 
     \end{subfigure}
     \hfill
     \begin{subfigure}[b]{0.24\textwidth}
         \centering
         \includegraphics[width=\textwidth]{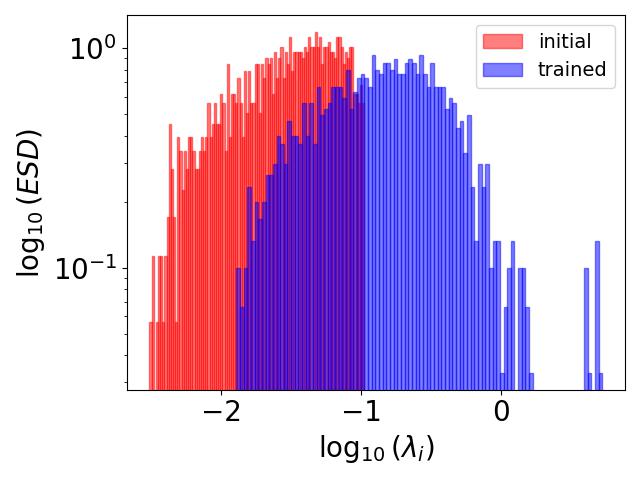}
         \caption{ ResNet18-L10 (SVHN) } 
     \end{subfigure}
     \caption{ESDs in VGG11, ResNet18 after $10$ steps with \texttt{FB-Adam}($\eta=0.01$) on MNIST, CIFAR10, SVHN.}
     \label{fig:esd_deeper_nn_main}
\end{figure}

\subsection{Applicability of Insights to Deeper NNs}
\label{subsec:applicability_deeper_nn}
We verify our findings on full-batch training and optimizer-dependent learning rates for the emergence of HT ESD with \texttt{FB-Adam} in practical deep NNs. We choose the VGG11 and ResNet18 models to train on the CIFAR10, MNIST and SVHN datasets. We avoid (very) large NNs and datasets due to GPU memory constraints for full-batch training and instead focus on the paper's main theme. Figure~\ref{fig:esd_deeper_nn_main} illustrates the ESDs of selected layers in the deeper NNs and indicates the emergence of HT ESDs during the early phases of training (i.e just after $10$ steps) (See Appendix~\ref{app:sec:vgg_mnist} for further illustrations). The main takeaway from these results is the observation that such emergence in practical NNs is \textit{not limited to training till convergence with mini-batches} \citep{martin2021implicit}. Furthermore, since the power-law fits have been used to determine if a layer is under-trained or over-trained \cite{martin2021predicting, zhou2023temperature}, our results highlight that such connections hold more nuance than considered by previous efforts. Thus
opening up a wide range of avenues to explore the interplay between the ESDs and training convergence (see also Appendix~\ref{app:sec:fw}).

\section{Conclusion}
This paper presents a different angle to study the emergence of HT ESDs during NN training. Unlike existing studies that mostly focused on the mini-batch settings, we show that full-batch \texttt{GD} or \texttt{Adam} can still lead to HT ESDs in the weight matrices after only a few optimizer updates with large $\eta$. Our paper also connects with several ongoing studies in this field. In particular, our study analyzes the `$5+1$' phase model~\citep{martin2021implicit} of ESD evolution and sheds light on the transitions from a Marchenko-Pastur (MP) fit (i.e. random initialization) $\to$ ``Bulk+Spike'' $\to$ ``Bulk-Decay'' $\to$ HT distributions. We also showcase that the ``Bulk-Decay'' ESD can be treated as an intermediate state generated from diffusing the spike into the main bulk (see also Appendix ~\ref{spike movement}). Our paper also presents several surprising phenomena: (1) the emergence of the HT spectra seems to require only a single spike aligned with the teacher model; (2) the emergence of the HT spectra can appear early during training, way before the NN reaches a low training loss; (3) several factors, such as weight normalization and learning rate scheduling, can all contribute to the emergence of HT ESDs. Overall, our work connects the feature learning aspect of the bulk + spike ESD (Section~\ref{sec:one_step_fb_adam_update}) with the early emergence of HT ESDs (Section~\ref{sec:HT_phenomenon}) and presents a unique perspective to study the correlations between HT ESDs and generalization in NNs (especially with \texttt{FB-Adam}).

\section*{Acknowledgments}
This work is supported by DOE under Award Number DE-SC0025584 and Dartmouth College.

\bibliography{references}
\bibliographystyle{tmlr}

\appendix

\newpage
\section{Generating \emph{Very} Heavy-Tailed Spectra without Gradient Noise}\label{app:heaviest}

\label{fig:app: more heavy_spectra}
\begin{figure}[h!]
     \centering
     \begin{subfigure}[b]{0.32\textwidth}
         \centering
         \includegraphics[width=\textwidth]{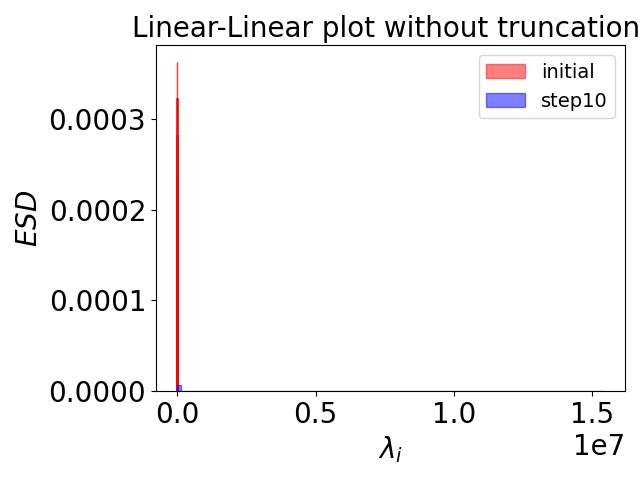}
         \caption{ \texttt{GD} $\eta=5000$ \label{fig:not_clear} } 
     \end{subfigure}
     \hfill
     \begin{subfigure}[b]{0.32\textwidth}
         \centering
         \includegraphics[width=\textwidth]{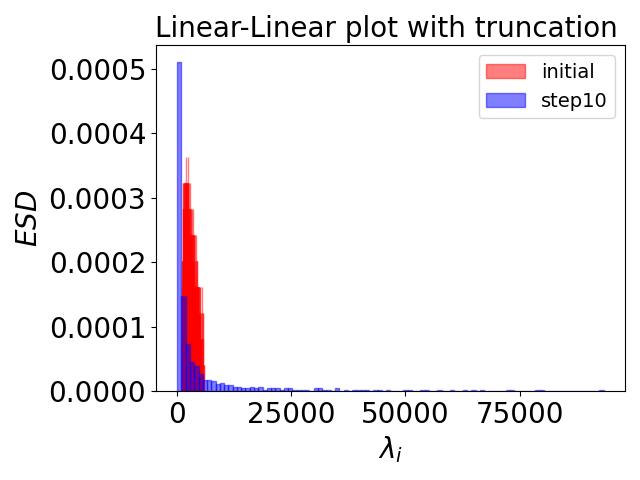}
         \caption{ \texttt{GD} $\eta=5000$   } 
     \end{subfigure}
     \hfill
     \begin{subfigure}[b]{0.32\textwidth}
         \centering
         \includegraphics[width=\textwidth]{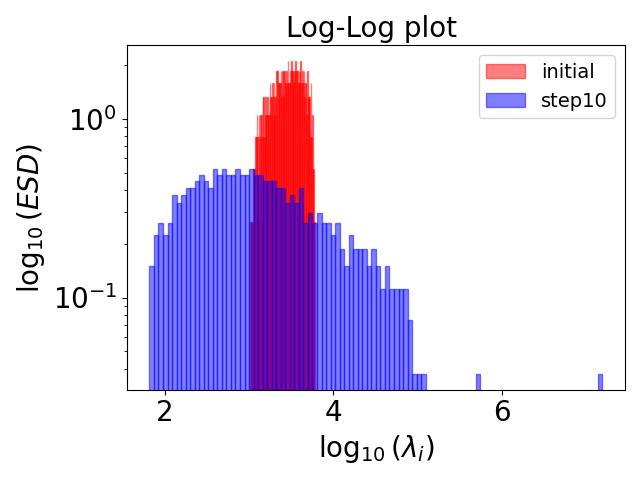}
         \caption{ \texttt{GD} $\eta=5000$ } 
     \end{subfigure}

     \begin{subfigure}[b]{0.32\textwidth}
         \centering
         \includegraphics[width=\textwidth]{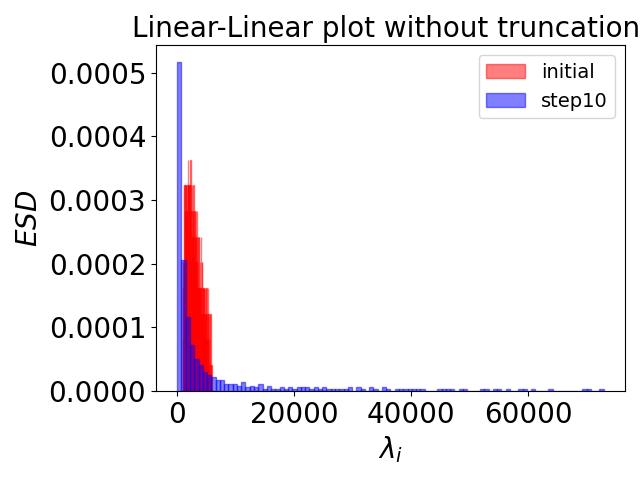}
         \caption{ \texttt{FB-Adam} $\eta=3$   } 
     \end{subfigure}
     \hfill
     \begin{subfigure}[b]{0.32\textwidth}
         \centering
         \includegraphics[width=\textwidth]{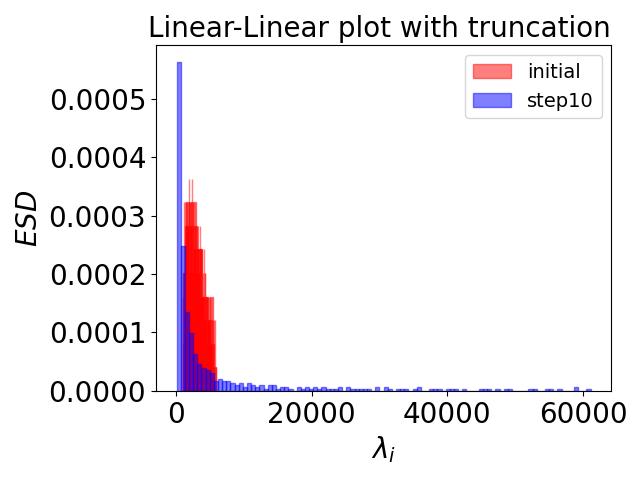}
         \caption{ \texttt{FB-Adam} $\eta=3$  } 
     \end{subfigure}
     \hfill
     \begin{subfigure}[b]{0.32\textwidth}
         \centering
         \includegraphics[width=\textwidth]{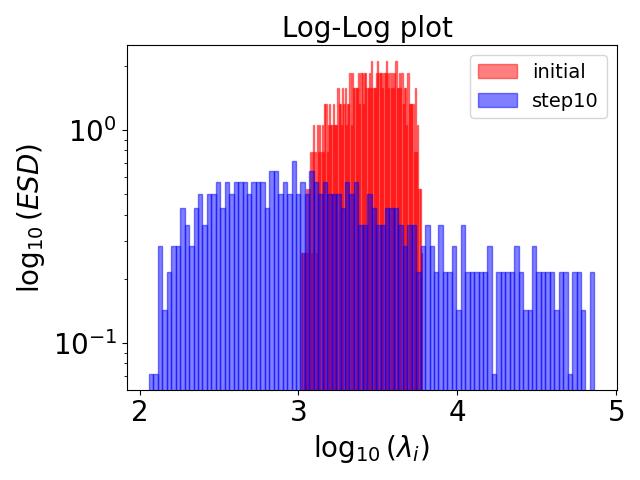}
         \caption{ \texttt{FB-Adam} $\eta=3$ } 
     \end{subfigure}
        \caption{ Emergence of HT spectra after 10 \texttt{GD}/\texttt{FB-Adam} steps with weight normalization. Here $n=4000$, $d=500$, $h=3000$, $\sigma_* = \texttt{softplus}, \sigma = \texttt{tanh}, \rho_e = 0.3$.   \texttt{PL\_Alpha\_KS}  for \texttt{GD} (first row) is 1.58, and \texttt{PL\_Alpha\_KS} for \texttt{FB-Adam} (second row) is 1.59. }
    \label{fig:very_HT}

\end{figure}

In this section, we present two examples of highly HT ESD, shown in Figure~\ref{fig:very_HT}. If one only observes the linear-linear plot (first column), the shapes can be invisible due to large eigenvalues. For example, see Figure~\ref{fig:not_clear}. After truncation, they become clearer (middle column). To quantitatively verify our observations, we adopted the WeightWatcher~\citep{martin2021predicting} Python APIs to fit a power-law (PL) distribution to these ESDs. We find that both ESDs have a PL coefficient \texttt{PL\_Alpha\_KS} smaller than~2. As discussed in~\citet[Table 3]{martin2021implicit}, this indicates that these ESDs have entered the \emph{very} HT regime. More importantly, these ESDs are generated without gradient noise after only 10 \texttt{GD}/\texttt{FB-Adam} steps.

\section{Proof of Theorem \ref{thm:fb_adam_norms}}
\label{app:thm:fb_adam_norms:proof}

In this section, we prove Theorem \ref{thm:fb_adam_norms}, and conduct numerical simulations for empirical verification. We begin by stating and discussing the main assumptions of the theorem.

\begin{assumption}
\label{asmpt:gauss_init}
\textbf {Gaussian Initialization.} The entries of the weights are sampled independently as  $\left[\mW_0\right]_{i j} \stackrel{\text { i.i.d. }}{\sim} \mathcal{N}(0,1) $ and $\left[a\right]_{i} \stackrel{\text { i.i.d. }}{\sim} \mathcal{N}(0,1), \forall i \in [h], j \in [d]$.
\end{assumption}

\begin{assumption}
\label{asmpt:norm_act}
\textbf {Normalized Activation.} The nonlinear activation $\sigma$ has $\lambda_\sigma$-bounded first three derivatives almost surely. In addition, $\sigma$ satisfies $\mathbb{E}[\sigma(z)]=0, \mathbb{E}[z \sigma(z)] \neq 0$, for $z \sim \mathcal{N}(0,1)$.
\end{assumption}

\textbf{Discussion.} Consider $\sigma: \sR \to \sR$ to be the $\texttt{tanh}$ function and $z \sim \mathcal{N}(0,1)$. Since $\sigma(z)=\texttt{tanh}(z)=\frac{e^z-e^{-z}}{e^z+e^{-z}}$, it is easy to check that $\sigma(z)$ has $1-$bounded first three derivatives and it is an odd function satisfying $\sigma(z)=-\sigma(-z)$, we have:
\begin{align}
    \mathbb{E}_{p(z)}[\sigma(z)]&=\int_{-\infty}^{\infty}\sigma(z)p(z)\mathrm{d}z =\int_{-\infty}^{0}\sigma(z)p(z)\mathrm{d}z+\int_{0}^{\infty}\sigma(z)p(z)\mathrm{d}z=0,
\end{align}
where $p(z)=\frac{1}{\sqrt{2 \pi} } e^{-\frac{z^2}{2}}$. Next, observe that
\begin{equation}
\label{eq:ztanh_ineq}
z \texttt{tanh}(z)=\frac{z(e^z-e^{-z})}{e^z+e^{-z}} \geq 0,
\end{equation}
where $z \texttt{tanh}(z)=0$ holds only when $z=0$. Let $\varepsilon>0$ and expand the expectation $\mathbb{E}_{p(z)}[z\sigma(z)]$ as follows:
\begin{align}
\begin{split}
    \mathbb{E}_{p(z)}[z\sigma(z)] &= \int_{-\infty}^{\infty}z\sigma(z)p(z)\mathrm{d}z =\int_{-\infty}^{0-\varepsilon}z\sigma(z)p(z)\mathrm{d}z + \int_{0+\varepsilon}^{\infty}z\sigma(z)p(z)\mathrm{d}z + \int_{0-\varepsilon}^{0+\varepsilon}z\sigma(z)p(z)\mathrm{d}z.\\
\end{split}
\end{align}
From equation \ref{eq:ztanh_ineq}, the first two terms are $> 0$, i.e:
\begin{equation}
    \int_{-\infty}^{0-\varepsilon}z\sigma(z)p(z)\mathrm{d}z > 0, \hspace{20pt}  \int_{0+\varepsilon}^{\infty}z\sigma(z)p(z)\mathrm{d}z>0
\end{equation}
whereas for $\varepsilon \to 0$, the third term can be bounded as:
\begin{equation}
    \int_{0-\varepsilon}^{0+\varepsilon}z\sigma(z)p(z)\mathrm{d}z \geq 0
\end{equation}
By combining the above results, $\sigma$ satisfies: $\mathbb{E}_{p(z)}[z\sigma(z)]>0 \Rightarrow    \mathbb{E}_{p(z)}[z\sigma(z)]\neq 0$.  \hfill

\begin{assumption}
\label{asmpt:ST}
    \textbf{Teacher-Student Setup.} The target labels are generated by the single index teacher model as $y_i=F^*\left(\vx_i\right)+\xi_i$, where $\vx_i \stackrel{\text { i.i.d. }}{\sim} \mathcal{N}(0, \mI), \xi_i$ is i.i.d. Gaussian noise with mean 0 and variance $\rho_{e}^2$, and the teacher $F^*$ is $\lambda_\sigma$-Lipschitz with $\norm{F^*}_{L^2}=\Theta_d(1)$.
\end{assumption}

\textbf{Discussion.}  Recall that our teacher model is given by $F^*(\vx_i) = \sigma_*( \vbeta^{*\top} \vx_i)$, where $\vbeta^* \in \sS^{d-1}, \vx_i \in \sR^d, i \in [n]$, and $\sigma_*(z)=\log(1+e^z)$ is the \texttt{softplus} function. Note that the derivative $\sigma_*^\prime(z)=\frac{e^z}{1+e^z}<1$ is bounded, and $ \norm{\vbeta^{*\top} \vx_i - \vbeta^{*\top} \vx_j}_2 \leq \norm{\vbeta^*}_2 \norm{\vx_i - \vx_j}_2, \forall \vx_i, \vx_j \in \sR^d$. This gives us:
\begin{equation}
     \norm{\sigma_*(\vbeta^{*\top} \vx_i) - \sigma_*(\vbeta^{*\top} \vx_j)}_2 \leq \norm{ \vbeta^{*\top} \vx_i - \vbeta^{*\top} \vx_j }_2 \leq \norm{\vbeta^*}_2 \norm{\vx_i -\vx_j}_2, \hspace{5pt} \forall \vx_i, \vx_j \in \sR^d,
\end{equation}
which implies that $F^*$ is a $\norm{\vbeta^*}_2$-Lipschitz function. Next, we consider $z=\vbeta^{*\top} \vx$, for $\vx \sim \gN(\vzero,\mI_d)$, which implies $z \sim \gN(0,\norm{\vbeta^*}_2^2)$, and bound $\sigma_*(z)$ by $0 < \sigma_*(z) < g_{\sigma_*}(z)$, where $g_{\sigma_*}(z)$ is:
\begin{equation}
    g_{\sigma_*}\left( z \right) =\begin{cases}
	1, &             z < 0\\
	z+1, &    z \geq 0.\\
\end{cases}
\end{equation}
Based on these results, we calculate $\norm{F^*}_{L^2}$ as follows.
\begin{align}
    \norm{F^*}_{L^2}^2 = \int_{\sR} \sigma_*(z)^2 \mathrm{d} \mu <\int_{\sR} g_{\sigma_*}(z)^2 \mathrm{d} \mu
\end{align}
where $\mathrm{d}\mu = \frac{1}{\sqrt{2 \pi} \norm{\vbeta^*}_2} e^{-\frac{z^2}{2 \norm{\vbeta^*}_2^2}} \mathrm{d}z$ is the Gaussian measure. Further expansion of the upper bound gives:

\begin{align*}
\begin{split}
    \int_{\sR} g_{\sigma_*}(z)^2 \mathrm{d} \mu &= \int_{\sR} g_{\sigma_*}(z)^2 \frac{1}{\sqrt{2 \pi}\norm{\vbeta^*}_2} e^{-\frac{z^2}{2 \norm{\vbeta^*}_2^2}} \mathrm{d} z \\
    &= \int_{-\infty}^0 \frac{1}{\sqrt{2 \pi}\norm{\vbeta^*}_2} e^{-\frac{z^2}{2 \norm{\vbeta^*}_2^2}} \mathrm{d} z + \int_{0}^{+\infty}(z^2+2z+1)\frac{1}{\sqrt{2 \pi}\norm{\vbeta^*}_2} e^{-\frac{z^2}{2 \norm{\vbeta^*}_2^2}} \mathrm{d} z\\
    &=1+\frac{1}{2}\norm{\vbeta^*}^2_2 +\int_{-\infty}^{+\infty}|z|\frac{1}{\sqrt{2 \pi}\norm{\vbeta^*}_2} e^{-\frac{z^2}{2 \|\beta\|_2^2}} d z \\
    &=1 + \frac{1}{2}\norm{\vbeta^*}_2^2 + \sqrt{\frac{2}{\pi}}\norm{\vbeta^*}_2 < \infty.
\end{split}
\end{align*}
Since $\vbeta^* \in \sS^{d-1}$, we get: $0 < \norm{F^*}_{L^2}^2 < 1 + \frac{1}{2}\norm{\vbeta^*}_2^2 + \sqrt{\frac{2}{\pi}}\norm{\vbeta^*}_2$, and $\norm{F^*}_{L^2}=\Theta_d(1)$.  \hfill 

\subsection{Norms of One-Step Update Matrix}

We begin by formulating the full-batch gradient for the first step ($\mG_0$) as follows:

\begin{align}
\begin{split}    
    \mG_0 &= \frac{1}{n\sqrt{d}} \left[  \frac{1}{\sqrt{h}}\left(\va\vy^\top -\frac{1}{\sqrt{h}} \va\va^\top\sigma\left(\frac{1}{\sqrt{d}}\mW_0\mX^\top\right) \right) \odot \sigma'\left(\frac{1}{\sqrt{d}}\mW_0\mX^\top\right) \right]\mX \\
    \mG_0 &= \underbrace{\frac{1}{n} \cdot \frac{\mu_1}{\sqrt{hd}} \va\vy^\top\mX }_{\mA} + \frac{1}{n} \cdot \frac{1}{\sqrt{hd}} \left(\va\vy^{\top} \odot \sigma_{\perp}^{\prime}\left(\frac{1}{\sqrt{d}}\mW_0\mX^\top\right)\right)\mX \\
     &\hspace{20pt}-\frac{1}{n} \cdot \frac{1}{h\sqrt{d}} \left(\va\va^\top \sigma\left(\frac{1}{\sqrt{d}}\mW_0\mX^\top\right)^{\top} \odot \sigma^{\prime}\left(\frac{1}{\sqrt{d}}\mW_0\mX^\top\right)\right)\mX.
\end{split}
\end{align}
Here we utilized the orthogonal decomposition of the activation function: $\sigma^{\prime}(z)=\mu_1+\sigma_{\perp}^{\prime}(z)$ to the second equality. Due to Stein's lemma \citep{stein1981estimation}, we know that $\mathbb{E}[z \sigma(z)]=\mathbb{E}\left[\sigma^{\prime}(z)\right]=\mu_1$, and hence $\mathbb{E}\left[\sigma_{\perp}^{\prime}(z)\right]=0$ for $z \sim \mathcal{N}(0,1)$.
\begin{lemma}
\label{asyrank}
      (\citep{ba2022high}): Given Assumptions \ref{asmpt:gauss_init},\ref{asmpt:norm_act}, and  \ref{asmpt:ST}, let $\mG_0=\frac{1}{\eta }\left(\mW_0-\mW_1\right)$ and $\boldsymbol{A}:=\frac{1}{n} \cdot \frac{\mu_1}{\sqrt{hd}} \va\vy^\top\mX $. Then there exists a constant $c$, such that for sufficiently large $n$:
    \begin{equation}
        \mathbb{P}\left(\left\|\mG_0-\boldsymbol{A}\right\|_2 \leq \frac{2 \log ^2 n}{\sqrt{n}}\left\|\boldsymbol{G}_0\right\|_2\right) \geq 1-n e^{-c \log ^2 n}-e^{-c n}.
    \end{equation}
\end{lemma}
This lemma implies that $\mG_0$ can be approximated by a rank one matrix $\mA$ under the operator norm. Now, to analyze the \texttt{FB-Adam} update, recall from equation \ref{adam_one_step_update} that $ \widetilde{\mG}_0 = \frac{1-\beta_1}{\sqrt{1 - \beta_2}} \operatorname{sign}(\mG_0).$
Observe that the essence of the first step \texttt{FB-Adam} update lies in the sign matrix $\operatorname{sign}(\mG_0)$. Based on the expansion of $\mG_0$, we leverage the rank-1 approximation matrix $\mA$ to state the following lemma.
\begin{lemma}
\label{lemma:sign_A_sign_G0}
     Given $\mG_0=\frac{1}{\eta }\left(\mW_1-\mW_0\right)$ and rank-1 matrix $\boldsymbol{A}:=\frac{\mu_1}{n\sqrt{hd}} \va\vy^\top\mX $, then for sufficiently large $n$:
     \begin{equation}
         \|\operatorname{sign}(\boldsymbol{A})-\operatorname{sign}(\mG_0)\|_2=0, \hspace{10pt} \operatorname{sign}(\boldsymbol{A})=\operatorname{sign}(\mG_0) \hspace{10pt} \text{with high probability}
     \end{equation} 
\end{lemma}
\paragraph{Proof of Lemma \ref{lemma:sign_A_sign_G0}.}
Let $a_{min}=\min_{i>0,j>0}|[\boldsymbol{A}]_{ij}|$. Since our analysis is based on large (fixed) $h$, from Lemma \ref{asyrank}, we almost surely have:
\begin{align}
\begin{split}   &\forall \delta>0, \exists k>0 , \forall n>k,\left\|\mG_0-\boldsymbol{A}\right\|_2< \delta \\
\implies &\forall \delta >0, \exists k>0 , \forall n>k,\left\|\mG_0-\boldsymbol{A}\right\|_F\leq\min{\{\sqrt{h},\sqrt{d}\}} \left\|\mG_0-\boldsymbol{A}\right\|_2\leq\min{\{\sqrt{h},\sqrt{d}\}}\delta
\end{split}
\end{align}
Considering $\delta=\frac{\sqrt{a_{min}}}{\sqrt{2}\cdot\min{\{\sqrt{h},\sqrt{d}\}}}$ gives us:
\begin{equation}
    |[\mG_{0}]_{ij}-[\boldsymbol{A}]_{ij}|^2 <\left\|\mG_0-\boldsymbol{A}\right\|_F^2 \leq \frac{a_{min}}{2},
\end{equation}
which implies $\exists k>0$, such that $\forall n>k$:
\begin{equation}
     \operatorname{sign}([\mG_{0}]_{ij})=\operatorname{sign}([\boldsymbol{A}]_{ij})
\end{equation}
Thus, $\|\operatorname{sign}(\boldsymbol{A})-\operatorname{sign}(\mG_0)\|_2=0$, $\operatorname{sign}(\boldsymbol{A})=\operatorname{sign}(\mG_0)$.  \hfill 
Next, we show that every entry of the matrix $\mA$ is not exactly $0$ almost surely.

\begin{proposition}
    \label{prop:A_nonzero}
    Let $\boldsymbol{A}:=\frac{1}{n} \cdot \frac{\mu_1}{\sqrt{hd}} \va\vy^\top\mX \in \sR^{h \times d}$, then $A_{i,j} \ne 0, \forall i \in [h], j \in [d]$ almost surely. 
\end{proposition}

\begin{definition}
    Given two measurable spaces $(\Omega,\mathcal{M},\mathbb{\mu})$, $(\Omega,\mathcal{M},\mathbb{\nu})$, we say $\mathbb{\nu}$ is absolutely continuous with respect to $\mathbb{\mu}$ if and only if  \begin{align*}
        \mathbb{\mu}(\mB)=0 \Rightarrow \mathbb{\nu}(\mB)=0, \hspace{10pt} \forall \mB \in \mathcal{M}.
    \end{align*} 
    We denote it as $\mathbb{\nu} \ll \mathbb{\mu}$.
\end{definition}

\begin{lemma}
    \label{lemma:abc}
    \citep{moran1968introduction} Given two measurable space $(\sR^n,\mathcal{B}(\sR^n),\mathbf{m}^n), $ $(\sR^n,\mathcal{B}(\sR^n),\mathcal{L}^n)$, where $\mathbf{m}^n$ is the Gaussian measure, $\mathcal{L}^n$ is the Lebesgue measure, we have $\mathbf{m}^n \ll \mathcal{L}^n$.

\end{lemma}
By {\em{Radon-Nikodym Theorem}} \citep{moran1968introduction}, we can define the Gaussian measure using the Lebesgue integral:
\begin{equation}
    \mathcal{L}^n(E)=\int_{E} \mathrm{d} \mathcal{L}^n; \; \mathbf{m}^n(E)=\int_{E} \mathrm{d} \mathbf{m}^n=\int_{E}\varPhi \mathrm{d}  \mathcal{L}^n, \hspace{10pt} \forall E \in \mathcal{B}(\sR^n),
\end{equation}
where $\varPhi$ is probability density function of $\mathbf{m}^n$ respect to $\mathcal{L}^n$. In the problem we consider, there are two groups of Gaussian measures: $a_i$ and $(X_{11},X_{12},...,X_{nd},\xi_1,...,\xi_n)$, which induce $(\sR,\mathbb{B}(\sR),\mathbf{m})$ and $(\sR^{nd+n},\mathbb{B}(\sR^{nd+n}),\mathbf{m}^{nd+n}) $ respectively. Here $a_i $ is the $i_{th}$ element of  $\va$; $X_{ij}$ is the element of $\mX$; $\xi_i$ is the Gaussian noise random variable.
\paragraph{Proof of Proposition ~\ref{prop:A_nonzero}.} 
Consider $\boldsymbol{A}:=\frac{1}{n} \cdot \frac{\mu_1}{\sqrt{hd}} \va\vy^\top\mX \in \sR^{h \times d}$  ,  if we can prove the following: 
\begin{equation*}
    \forall i \in [h],  j \in [d], \mathbb{P}(A_{i,j}=0)=0,
\end{equation*}
we can further have $\mathbb {P}(\exists i,j, \text{s.t } A_{i,j}=0)\leq \sum_{i=1}^h \sum_{j=1}^d \mathbb {P}( A_{i,j}=0)=0$, which means  $A_{i,j} \ne 0, \forall i \in [h], j \in [d]$ almost surely. So our goal is to prove  $\forall i \in [h],  j \in [d], \mathbb{P}(A_{i,j}=0)=0$. Given $\va \in \sR^{h\times1}, \vy^\top \mX \in \sR^{1\times d}$, notice that 
\begin{equation*}
    \{ A_{i,j}=0 \} \Leftrightarrow  \{ a_{i}=0 \} \bigcup \{(y^\top X)_j=0 \}.
\end{equation*}
Since $\va$ and $\vy^\top\mX$ are independent, we aim to prove $\forall i \in [h],  j \in [d]$, $\mathbf{m}^{nd+n}((y^\top X)_j=0)=0$ and $\mathbf{m} (a_{i}=0)=0 $.

\textbf{We first show $\forall  j \in [d], \mathbb{P}((y^\top X)_j=0)=0$}. Consider $y=\sigma_{*}(\mX \vbeta^*)+\xi$, where $\sigma_*$ is a \texttt{Softplus} function and $\vbeta^*=(b_1,...,b_d)^\top$ to get:
\begin{equation}
    (y^\top X)_j=\sum_{i=1}^n X_{ij} \left[ \ln \left(\exp\left(\sum_{k=1}^d b_k X_{ik}\right)+1\right)+\xi_i\right]
\end{equation}
It is easy to observe that $(y^\top X)_j$ can be written as a function:
\begin{equation}
    (y^\top X)_j=f_j(X_{11},\cdots,X_{1d},\cdots,X_{nd},\xi_1,\cdots, \xi_n).
\end{equation}
We can easily verify  $f_j: \sR^{nd+n} \rightarrow \sR$ is continuously differentiable of the first order,  and we denote it as $f_j \in \mathbf{C}^1 ( \sR^{nd+n})$. Considering the set  
\begin{align}
\begin{split}
    \mathcal{M}_1 &= \left\{  (X_{11},\cdots,X_{1d},\cdots,X_{nd},\xi_1,\cdots, \xi_n) \in \sR^{nd+n} \right.\\
   &\left. \hspace{20pt}| f_j(X_{11},\cdots,X_{1d},\cdots,X_{nd},\xi_1,\cdots, \xi_n)=0\right\},
\end{split}
\end{align}
due to   $f_j \in \mathbf{C}^1 ( \sR^{nd+n})$  and rank$(\mathbf{D}f_j)=$ rank$(\frac{\partial f_j}{\partial X_{11} },\cdots, \frac{\partial f_j}{\partial X_{1d} },\cdots,\frac{\partial f_j}{\partial X_{nd} },\frac{\partial f_j}{\partial \xi_1 },\cdots,\frac{\partial f_j}{\partial \xi_n })=1$, by {\em {Implicit Function Theorem}} \citep{zorich2016mathematical},  we have $\mathcal{M}_1$ is a  $C^1$  $(nd+n-1)$-dim sub-manifold. Therefore  $\mathcal{L}^{nd+n}(\mathcal{M}_1)=\int{\mathcal{M}_1}\mathrm{d} \mathcal{L}^{nd+n}=0$. 

$\bullet$ By Lemma ~\ref{lemma:abc},  $\mathbf{m}^{nd+n} \ll\mathcal{L}^{nd+n}$. Then $\mathbf{m}^{nd+n}(\mathcal{M}_1)=0$, we get  $\forall j \in [d] , \mathbf{m}^{nd+n}((y^\top X)_j=0)=0.$

$\bullet$ Observe that since any single point set is a zero-measure set for Lebesgue measure $\mathcal{L}^1$ and $\mathbf{m}\ll \mathcal{L}^1 $, we get $\forall i \in [h], \mathbf{m} (a_{i}=0)=0 $.

Since  we have proved $\forall i \in [h],  j \in [d],$ $\mathbf{m}^{nd+n} ((y^\top X)_j=0)=0$ and $\mathbf{m} (a_{i}=0)=0 $, we get $\forall i \in [h],  j \in [d], \mathbb{P}(A_{i,j}=0)=0$. Thus proving the proposition.  \hfill 

\par

\begin{lemma}
\label{lemma:forster}
    \citep{forster2001relations} Let $\mM \in\{-1,+1\}^{h \times d}$ and $\mM^{\prime} \in \mathbb{R}^{h \times d}$ such that $\operatorname{sign}\left(\mM_{i, j}\right)=\operatorname{sign}\left(\mM_{i, j}^{\prime}\right)$ for all $i \in [h], j \in [d]$. Then the following holds:
    \begin{equation}
        \operatorname{rank}\left(\mM^{\prime}\right) \geq \frac{\sqrt{hd}}{\norm{\mM}_2} .
    \end{equation}
\end{lemma}

From Lemma \ref{lemma:sign_A_sign_G0} and Proposition \ref{prop:A_nonzero} it is clear that $\operatorname{sign}(\mG_0)$ almost surely only contains $\{-1,1\}$. Now, by combining Lemma \ref{lemma:forster} and Lemma \ref{asyrank} for sufficiently large $n$, almost surely leads to:
\begin{align}
\begin{split}  \operatorname{rank}(\boldsymbol{A})=1 \geq \frac{\sqrt{hd}}{\|\operatorname{sign}(\boldsymbol{A})\|_2} 
\implies 1 \geq \frac{\sqrt{hd}}{\|\operatorname{sign}(\mG_0)\|_2}
\end{split}
\end{align}
Therefore, we have:
\begin{equation}
\label{eq:fb_adam_update_bigOmega}
    \norm{\widetilde{\mG}_0}_2 = \Omega_{d, \mathbb{P}}(\sqrt{hd})
\end{equation}
Additionally:
\begin{align}
\begin{split}
\label{eq:fb_adam_update_bigO}
     \norm{\widetilde{\mG}_0}_2 \leq \norm{\widetilde{\mG}_0}_F= \norm{\frac{1-\beta_1}{\sqrt{1 - \beta_2}} \operatorname{sign}(\mG_0)}_F = \frac{1-\beta_1}{\sqrt{1 - \beta_2}} \sqrt{hd}
     \implies \norm{\widetilde{\mG}_0}_2=O(\sqrt{hd}).
\end{split}
\end{align}

Finally, by combined equations \ref{eq:fb_adam_update_bigO} and  \ref{eq:fb_adam_update_bigOmega}, we  get: 
\begin{equation}
 \norm{\widetilde{\mG}_0}_2 = \Theta_{d, \mathbb{P}}(\sqrt{hd}), \hspace{10pt} \norm{\widetilde{\mG}_0}_F = \Theta_{d, \mathbb{P}}(\sqrt{hd}).
\end{equation}
Thus proving the theorem. \hfill 

\subsection{Numerical Simulations} 
\begin{figure}[h!]
     \centering
     \begin{subfigure}[b]{0.30\textwidth}
         \centering
         \includegraphics[width=\textwidth]{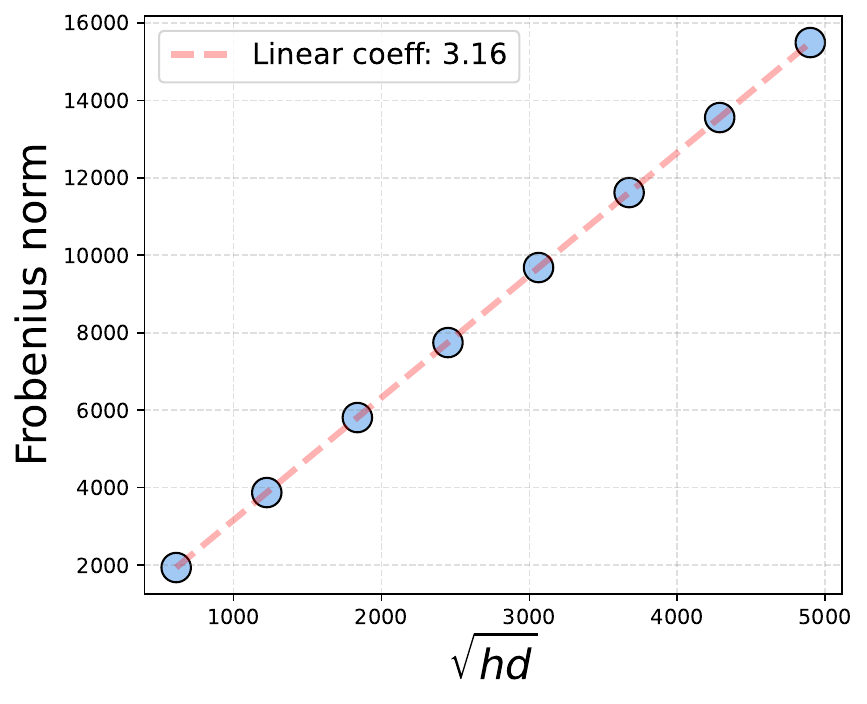}
         \caption{Frobenius norm}
     \end{subfigure}
     \begin{subfigure}[b]{0.30\textwidth}
         \centering
         \includegraphics[width=\textwidth]{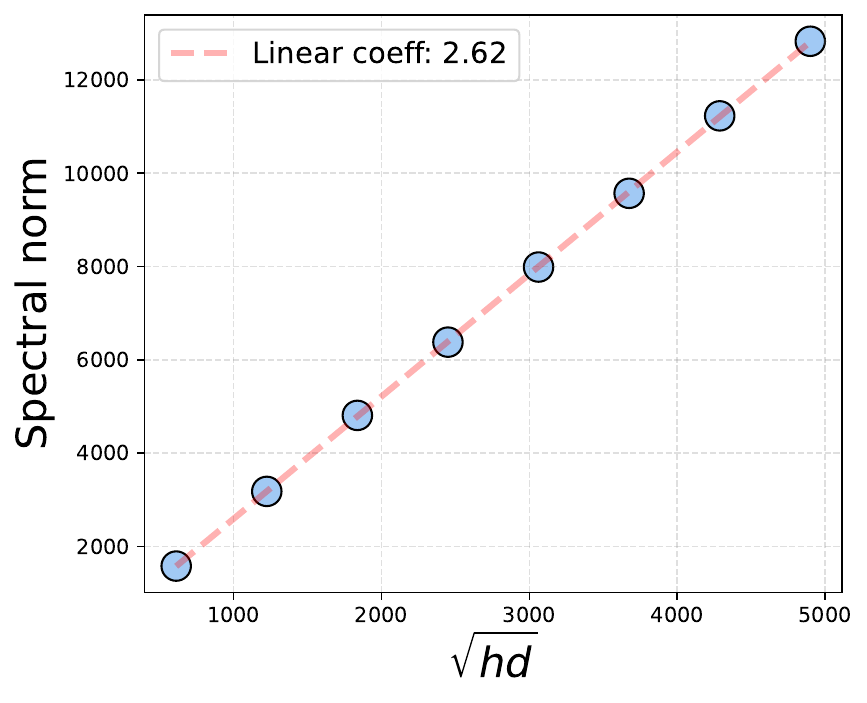}
         \caption{Spectral norm}
     \end{subfigure}
        \caption{Plots of $\norm{\widetilde{\mG}_0}_F, \norm{\widetilde{\mG_0}}_2$ with varying $n, d, h$ and $\beta_1=0.9, \beta_2=0.999$. }
        \label{fig:simulation_for_proof}
\end{figure}

We consider multiple sets of $n,h,d$ (see Table \ref{table_param_proof}) for one-step \texttt{FB-Adam} and plot the Frobenius norm and spectral norm of $\widetilde{\mG}_0$. Figure \ref{fig:simulation_for_proof} shows a linear relationship of the norms with $\sqrt{hd}$, which validates the results in our theorem:  $\left\|\widetilde{\mG}_0\right\|_2=\Theta_{d, \mathbb{P}}(\sqrt{hd}),\hspace{10pt} \left\|\widetilde{\mG}_0\right\|_F=\Theta_{d, \mathbb{P}}(\sqrt{hd})$.

\begin{table*}[h!]

\centering
{\scriptsize
\setlength\tabcolsep{20pt}
\begin{tabular}{ccccccc}
\hline
\gb \multicolumn{1}{c}{Index} & \multicolumn{1}{c}{$n$} & \multicolumn{1}{c}{$h$} & \multicolumn{1}{c}{$d$} & \multicolumn{1}{c}{Optimizer}  \\
\hline
\hline
    0 & 1000 & 750 & 500  &  \texttt{FB-Adam}   \\
\gb 1 & 2000 & 1500 & 1000  &  \texttt{FB-Adam}   \\
    2 & 3000 & 2250 & 1500  &  \texttt{FB-Adam}   \\
\gb 3 & 4000 & 3000 & 2000  &  \texttt{FB-Adam}   \\
    4 & 5000 & 3750 & 2500  &  \texttt{FB-Adam}   \\
\gb 5 & 6000 & 4500 & 3000  &  \texttt{FB-Adam}   \\
    6 & 7000 & 5250 & 3500  &  \texttt{FB-Adam}   \\
\gb 7 & 8000 & 6000 & 4000  &  \texttt{FB-Adam}   \\
\bottomrule

\end{tabular}
}
\caption{Parameters for Figure \ref{fig:simulation_for_proof}}
\label{table_param_proof}
\end{table*}


\section{Singular Vector Alignments of Weights and Optimizer Updates}
\label{subsec:singular_vector_alignments}

In this section, we present a singular-vector perspective on the emergence of HT ESD after multiple update steps.
By considering an `update' matrix $\mM_t \in \sR^{h \times d}$ , we formulate the weight updates as:
\begin{align}
    \mW_{t+1} = \mW_t + \mM_t,
\end{align} 
where $\mM_t$ is the optimizer update matrix based on \texttt{GD} ($\mM_t = -\eta \mG_t)$ or \texttt{FB-Adam} ($\mM_t = -\eta \widetilde{\mG}_t)$. 
We leverage methods from the rich literature on signal recovery in spiked matrix models \citep{ shabalin2013reconstruction,landau2023singular, el2018detection, gavish2017optimal, troiani2022optimal} to analyze the role of singular vectors of $\mM_t$ in transforming the ESD of $\mW_t^\top\mW_t$ gradually into a HT distribution. Let $b=\min(h, d)$, we consider the SVD of $\mW_{t+1}, \mW_t, \mM_t$ as:
\begin{align}
    \mW_{t+1} = \mU_{W_{t+1}}\mS_{W_{t+1}}\mV_{W_{t+1}}^\top ,
    \hspace{10pt} \mW_{t} = \mU_{W_{t}}\mS_{W_{t}}\mV_{W_{t}}^\top , 
    \hspace{10pt} \mM_{t} = \mU_{M_{t}}\mS_{M_{t}}\mV_{M_{t}}^\top,
\end{align}
where $\mU_{W_{t+1}}, \mU_{W_{t}}, \mU_{M_{t}} \in \sR^{h \times b}$, $\mS_{W_{t+1}}, \mS_{W_{t}}, \mS_{M_{t}} \in \sR^{b \times b}$ and $\mV_{W_{t+1}}, \mV_{W_{t}}, \mV_{M_{t}} \in \sR^{b \times d}$.



\begin{definition}
    \citep{landau2023singular}. The `overlaps' between two singular vector matrices $\mJ, \mQ \in \sR^{a \times b}$ is defined as:
    $\gO(\mJ, \mQ) = (\mJ^\top \mQ)^{\circ 2} \in \sR^{b \times b}$
\end{definition}

\paragraph{Overlaps during training.} From the finite rank spiked matrix model, we know that the singular values $\mS_{W_1}$ are non-linear transformations (also termed as `inflations' \citep{landau2023singular}) of $\mS_{M_0}$, and $\mU_{W_1}, \mV_{W_1}$ are rotated variants of $\mU_{M_0}, \mV_{M_0}$ respectively. Formally, let $\hat{s}_1 \ge \hat{s}_2 \cdots \ge \hat{s}_b$ denote the singular values of $\mW_1$, and let $s_1 \ge s_2 \cdots \ge s_b$ denote the singular values of $\mM_0$. Let $\hat{\vu}_j \in \sR^h, \hat{\vv}_j \in \sR^d$ represent the left and right singular vectors of $\mW_1$ corresponding to singular value $\hat{s}_j$. Similarly, let $\vu_k \in \sR^h, \vv_k \in \sR^d$ represent the left and right singular vectors of $\mM_0$ corresponding to singular value $s_k$. Owing to the rotational invariant nature of the Gaussian matrix $\mW_0$, the alignment values $\mathbb{E}\left[ (\hat{\vu}_j^\top \vu_k )^2 \right], \mathbb{E}\left[ (\hat{\vv}_j^\top \vv_k )^2 \right], \forall j, k \in \{1, \cdots, b\}$ can be computed solely based on $\hat{s}_j, s_k$ (see \citet{landau2023singular, mingo2017free}). 
In this one-step context, we show that the \emph{large} $\eta$ (from Section \ref{sec:one_step_fb_adam_update}) leads to outlier alignment values in the overlap plots.

\begin{figure}[t!]
     \centering
     \begin{subfigure}[b]{0.24\textwidth}
         \centering
         \includegraphics[width=\textwidth]{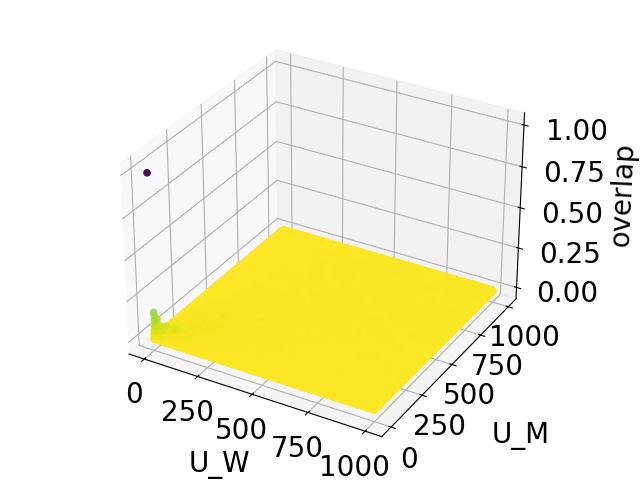}
         \caption{ $\gO(\mU_{W_1}, \mU_{M_{0}})$ } 
         \label{fig:main:overlap_fb_adam_lr1_u_o1}
     \end{subfigure}
     \hfill
     \begin{subfigure}[b]{0.24\textwidth}
         \centering
         \includegraphics[width=\textwidth]{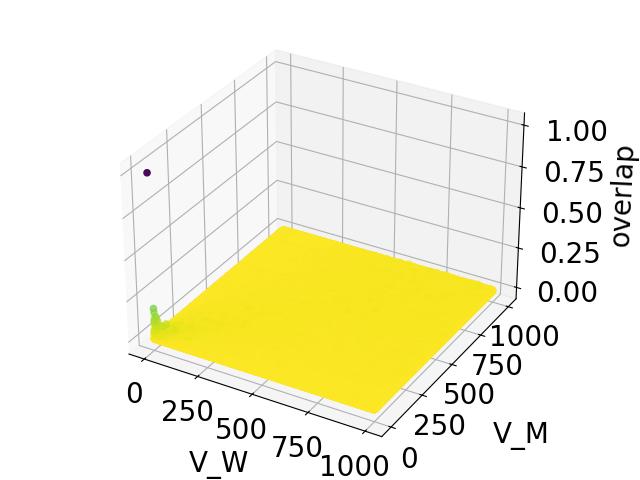}
         \caption{ $\gO(\mV_{W_1}, \mV_{M_{0}})$ } 
         \label{fig:main:overlap_fb_adam_lr1_v_o1}
     \end{subfigure}
     \begin{subfigure}[b]{0.25\textwidth}
         \centering
         \includegraphics[width=\textwidth]{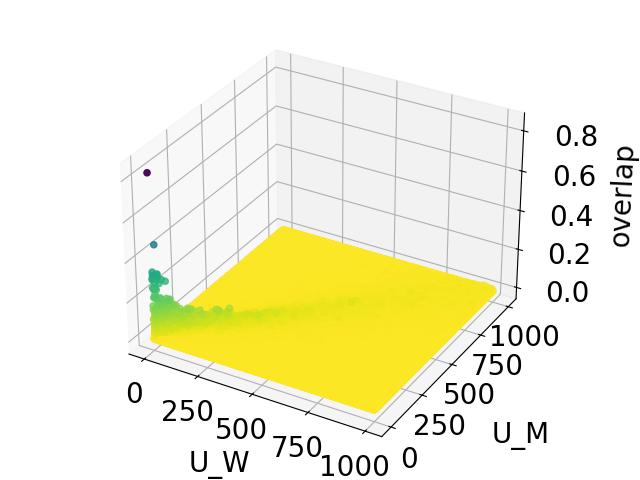}
         \caption{$\gO(\mU_{W_{10}}, \mU_{M_9})$ } 
         \label{fig:main:overlap_fb_adam_lr1_u_o10}
     \end{subfigure}
     \begin{subfigure}[b]{0.25\textwidth}
         \centering
         \includegraphics[width=\textwidth]{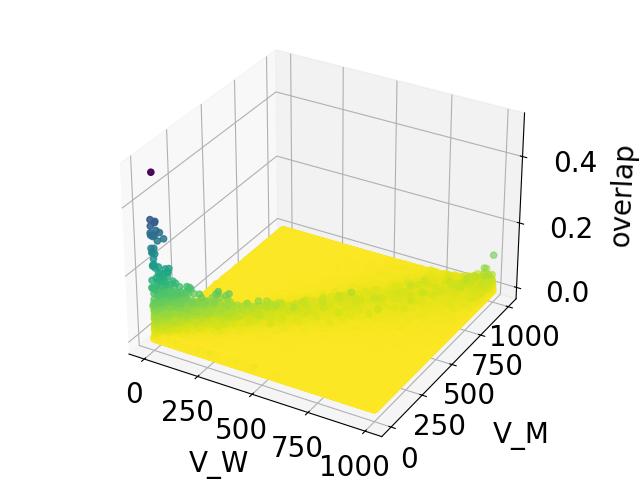}
         \caption{ $\gO(\mV_{W_{10}}, \mV_{M_9})$ } 
         \label{fig:main:overlap_fb_adam_lr1_v_o10}
     \end{subfigure}
        \caption{ Overlaps after one step (a), (b) and $10$ steps (c), (d) with \texttt{FB-Adam}($\eta=1$). }
        \label{fig:main:overlap_fb_adam_lr1}
\end{figure}

In particular, observe from Figure~\ref{fig:main:overlap_fb_adam_lr1_u_o1}, Figure~\ref{fig:main:overlap_fb_adam_lr1_v_o1} that $(\hat{\vu}_1^\top \vu_1 )^2, (\hat{\vv}_1^\top \vv_1 )^2$ (i.e. alignments of top singular vectors of $\mW_1, \mM_0$) have high values which are close to $1$ and correlate with the presence of a spike in the ESD (Table~\ref{tab:ADAM_transition}) with $\eta=1, t=1$. Furthermore, Figure~\ref{fig:main:overlap_fb_adam_lr1_u_o10}, Figure~\ref{fig:main:overlap_fb_adam_lr1_v_o10} show a decay of overlap values across the diagonal that tends to correlate with the HT ESD emergence (Table~\ref{tab:ADAM_transition}) with $\eta=1, t=10$.
Similar observations for \texttt{GD} are presented in Appendix~\ref{app:add_exp}. 

\section{Discussions on Mean-Field Initialization }
\label{app:mean_field_discussion}
Our setup employs the widely studied NTK initialization \citep{wang2023spectral} for the two-layer NNs. Alternatively, previous studies have also focused on mean-field-based initialization \citep{ba2022high, moniri2023theory} to analyze the role of one-step optimizer updates. The mean-field initialization for a two-layer NN can be formulated as:
\begin{align}
    \label{eq:mean_field_nn}
    f(\vx_i) &= \frac{1}{\sqrt{h}} \va^\top \sigma\left(\mW \vx_i \right).
\end{align}
Here $\mW \in \sR^{h \times d}, \va \in \sR^h$ are the first and second layer weights respectively, with entries sampled as $\left[\mW_0\right]_{i j} \stackrel{\text { i.i.d. }}{\sim} \mathcal{N}(0,\frac{1}{d})$, $\left[a\right]_{i} \stackrel{\text { i.i.d. }}{\sim} \mathcal{N}(0,\frac{1}{h}), \forall i \in [h], j \in [d]$. Notice the change in scale of the entries and the $\frac{1}{\sqrt{h}}$ scaling factor in equation \ref{eq:mean_field_nn}. In this setup:  $\norm{\mW_0}_2 = \Theta_{d, \sP}(1), \norm{\mW_0}_F = \Theta_{d, \sP}(\sqrt{h})$.

 This section provides additional results and discussions for the mean-field initialization. We claim that it is straightforward to extend the conclusions regarding the scale of $\eta$ for one-step \texttt{FB-Adam} to this setting. In particular, with \emph{large} $\eta$ and multiple optimizer steps, we observe the emergence of HT ESDs. Additionally, the alignments of singular vectors of the weight and corresponding optimizer update matrices also remain a potential contributor to the emergence of HT ESDs, consistent with the discussion in the main paper.

\subsection{Scaling $\eta$ and Alignment of Singular Vectors for \texttt{FB-Adam} }

\begin{figure}[h!]
     \centering
     \begin{subfigure}[b]{0.24\textwidth}
         \centering
         \includegraphics[width=\textwidth]{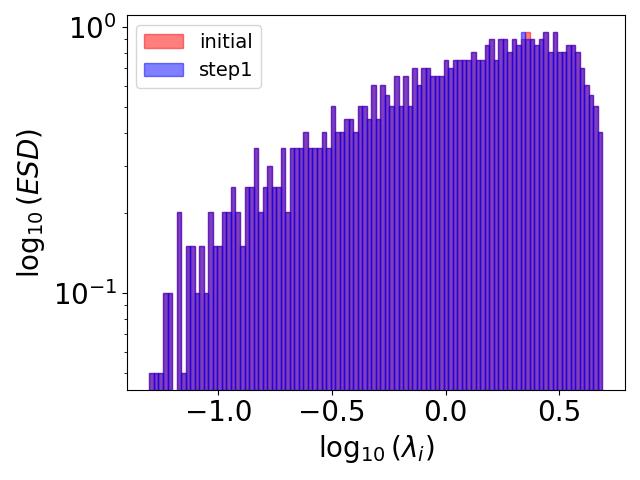}
         \caption{$\eta=0.1$}
     \end{subfigure}
     \hfill
     \begin{subfigure}[b]{0.24\textwidth}
         \centering
         \includegraphics[width=\textwidth]{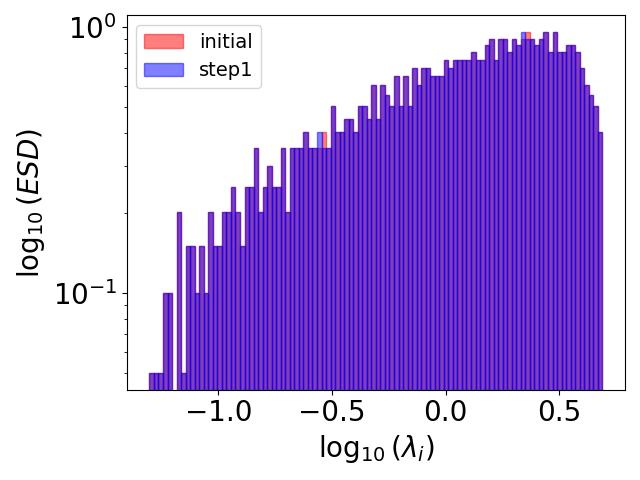}
         \caption{$\eta=1$}
     \end{subfigure}
     \hfill
     \begin{subfigure}[b]{0.24\textwidth}
         \centering
         \includegraphics[width=\textwidth]{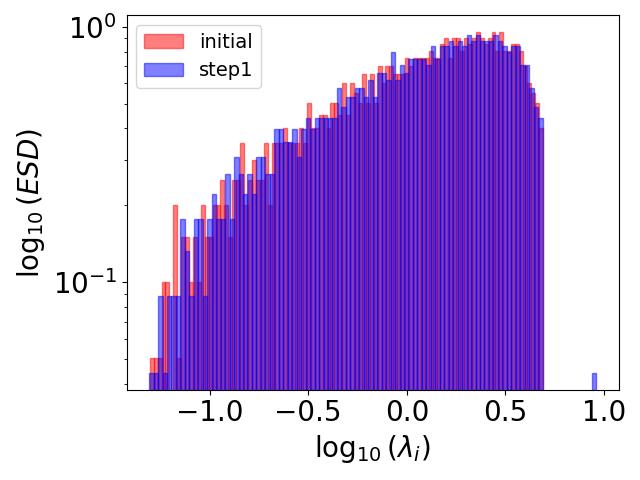}
         \caption{$\eta=100$}
     \end{subfigure}
     \hfill
     \begin{subfigure}[b]{0.24\textwidth}
         \centering
         \includegraphics[width=\textwidth]{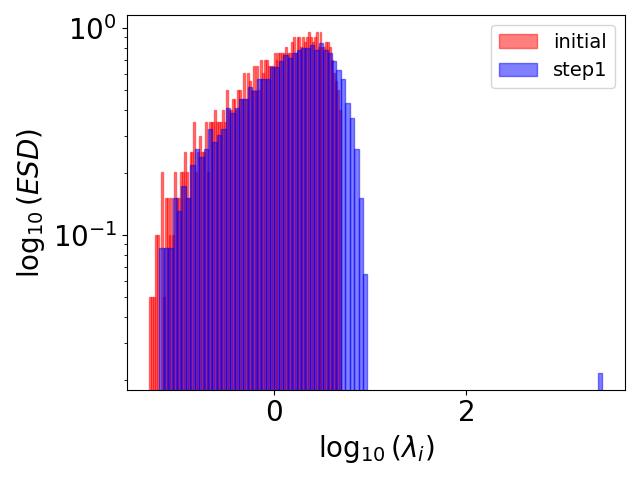}
         \caption{$\eta=2000$}
     \end{subfigure}
     
    \caption{Evolution of ESD of $\mW^\top\mW$ after one step \texttt{GD} optimizer update in mean-field setting. Here $n=2000$, $d=1000$, $h=1500$, $\sigma_* = \texttt{softplus}, \sigma = \texttt{tanh}, \rho_e = 0.3$.
}
\label{fig:appendix:gd_W_esd_one_step_mean_field}
\end{figure}

\textbf{A note on notation from \citet{ba2022high}:}
In our setup, we denote $\mG_0 = \frac{1}{\eta}(\mW_1 - \mW_0)$, whereas \citet{ba2022high} consider $\mG_0 = \frac{1}{\eta\sqrt{h}}(\mW_1 - \mW_0)$. Thus, in the mean-field setting, the learning rates we obtain are simply the scaled versions of theirs by a factor of $\sqrt{h}$. To this end, \citet{ba2022high} showed that $\eta=\Theta(h)$ (scaling adjusted to our notation) is (sufficiently) large for the \texttt{GD} update $\mG_0$ (see Figure ~\ref{fig:appendix:gd_W_esd_one_step_mean_field}). One can also verify that the results of Theorem ~\ref{thm:fb_adam_norms} for \texttt{FB-Adam} can be scaled and extended to this setting.

\begin{corollary}
\label{cor:mean_field_lr_scale}
    Under the assumptions of Theorem \ref{thm:fb_adam_norms}, we obtain the following scaling for $\eta$ in the mean-field initialization setting:
    \begin{align}
    \begin{split}
        \eta = \Theta(1/\sqrt{d}) \implies \norm{\mW_1 - \mW_0}_F \asymp \norm{W_0}_F \\
        \eta = \Theta\left(1/\sqrt{hd}\right) \implies \norm{\mW_1 - \mW_0}_2 \asymp \norm{W_0}_2,
    \end{split}
    \end{align}
    where $\norm{\mW_0}_2 = \Theta_{d, \sP}(\sqrt{1}), \norm{\mW_0}_F = \Theta_{d, \sP}(\sqrt{h})$.
\end{corollary}

\begin{figure}[h!]
     \centering
     \begin{subfigure}[b]{0.24\textwidth}
         \centering
         \includegraphics[width=\textwidth]{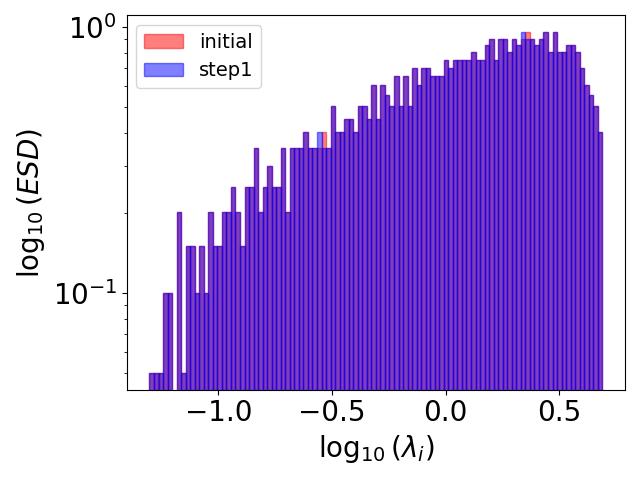}
         \caption{$\eta=0.00005$}
     \end{subfigure}
     \hfill
     \begin{subfigure}[b]{0.24\textwidth}
         \centering
         \includegraphics[width=\textwidth]{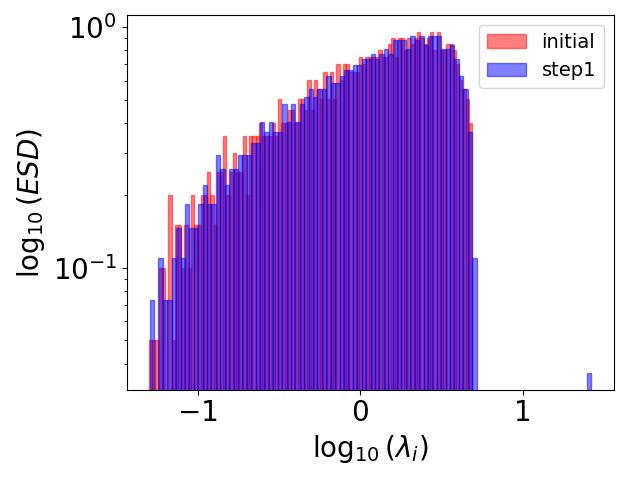}
         \caption{$\eta=0.005$}
     \end{subfigure}
     \hfill
     \begin{subfigure}[b]{0.24\textwidth}
         \centering
         \includegraphics[width=\textwidth]{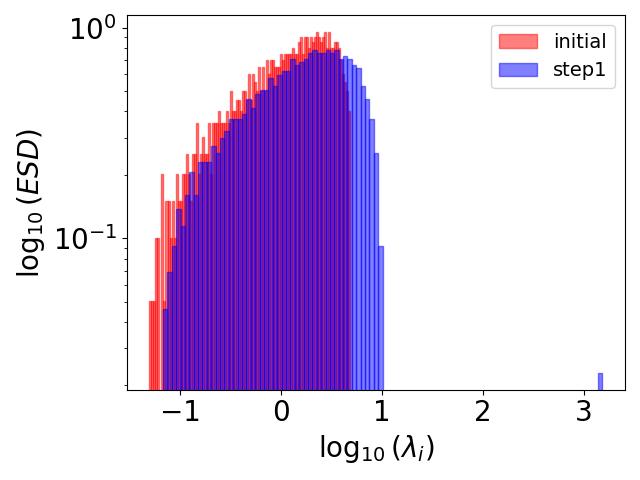}
         \caption{$\eta=0.04$}
         \label{fig:appendix:gd_fb_adam_W_esd_one_step_mean_field_lr0.04}
     \end{subfigure}
     \hfill
     \begin{subfigure}[b]{0.24\textwidth}
         \centering
         \includegraphics[width=\textwidth]{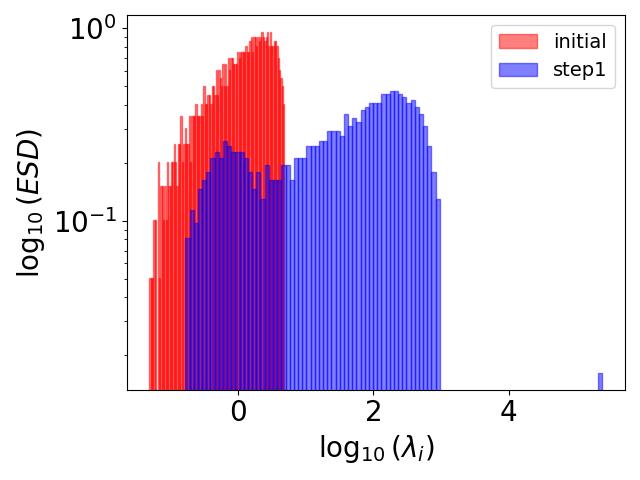}
         \caption{$\eta=0.5$}
     \end{subfigure}
     
    \caption{Evolution of ESD of $\mW^\top\mW$ after one step \texttt{FB-Adam} update in mean-field setting. Here $n=2000$, $d=1000$, $h=1500$, $\sigma_* = \texttt{softplus}, \sigma = \texttt{tanh}, \rho_e = 0.3$.
}
        \label{fig:appendix:gd_fb_adam_W_esd_one_step_mean_field}
\end{figure}

\begin{figure}[t!]
     \centering
     \begin{subfigure}[b]{0.24\textwidth}
         \centering
         \includegraphics[width=\textwidth]{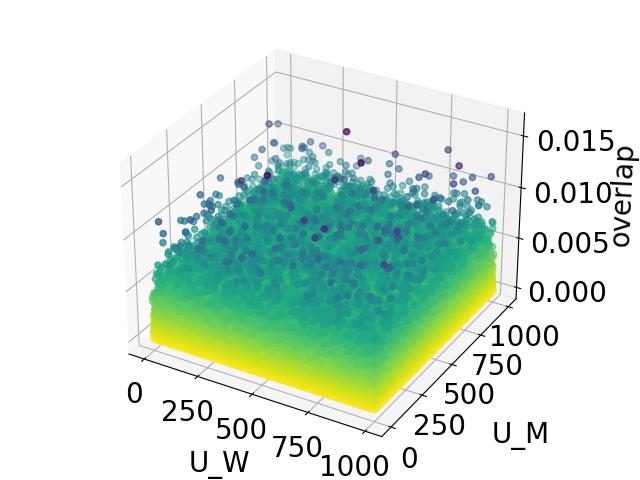}
         \caption{$\gO(\mU_{W_0}, \mU_{M_0})$}

     \end{subfigure}
     \hfill
     \begin{subfigure}[b]{0.24\textwidth}
         \centering
         \includegraphics[width=\textwidth]{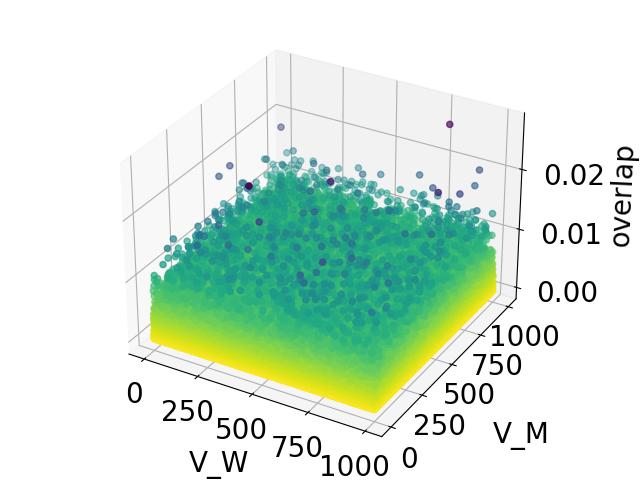}
         \caption{$\gO(\mV_{W_0}, \mV_{M_0})$}
     \end{subfigure}
     \hfill
     \begin{subfigure}[b]{0.24\textwidth}
         \centering
         \includegraphics[width=\textwidth]{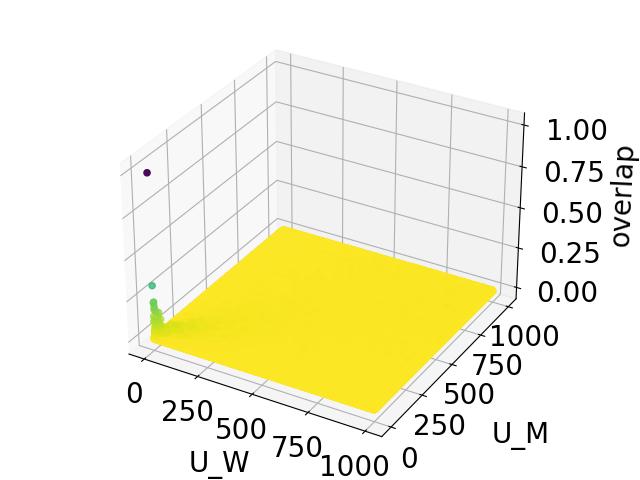 }
         \caption{$\gO(\mU_{W_1}, \mU_{M_0})$}
         
     \end{subfigure}
     \hfill
     \begin{subfigure}[b]{0.24\textwidth}
         \centering
         \includegraphics[width=\textwidth]{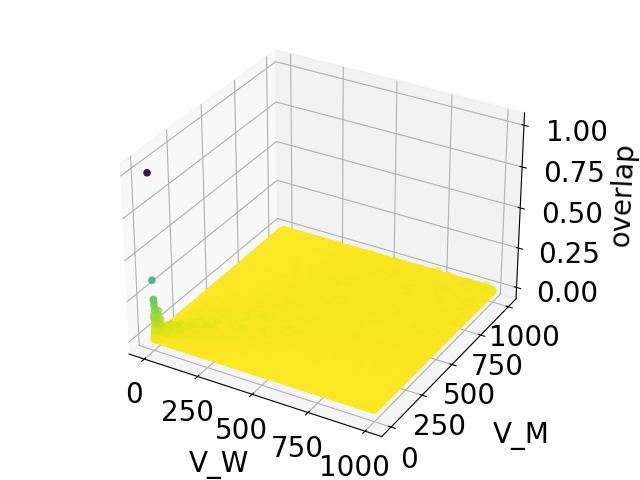}
         \caption{$\gO(\mV_{W_1}, \mV_{M_0})$}
     \end{subfigure}
     
    \caption{Overlaps after one \texttt{FB-Adam} update with $\eta=0.04$ in mean-field setting.}
        \label{fig:appendix:gd_fb_adam_overlap_one_step_mean_field}
\end{figure}

\paragraph{Singular Vector Overlaps after one-step \texttt{FB-Adam} update.} Following the main paper, we compute the following overlap metrics: $\gO(\mU_{W_0}, \mU_{M_0}), \gO(\mV_{W_0}, \mV_{M_0}), \gO(\mU_{W_{1}}, \mU_{M_0}), \gO(\mV_{W_{1}}, \mV_{M_0})$. For \texttt{FB-Adam} with $\eta=0.04$ (which is a large $\eta$ in this setting), Figure ~\ref{fig:appendix:gd_fb_adam_overlap_one_step_mean_field} shows the outliers for $\gO(\mU_{W_{1}}, \mU_{M_0}), \gO(\mV_{W_{1}}, \mV_{M_0})$ corresponding to the spike in Figure~\ref{fig:appendix:gd_fb_adam_W_esd_one_step_mean_field_lr0.04}. Thus aligning with the results in Section~\ref{subsec:singular_vector_alignments}.

\begin{figure}[t!]
     \centering
     \begin{subfigure}[b]{0.24\textwidth}
         \centering
         \includegraphics[width=\textwidth]{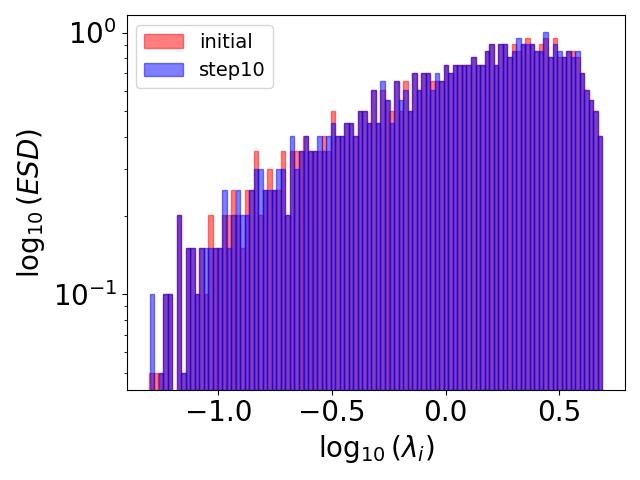}
         \caption{$\eta=0.00005$}
     \end{subfigure}
     \hfill
     \begin{subfigure}[b]{0.24\textwidth}
         \centering
         \includegraphics[width=\textwidth]{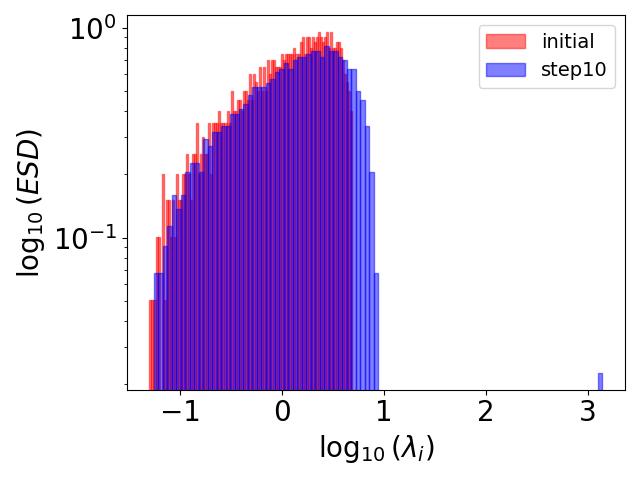}
         \caption{$\eta=0.005$}
     \end{subfigure}
     \hfill
     \begin{subfigure}[b]{0.24\textwidth}
         \centering
         \includegraphics[width=\textwidth]{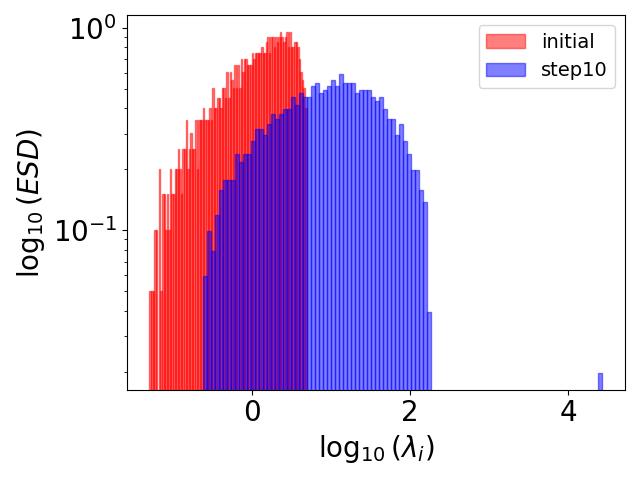}
         \caption{$\eta=0.04$}
     \end{subfigure}
     \hfill
     \begin{subfigure}[b]{0.24\textwidth}
         \centering
         \includegraphics[width=\textwidth]{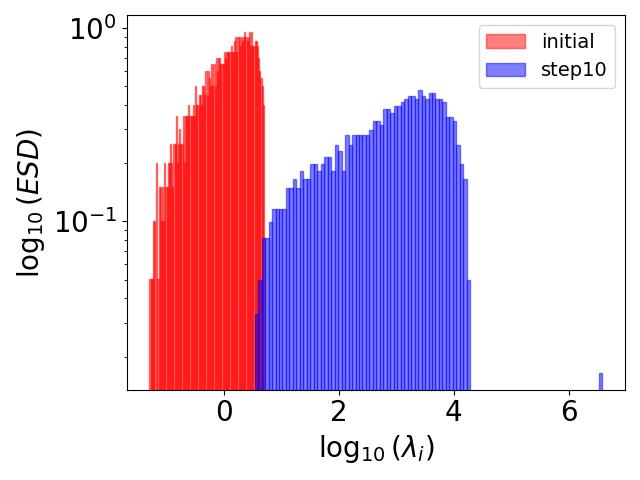}
         \caption{$\eta=0.5$}
     \end{subfigure}
     
    \caption{Evolution of ESD of $\mW^\top\mW$ after $10$ \texttt{FB-Adam} updates in mean-field setting. Here $n=2000$, $d=1000$, $h=1500$, $\sigma_* = \texttt{softplus}, \sigma = \texttt{tanh}, \rho_e = 0.3$.
}
\label{fig:appendix:gd_fb_adam_W_esd_ten_step_mean_field}
\vspace{-3mm}
\end{figure}

\begin{figure}[t!]
     \centering
     \begin{subfigure}[b]{0.24\textwidth}
         \centering
         \includegraphics[width=\textwidth]{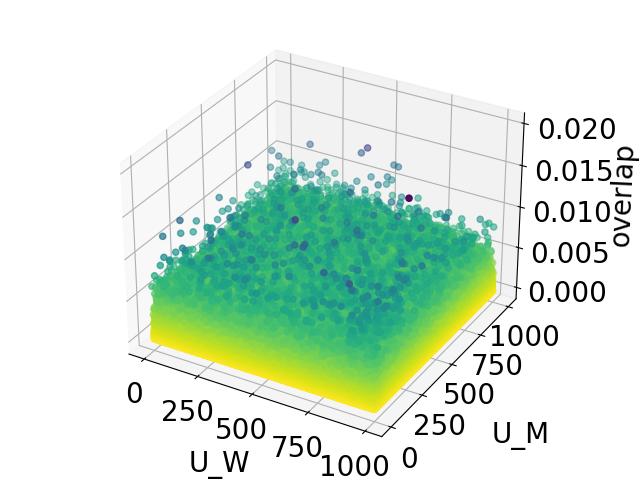}
         \caption{ $\eta=0.00005 $ \\ $ \gO(\mU_{W_{10}}, \mU_{M_9})$ } 
     \end{subfigure}
     \hfill
     \begin{subfigure}[b]{0.24\textwidth}
         \centering
         \includegraphics[width=\textwidth]{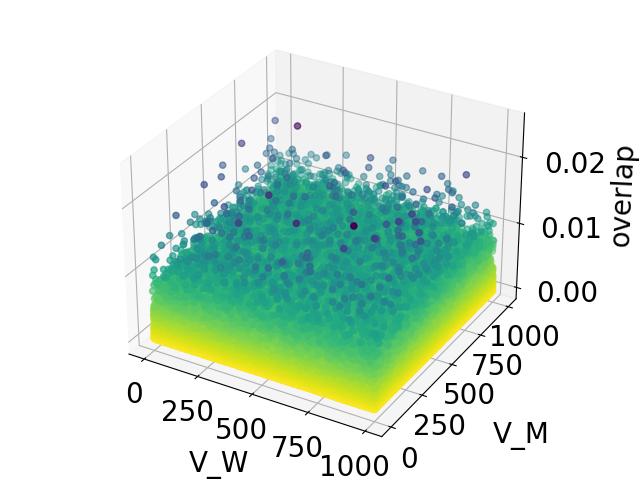}
         \caption{ $\eta=0.00005  $ \\ $ \gO(\mV_{W_{10}}, \mV_{M_9})$ } 
     \end{subfigure}
     \hfill
     \begin{subfigure}[b]{0.24\textwidth}
         \centering
         \includegraphics[width=\textwidth]{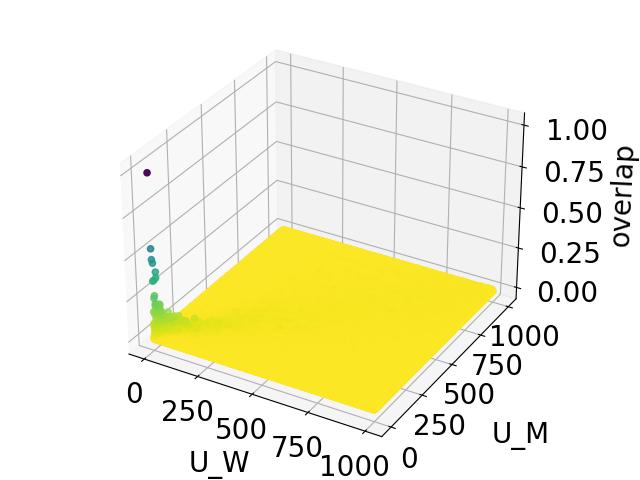}
         \caption{ $\eta=0.04   $ \\ $  \gO(\mU_{W_{10}}, \mU_{M_9})$ } 
     \end{subfigure}
     \hfill
     \begin{subfigure}[b]{0.24\textwidth}
         \centering
         \includegraphics[width=\textwidth]{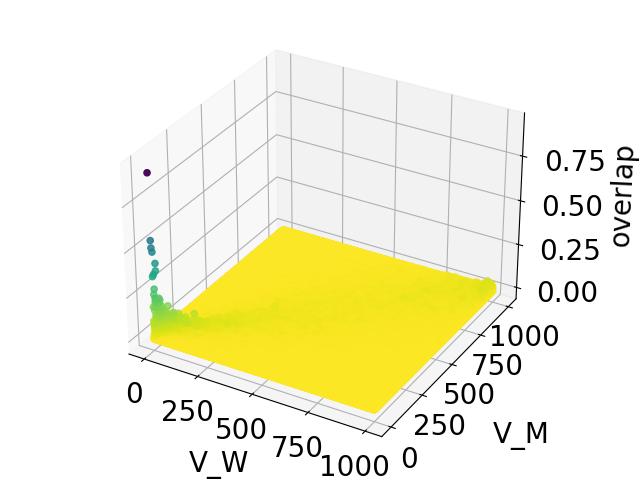}
         \caption{ $\eta=0.04  $ \\ $  \gO(\mV_{W_{10}}, \mV_{M_9})$ } 
     \end{subfigure}
        \caption{ Overlaps between singular vectors after $10$ \texttt{FB-Adam} updates with $\eta=0.00005$ (plots (a), (b)) and $\eta=0.04$  (plots (c), (d)) in the mean-field setting.}
        \label{fig:appendix:overlap_fb_adam_step10}
        \vspace{-3mm}
\end{figure}

\paragraph{Heavy-Tailed Phenomenon after Multiple Steps With Mean-Field Initialization.} Similar to the experiments in Section~\ref{sec:HT_phenomenon}, we employ the mean-field initialization and apply $10$ \texttt{FB-Adam} updates with various $\eta$ to compute $\gO(\mU_{W_{10}}, \mU_{M_9}), \gO(\mV_{W_{10}}, \mV_{M_9})$. Notice that for small $\eta=0.00005$, the ESD is expected to remain largely unchanged and the overlap plots illustrate random alignment values. However, for $\eta=0.04$, the outliers emerge in the overlap matrices in Figure~\ref{fig:appendix:overlap_fb_adam_step10} and correlate with the HT ESD in Figure~\ref{fig:appendix:gd_fb_adam_W_esd_ten_step_mean_field}. 

\paragraph{Note on the kernel lower bound.} Without loss of generality, consider a dataset $\mX$ for which we have a kernel bound for the test loss. \cite{ba2022high} have shown that the Bulk + Spike ESD after the first step with \texttt{GD} will result in a test loss that is much better than this kernel bound (For example, with $\eta=100$ in Figure~\ref{fig:appendix:gd_W_esd_one_step_mean_field}). Informally, our results imply that, by choosing $\eta$ for which a bulk + spike ESD occurs with FB-Adam, (with that same $\mX$), then the same feature learning behaviour (via increased KTA and $sim(W,\beta^*)$) represents that we are indeed better than the kernel bound. Nonetheless, we defer a rigorous characterization of this behaviour to future work (Appendix~\ref{app:sec:fw}).

\newpage
\section{ESD evolution over long training periods}
\label{app:sec:esd_evolution}

In the main text, we claimed and validated that the ESD of the hidden layer weight matrix exhibits heavy tails after multiple steps of \texttt{GD}/ \texttt{FB-Adam} with sufficiently large $\eta$. In the following ESD plots, we further justify our claim and illustrate that very long training $t=10000$ with small $\eta$ does not result in HT ESDs.

\subsection{ESD evolution with \texttt{GD}}

\begin{table}[h!]
\centering
\setlength{\tabcolsep}{1.5pt} 
\renewcommand{\arraystretch}{1.2}

\begin{tabular}{ |c| c|c|c|c|}
    \hline
    & $t = 1$ & $t = 10$ & $t = 100$ & $t = 10000$ \\ \hline

    \multirow{2}{*}{$\eta = 0.1$} & \adjustbox{valign=c}{\includegraphics[width=0.24\textwidth]{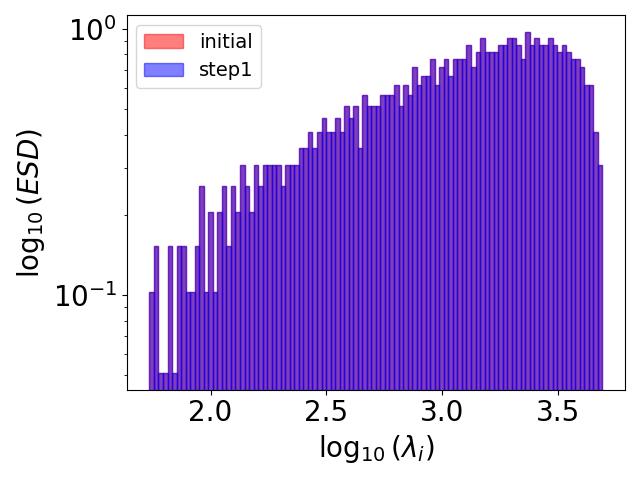}} & 
    \adjustbox{valign=c}{\includegraphics[width=0.24\textwidth]{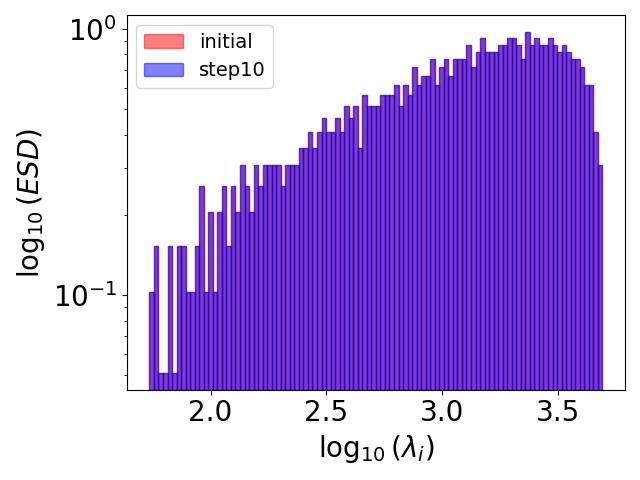}} & 
    \adjustbox{valign=c}{\includegraphics[width=0.24\textwidth]{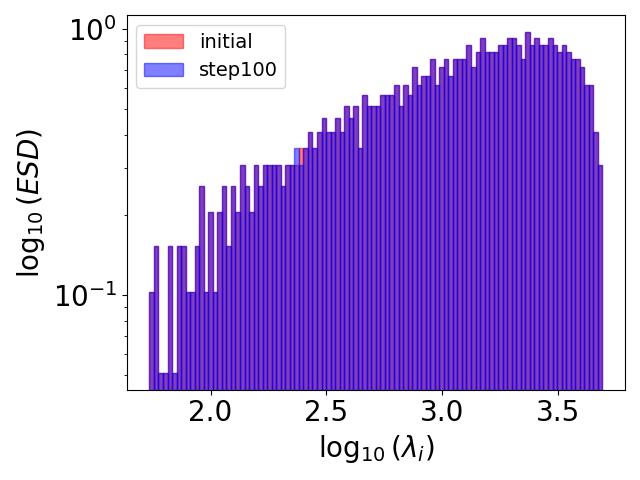}} & 
    \adjustbox{valign=c}{\includegraphics[width=0.24\textwidth]{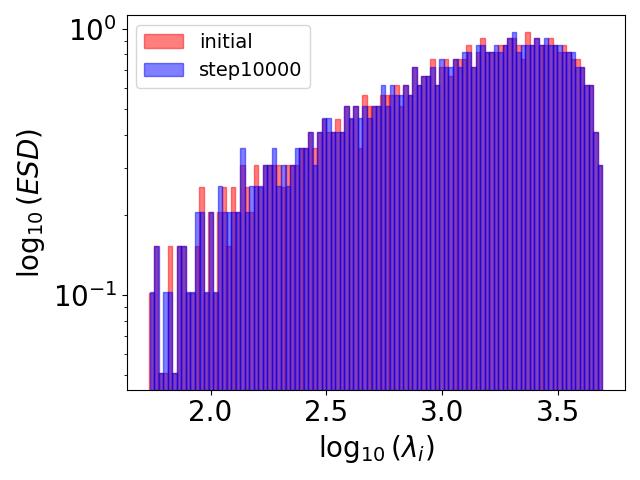}} \\ \hline

    \multirow{2}{*}{$\eta = 1$} & \adjustbox{valign=c}{\includegraphics[width=0.24\textwidth]{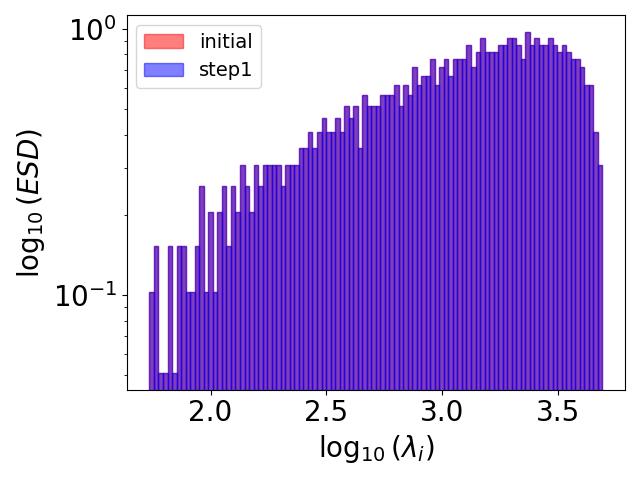}} & 
    \adjustbox{valign=c}{\includegraphics[width=0.24\textwidth]{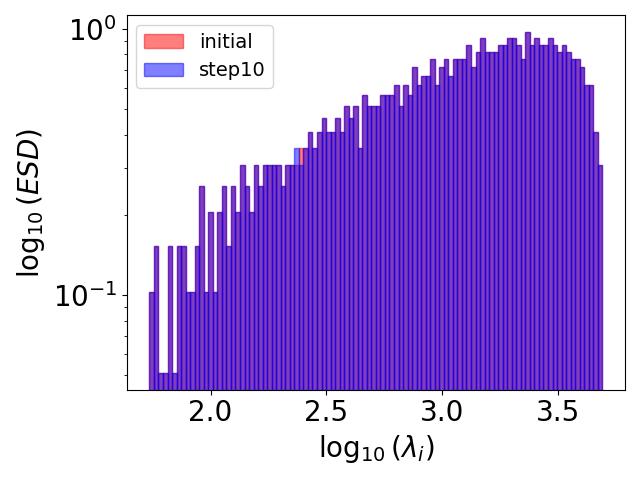}} & 
    \adjustbox{valign=c}{\includegraphics[width=0.24\textwidth]{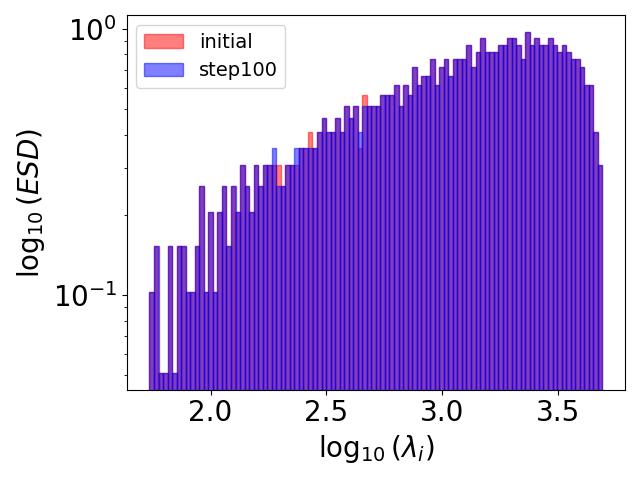}} & 
    \adjustbox{valign=c}{\includegraphics[width=0.24\textwidth]{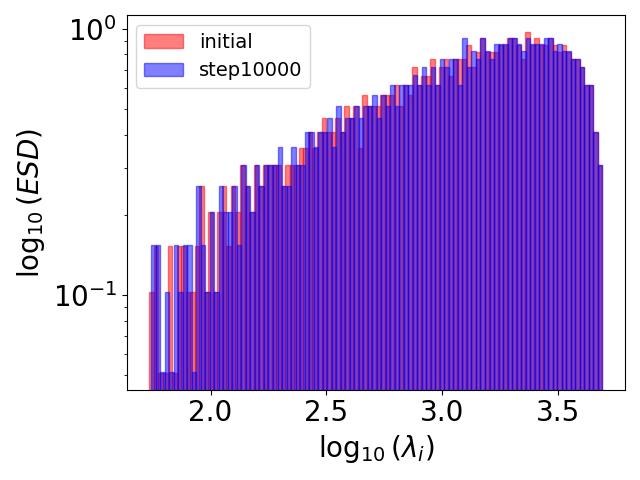}} \\ \hline

    \multirow{2}{*}{$\eta = 10$} & \adjustbox{valign=c}{\includegraphics[width=0.24\textwidth]{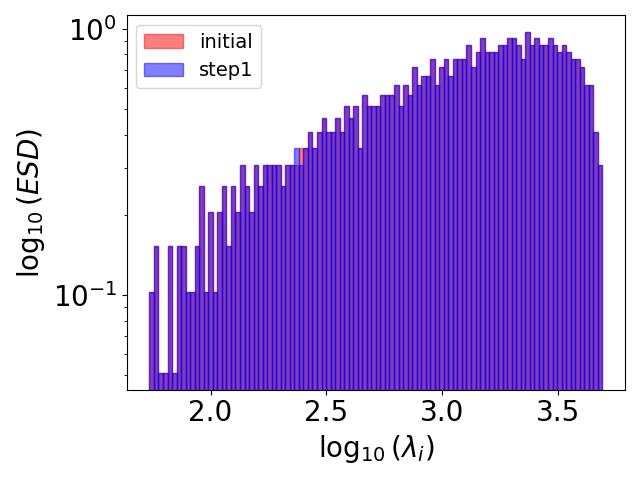}} & 
    \adjustbox{valign=c}{\includegraphics[width=0.24\textwidth]{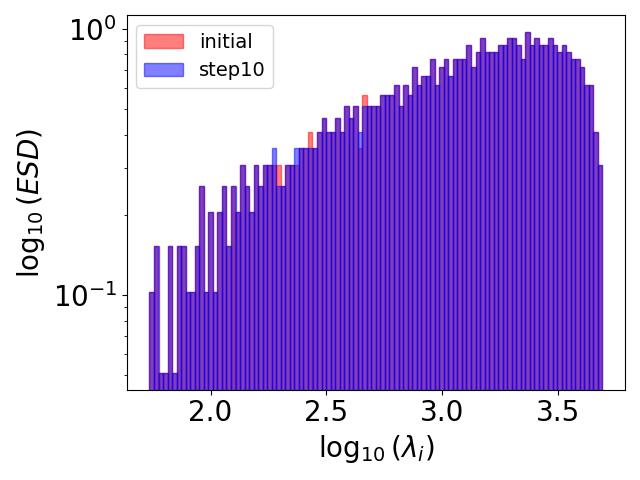}} & 
    \adjustbox{valign=c}{\includegraphics[width=0.24\textwidth]{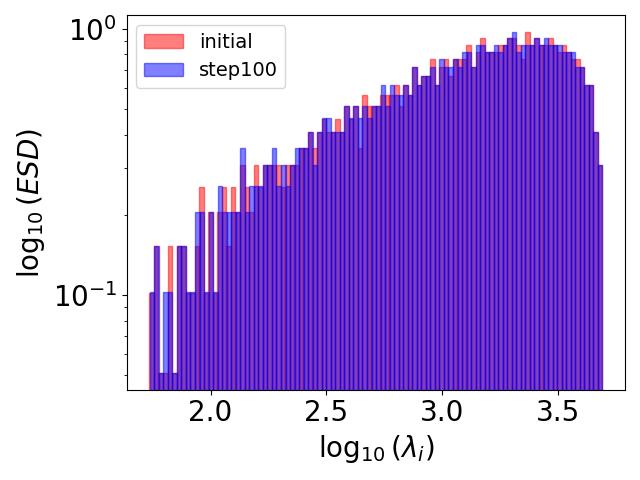}} & 
    \adjustbox{valign=c}{\includegraphics[width=0.24\textwidth]{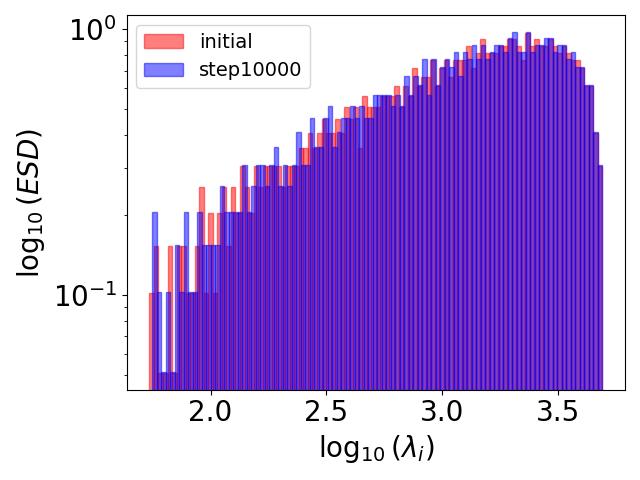}} \\ \hline

    \multirow{2}{*}{$\eta = 100$} & \adjustbox{valign=c}{\includegraphics[width=0.24\textwidth]{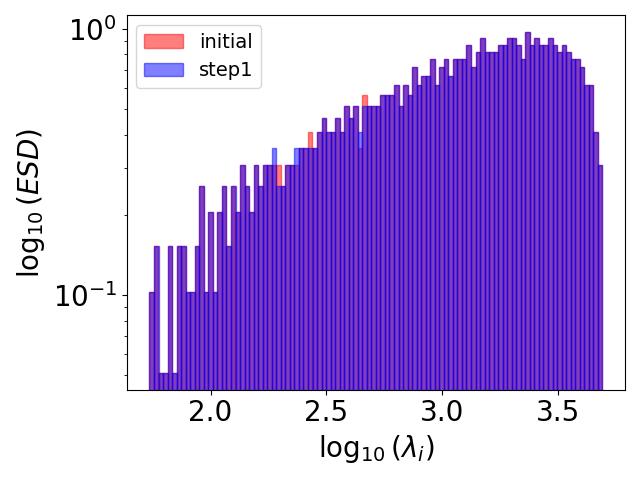}} & 
    \adjustbox{valign=c}{\includegraphics[width=0.24\textwidth]{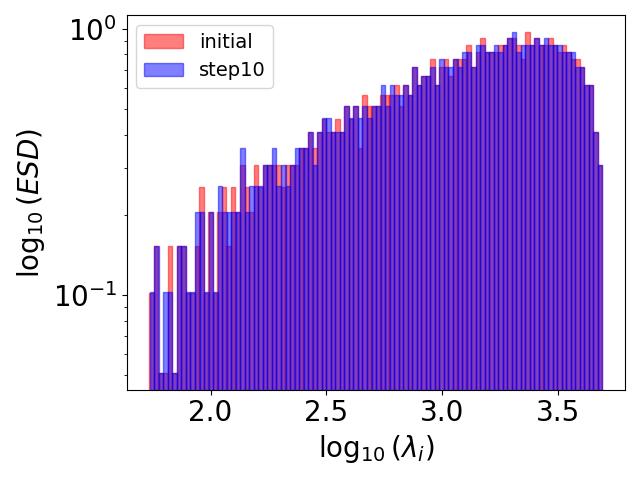}} & 
    \adjustbox{valign=c}{\includegraphics[width=0.24\textwidth]{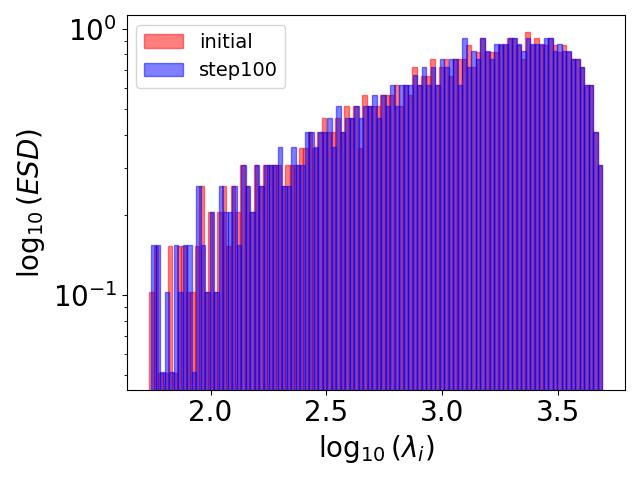}} & 
    \adjustbox{valign=c}{\includegraphics[width=0.24\textwidth]{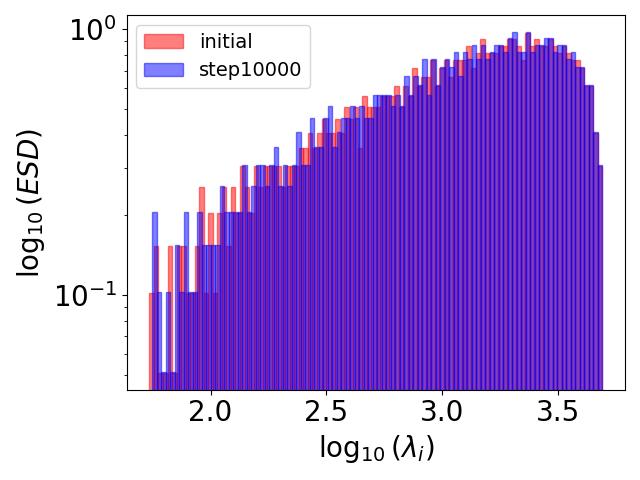}} \\ \hline

    \multirow{2}{*}{$\eta = 2000$} & \adjustbox{valign=c}{\includegraphics[width=0.24\textwidth]{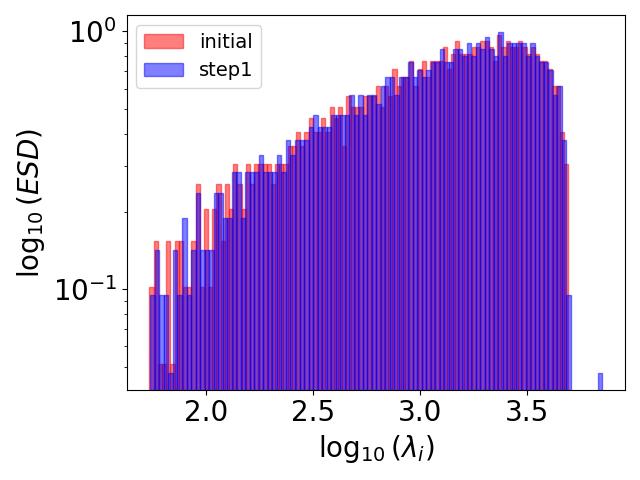}} & 
    \adjustbox{valign=c}{\includegraphics[width=0.24\textwidth]{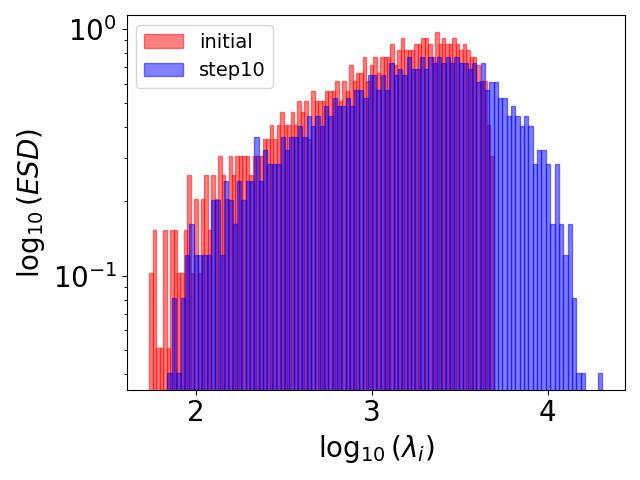}} & 
    \adjustbox{valign=c}{\includegraphics[width=0.24\textwidth]{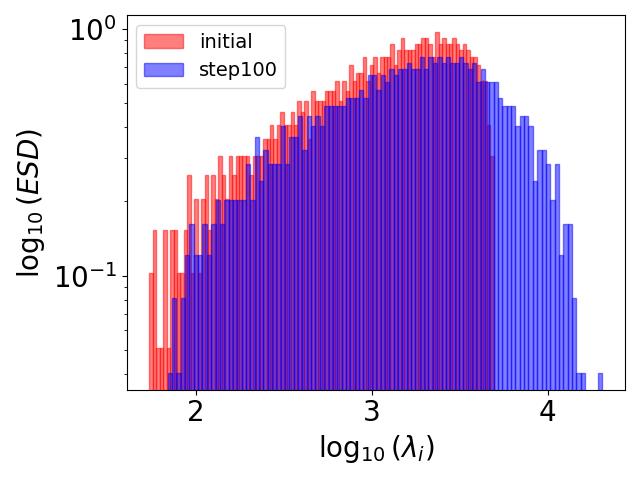}} & 
    \adjustbox{valign=c}{\includegraphics[width=0.24\textwidth]{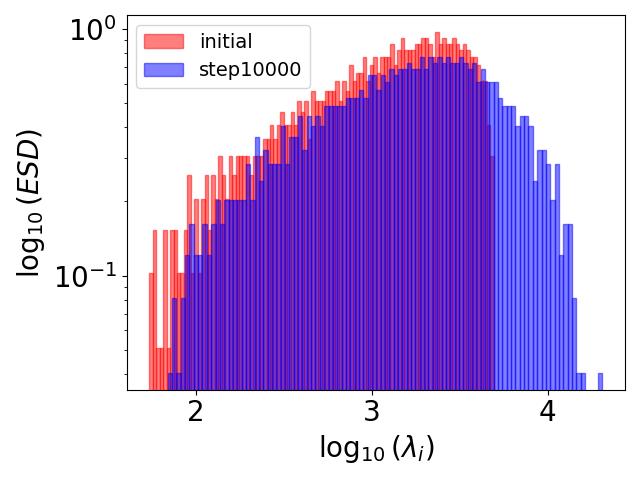}} \\ \hline
\end{tabular}
\caption{Evolution of ESD over different step times $t \in \{1,10,100,10000\}$ with different $\eta \in \{0.1,1,10,100,2000 \}$ for \texttt{GD} optimizer. Through this grid, we highlight the critical $\eta$ for the occurrence of a spike at $t=1$ and the effects of longer training on the emergence of HT ESD.}
\label{tab:GD_transition}
\end{table}

\clearpage
\subsection{ ESD evolution with \texttt{FB-Adam}}
\begin{table}[h!]
\centering
\setlength{\tabcolsep}{1.5pt} 
\renewcommand{\arraystretch}{1.2}

\begin{tabular}{|c|c|c|c|c|}
    \hline
    & $t = 1$ & $t = 10$ & $t = 100$ & $t = 10000$ \\ \hline
    \multirow{2}{*}{$\eta = 0.01$} & \adjustbox{valign=c}{\includegraphics[width=0.24\textwidth]{final_images/long_time_training/adam_lr_0.01_step_1_W0_initial_final_esd.jpg}} & 
    \adjustbox{valign=c}{\includegraphics[width=0.24\textwidth]{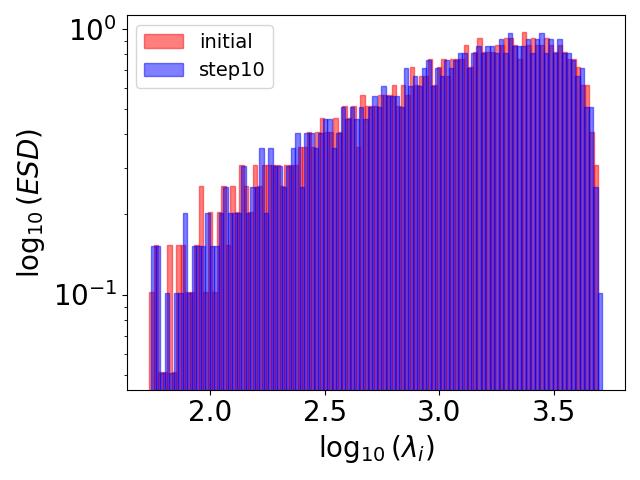}} & 
    \adjustbox{valign=c}{\includegraphics[width=0.24\textwidth]{final_images/long_time_training/adam_lr_0.01_step_100_W0_initial_final_esd.jpg}} & 
    \adjustbox{valign=c}{\includegraphics[width=0.24\textwidth]{final_images/long_time_training/adam_lr_0.01_step_10000_W0_initial_final_esd.jpg}} \\ \hline
    \multirow{2}{*}{$\eta = 0.1$} & \adjustbox{valign=c}{\includegraphics[width=0.24\textwidth]{final_images/long_time_training/adam_lr_0.1_step_1_W0_initial_final_esd.jpg}} & 
    \adjustbox{valign=c}{\includegraphics[width=0.24\textwidth]{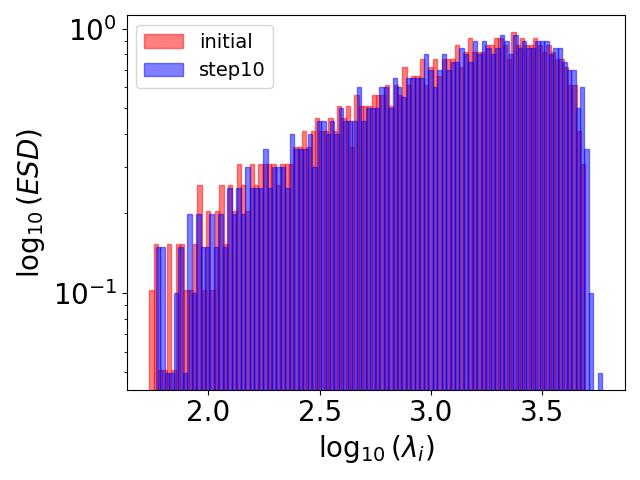}} & 
    \adjustbox{valign=c}{\includegraphics[width=0.24\textwidth]{final_images/long_time_training/adam_lr_0.1_step_100_W0_initial_final_esd.jpg}} & 
    \adjustbox{valign=c}{\includegraphics[width=0.24\textwidth]{final_images/long_time_training/adam_lr_0.1_step_10000_W0_initial_final_esd.jpg}} \\ \hline
    \multirow{2}{*}{$\eta = 1$} & \adjustbox{valign=c}{\includegraphics[width=0.24\textwidth]{final_images/long_time_training/adam_lr_1_step_1_W0_initial_final_esd.jpg}} & 
    \adjustbox{valign=c}{\includegraphics[width=0.24\textwidth]{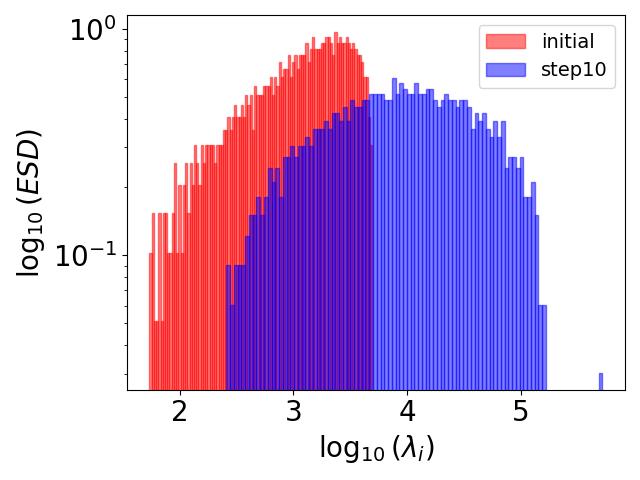}} & 
    \adjustbox{valign=c}{\includegraphics[width=0.24\textwidth]{final_images/long_time_training/adam_lr_1_step_100_W0_initial_final_esd.jpg}} & 
    \adjustbox{valign=c}{\includegraphics[width=0.24\textwidth]{final_images/long_time_training/adam_lr_1_step_10000_W0_initial_final_esd.jpg}} \\ \hline
    \multirow{2}{*}{$\eta = 10$} & \adjustbox{valign=c}{\includegraphics[width=0.24\textwidth]{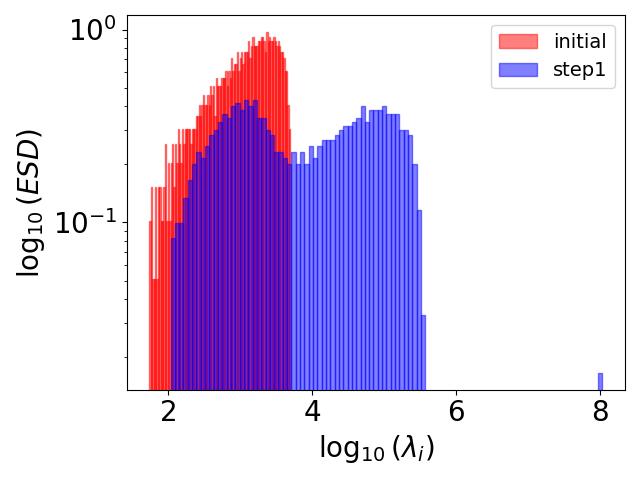}} & 
    \adjustbox{valign=c}{\includegraphics[width=0.24\textwidth]{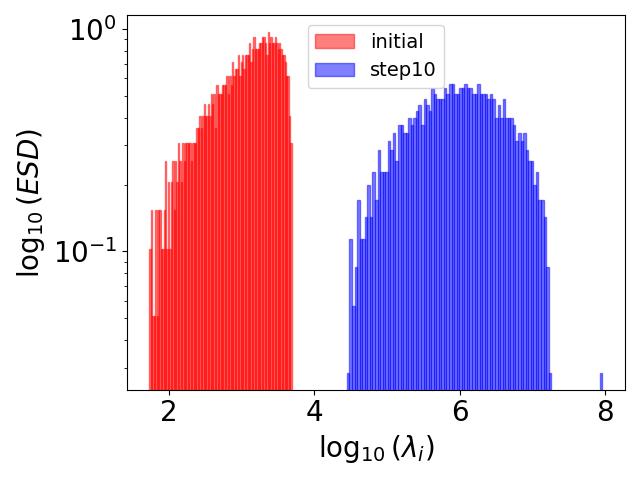}} & 
    \adjustbox{valign=c}{\includegraphics[width=0.24\textwidth]{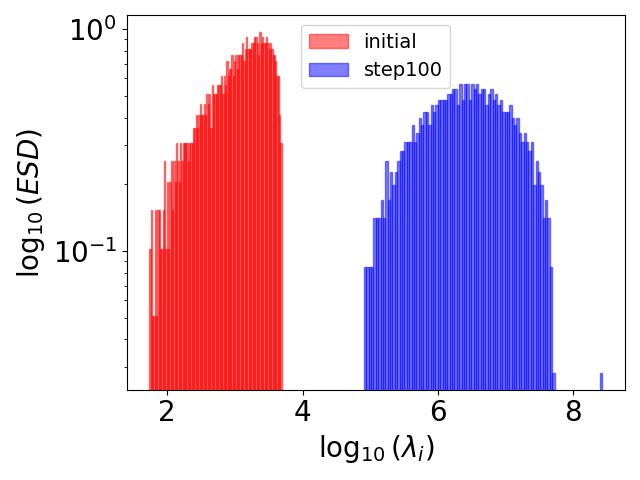}} & 
    \adjustbox{valign=c}{\includegraphics[width=0.24\textwidth]{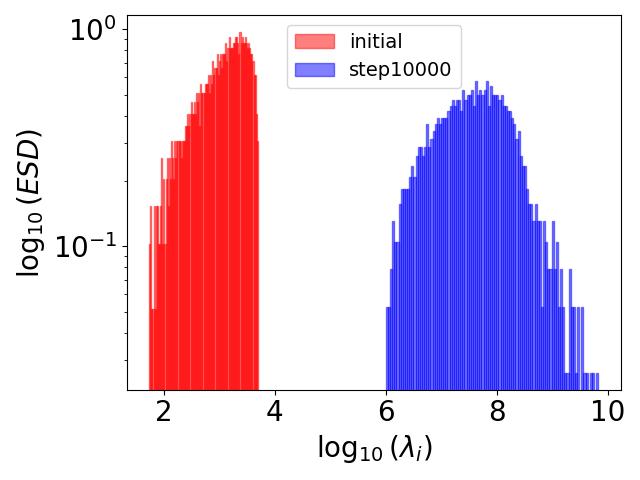}} \\ \hline
    \multirow{2}{*}{$\eta = 100$} & \adjustbox{valign=c}{\includegraphics[width=0.24\textwidth]{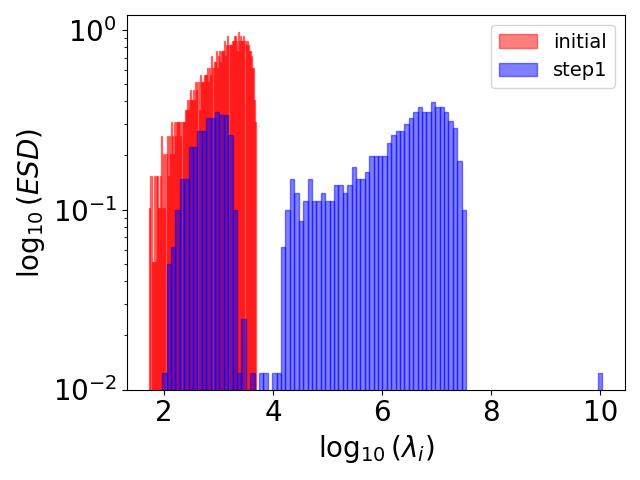}} & 
    \adjustbox{valign=c}{\includegraphics[width=0.24\textwidth]{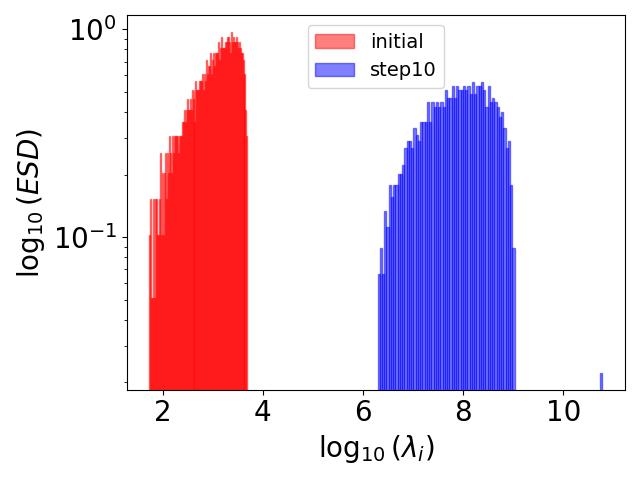}} & 
    \adjustbox{valign=c}{\includegraphics[width=0.24\textwidth]{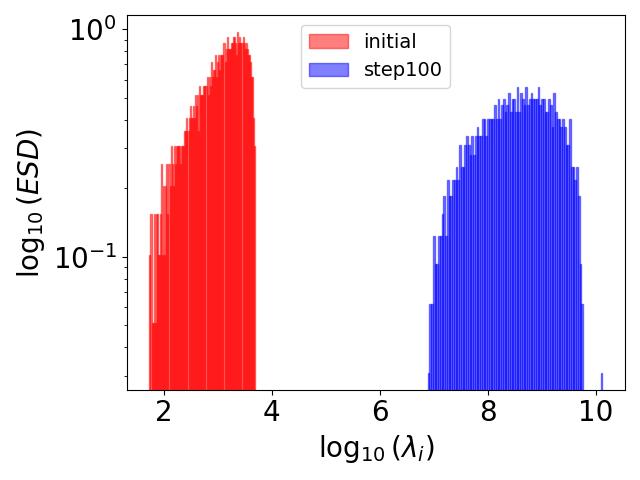}} & 
    \adjustbox{valign=c}{\includegraphics[width=0.24\textwidth]{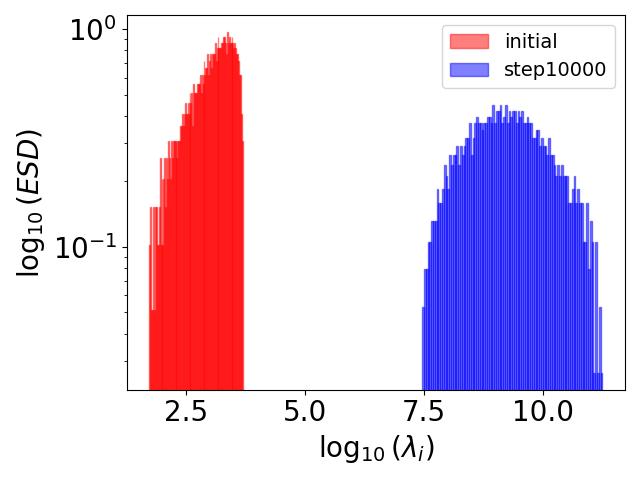}} \\ \hline

\end{tabular}
  \caption{Evolution of ESD over different step times $t \in \{1,10,100,10000\}$ with different $\eta \in \{0.01,0.1,1,10,100\}$ for \texttt{FB-Adam} optimizer. Through this grid, we highlight the critical $\eta$ for the occurrence of a spike at $t=1$ and the effects of longer training on the HT emergence.}
\label{tab:ADAM_transition}
\end{table}

\begin{table}[h!]
\centering
\setlength{\tabcolsep}{1.5pt} 
\renewcommand{\arraystretch}{1.2}
\begin{tabular}{|c|c|c|}
    \hline
    & $t = 1$ & $t = 10^6$ \\ \hline
    \multirow{2}{*}{$\eta = 0.01$} & \adjustbox{valign=c}{\includegraphics[width=0.25\textwidth]{final_images/long_time_training/adam_lr_0.01_step_1_W0_initial_final_esd.jpg}} & 
    \adjustbox{valign=c}{\includegraphics[width=0.25\textwidth]{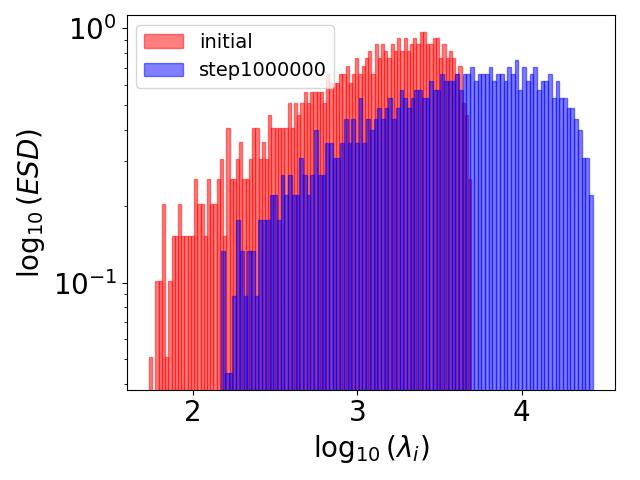}} \\ \hline
\end{tabular}
  \caption{ESD after $t=\{1, 10^6\}$ steps of \texttt{FB-Adam} with $\eta=0.01$. The plot showcases a scenario where a spike does not appear after the first step but results in HT ESD after extremely long training.}
\label{tab:ADAM_transition_t_1000000}
\end{table}

\newpage
\section{Additional Experiments}
\label{app:add_exp}

\paragraph{Hyperparameters:} In most of our experiments, we follow a consistent setup with $n=2000$, $n\_{test}=200$, $d=1000$, $h=1500$, $\lambda=0.01$, $\rho_e=0.3$, $\sigma_* = \texttt{softplus}, \sigma = \texttt{tanh}$. Additionally, we explicitly mention the parameter changes wherever applicable in our experiments.

\subsection{One-Step Optimizer Updates}
\label{app:subsec:one_step}

\begin{figure}[h!]
     \centering
     \begin{subfigure}[b]{0.24\textwidth}
         \centering
         \includegraphics[width=\textwidth]{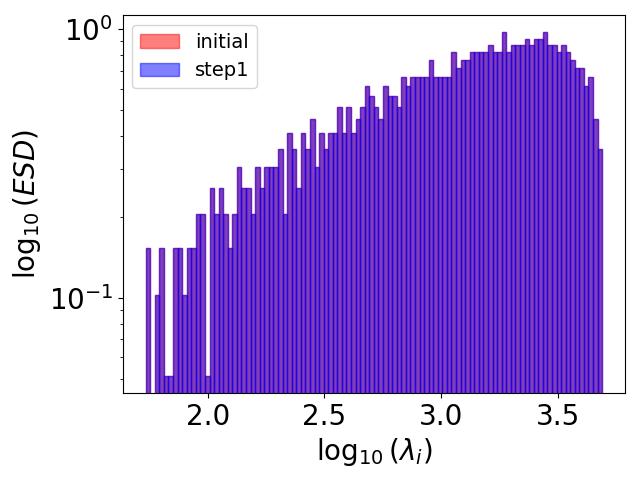}
         \caption{\texttt{GD} $\eta=1$}
     \end{subfigure}
     \hfill
     \begin{subfigure}[b]{0.24\textwidth}
         \centering
         \includegraphics[width=\textwidth]{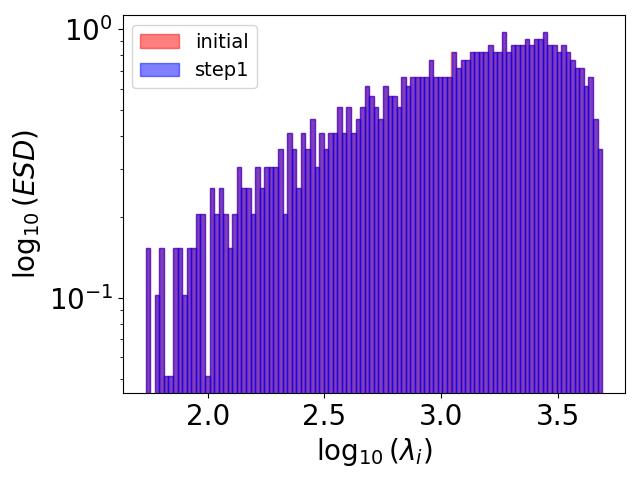}
         \caption{\texttt{GD} $\eta=10$}
     \end{subfigure}
     \hfill
     \begin{subfigure}[b]{0.24\textwidth}
         \centering
         \includegraphics[width=\textwidth]{final_images/fb_adam_lr1_W0_initial_final_step1_esd.jpg}
         \caption{\texttt{FB-Adam} $\eta=1$}
         \label{fig:app:gd_fb_adam_W_esd_n_2000_fb_adam_esd}
     \end{subfigure}
     \hfill
     \begin{subfigure}[b]{0.24\textwidth}
         \centering
         \includegraphics[width=\textwidth]{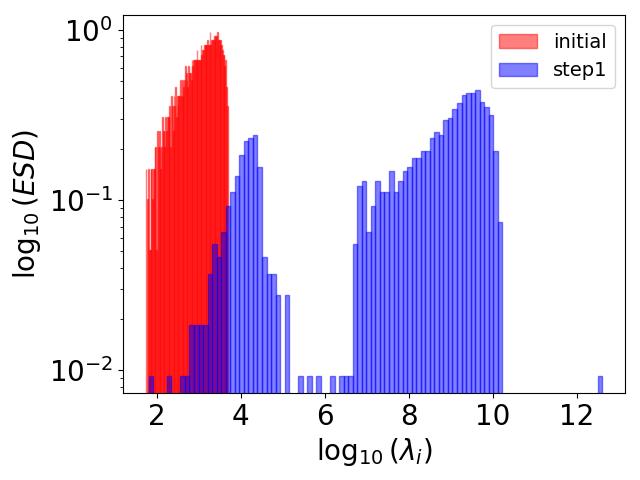}
         \caption{\texttt{FB-Adam} $\eta=2000$}
     \end{subfigure}
        \caption{Evolution of ESD of $\mW^\top\mW$ after one step optimizer update with varying learning rates. Here $n=2000$, $d=1000$, $h=1500$, $\sigma_* = \texttt{softplus}, \sigma = \texttt{tanh}, \rho_e = 0.3$.
}
        \label{fig:app:gd_fb_adam_W_esd_n_2000}
        \vspace{-2mm}
\end{figure}

\begin{figure}[h!]
     \centering
     \begin{subfigure}[b]{0.24\textwidth}
         \centering
         \includegraphics[width=\textwidth]{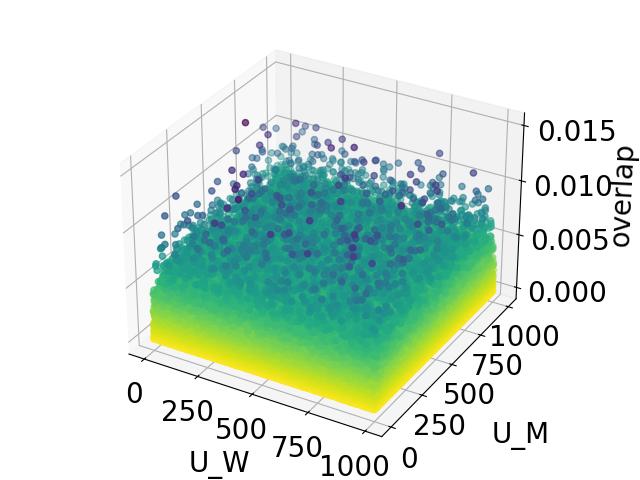}
         \caption{ $\gO(\mU_{W_0}, \mU_{M_{0}})$ } 
     \end{subfigure}
     \hfill
     \begin{subfigure}[b]{0.24\textwidth}
         \centering
         \includegraphics[width=\textwidth]{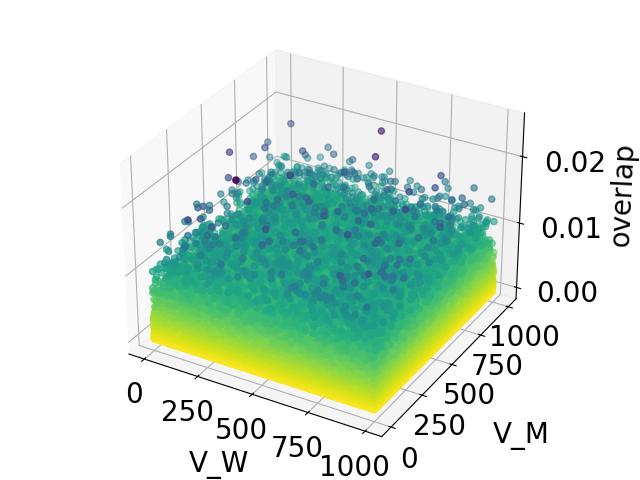}
         \caption{ $\gO(\mV_{W_0}, \mV_{M_{0}})$ } 
     \end{subfigure}
     \hfill
     \begin{subfigure}[b]{0.24\textwidth}
         \centering
         \includegraphics[width=\textwidth]{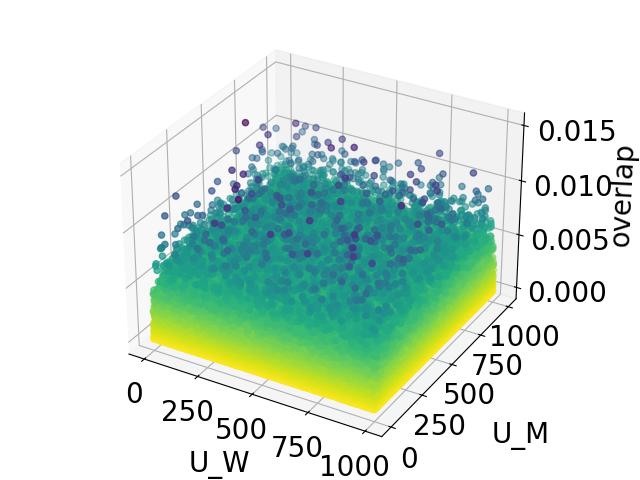}
         \caption{ $\gO(\mU_{W_1}, \mU_{M_{0}})$ } 
     \end{subfigure}
     \hfill
     \begin{subfigure}[b]{0.24\textwidth}
         \centering
         \includegraphics[width=\textwidth]{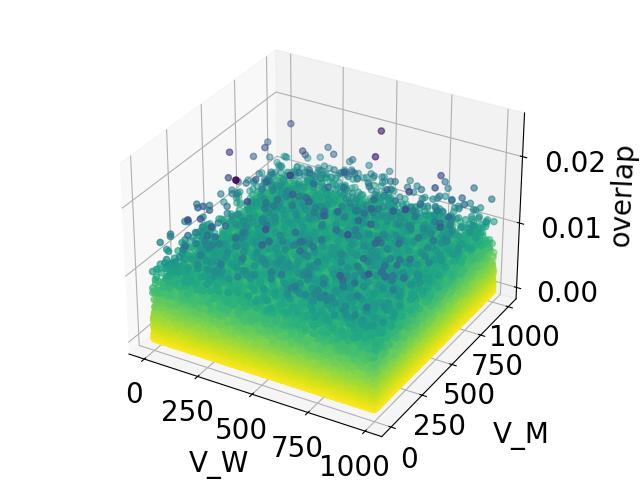}
         \caption{ $\gO(\mV_{W_1}, \mV_{M_{0}})$ } 
     \end{subfigure}
        \caption{ Overlaps between singular vectors after one step \texttt{GD} update with $\eta=0.1$. Here $n=2000$, $d=1000$, $h=1500$, $\sigma_* = \texttt{softplus}, \sigma = \texttt{tanh}, \rho_e = 0.3$. }
        \label{fig:app:overlap_gd_lr0.1_n_2000}
        \vspace{-2mm}
\end{figure}

\begin{figure}[h!]
     \centering
     \begin{subfigure}[b]{0.24\textwidth}
         \centering
         \includegraphics[width=\textwidth]{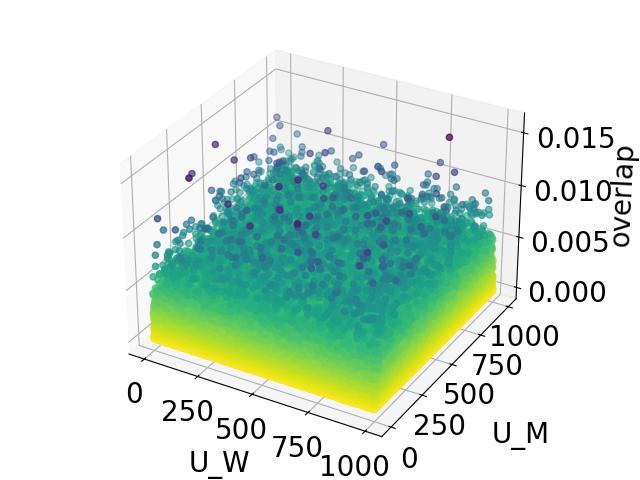}
         \caption{ $\gO(\mU_{W_0}, \mU_{M_{0}})$ } 
     \end{subfigure}
     \hfill
     \begin{subfigure}[b]{0.24\textwidth}
         \centering
         \includegraphics[width=\textwidth]{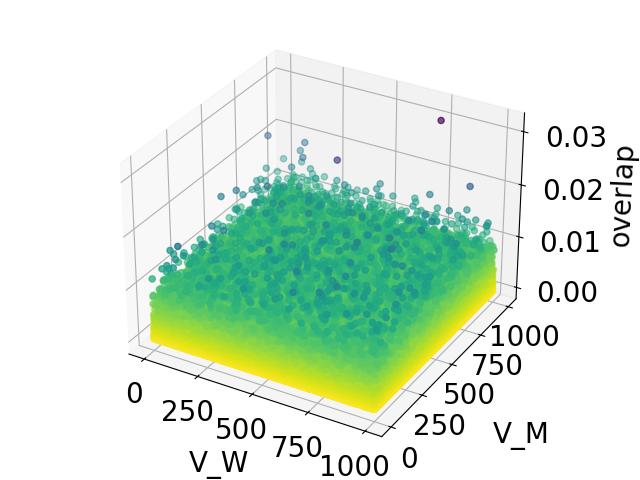}
         \caption{ $\gO(\mV_{W_0}, \mV_{M_{0}})$ } 
     \end{subfigure}
     \hfill
     \begin{subfigure}[b]{0.24\textwidth}
         \centering
         \includegraphics[width=\textwidth]{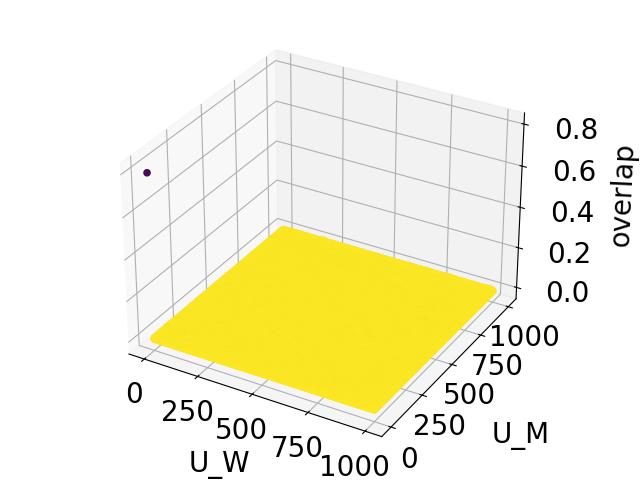}
         \caption{ $\gO(\mU_{W_1}, \mU_{M_{0}})$ } 
     \end{subfigure}
     \hfill
     \begin{subfigure}[b]{0.24\textwidth}
         \centering
         \includegraphics[width=\textwidth]{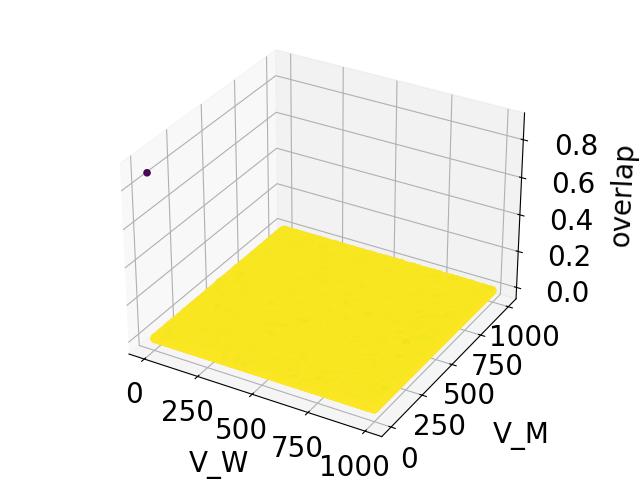}
         \caption{ $\gO(\mV_{W_1}, \mV_{M_{0}})$ } 
     \end{subfigure}
        \caption{ Overlaps between singular vectors after one step \texttt{FB-Adam} update with $\eta=0.1$. Here $n=2000$, $d=1000$, $h=1500$, $\sigma_* = \texttt{softplus}, \sigma = \texttt{tanh}, \rho_e = 0.3$. }
        \label{fig:app:overlap_fb_adam_lr0.1_n_2000}
\end{figure}

\begin{figure}[h!]
     \centering
     \begin{subfigure}[b]{0.24\textwidth}
         \centering
         \includegraphics[width=\textwidth]{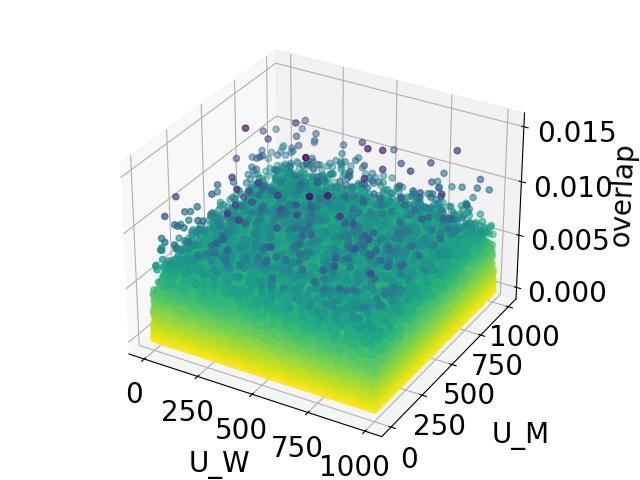}
         \caption{ $\gO(\mU_{W_0}, \mU_{M_{0}})$ } 
     \end{subfigure}
     \hfill
     \begin{subfigure}[b]{0.24\textwidth}
         \centering
         \includegraphics[width=\textwidth]{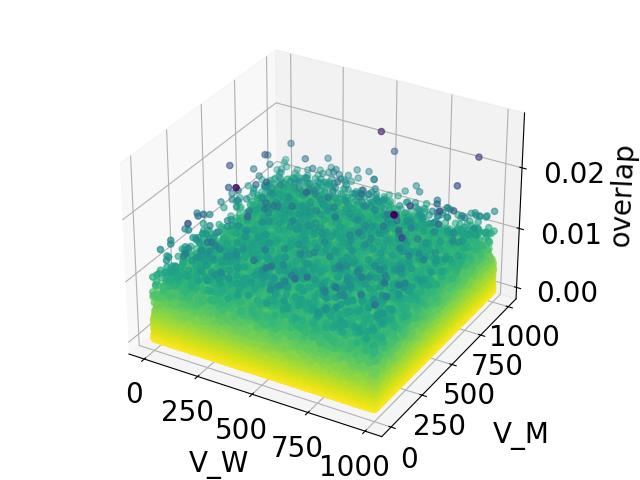}
         \caption{ $\gO(\mV_{W_0}, \mV_{M_{0}})$ } 
     \end{subfigure}
     \hfill
     \begin{subfigure}[b]{0.24\textwidth}
         \centering
         \includegraphics[width=\textwidth]{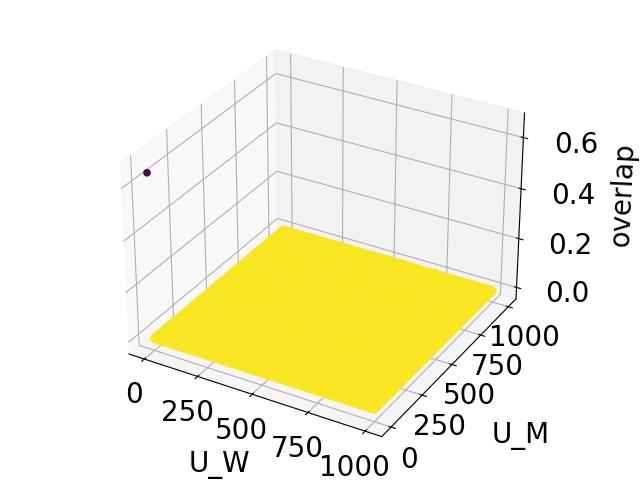}
         \caption{ $\gO(\mU_{W_1}, \mU_{M_{0}})$ } 
         \label{fig:app:overlap_gd_lr2000_n_2000_u_w1_m0}
     \end{subfigure}
     \hfill
     \begin{subfigure}[b]{0.24\textwidth}
         \centering
         \includegraphics[width=\textwidth]{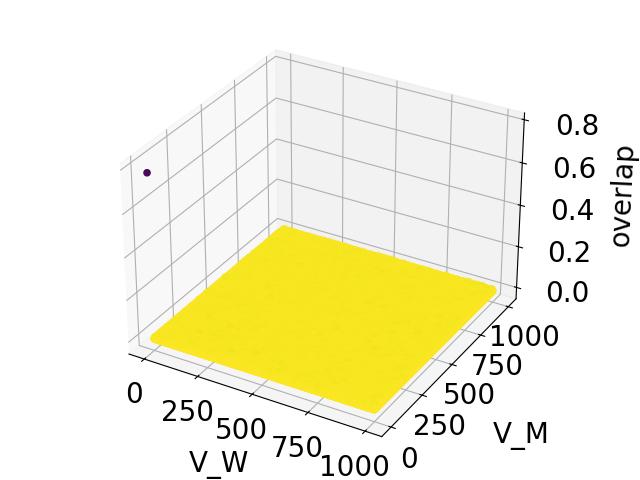}
         \caption{ $\gO(\mV_{W_1}, \mV_{M_{0}})$ } 
         \label{fig:app:overlap_gd_lr2000_n_2000_v_w1_m0}
     \end{subfigure}
        \caption{ Overlaps between singular vectors after one step \texttt{GD} update with $\eta=2000$. Here $n=2000$, $d=1000$, $h=1500$, $\sigma_* = \texttt{softplus}, \sigma = \texttt{tanh}, \rho_e = 0.3$. }
        \label{fig:app:overlap_gd_lr2000_n_2000}
\end{figure}

\begin{figure}[h!]
     \centering
     \begin{subfigure}[b]{0.32\textwidth}
         \centering
         \includegraphics[width=\textwidth]{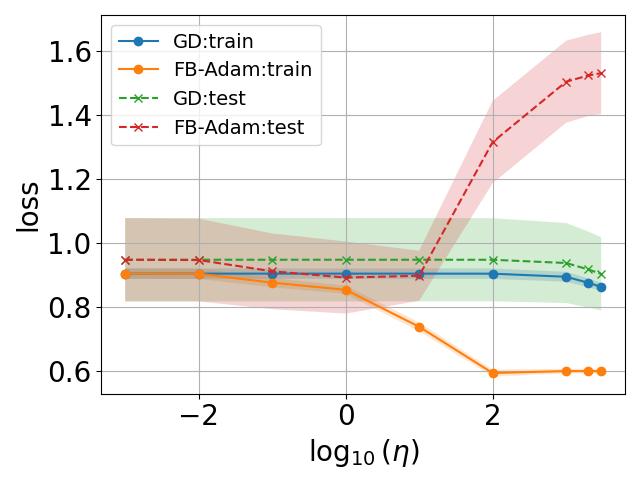}
         \caption{loss}
         \label{fig:app:gd_fb_adam_bulk_loss_alignments_n_4000_loss}
     \end{subfigure}
     \hfill
     \begin{subfigure}[b]{0.32\textwidth}
         \centering
         \includegraphics[width=\textwidth]{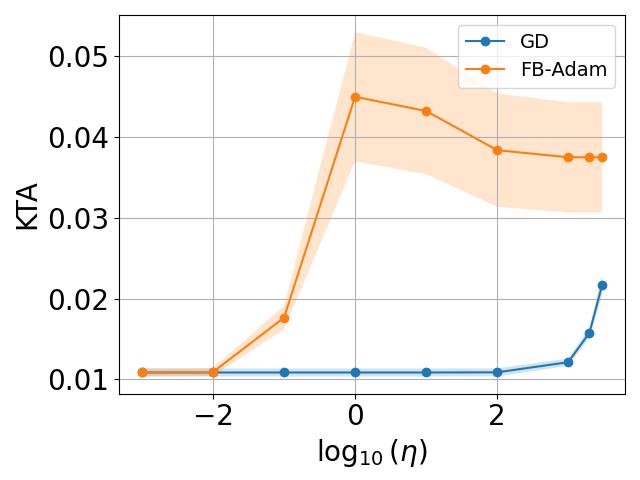}
         \caption{\texttt{KTA}}
         \label{fig:app:gd_fb_adam_bulk_loss_alignments_n_4000_kta}
     \end{subfigure}
     \hfill
     \begin{subfigure}[b]{0.32\textwidth}
         \centering
         \includegraphics[width=\textwidth]{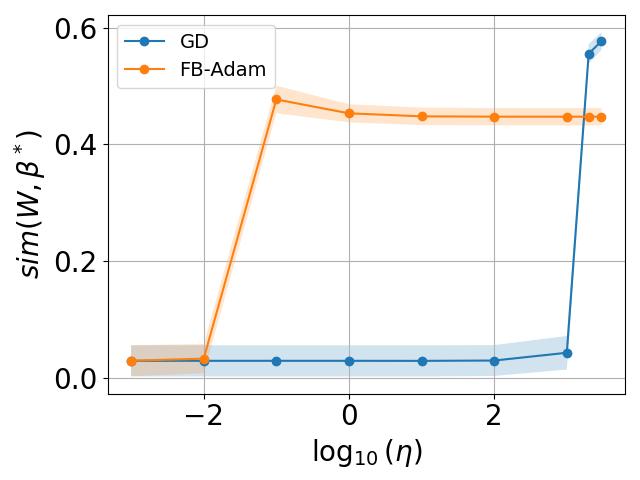}
         \caption{\texttt{sim}$(\mW, \vbeta^*)$}
         \label{fig:app:gd_fb_adam_bulk_loss_alignments_n_4000_sim}
     \end{subfigure}
        \caption{Train/test losses, \texttt{KTA}, \texttt{sim}$(\mW, \vbeta^*)$ for $f(\cdot)$ trained with one-step of  \texttt{GD}, \texttt{FB-Adam}. Here $n=4000$, $d=1000$, $h=1500$, $\sigma_* = \texttt{softplus}, \sigma = \texttt{tanh}, \rho_e = 0.3$.}
        \label{fig:app:gd_fb_adam_bulk_loss_alignments_n_4000}
\end{figure}

\begin{figure}[h!]
     \centering
     \begin{subfigure}[b]{0.32\textwidth}
         \centering
         \includegraphics[width=\textwidth]{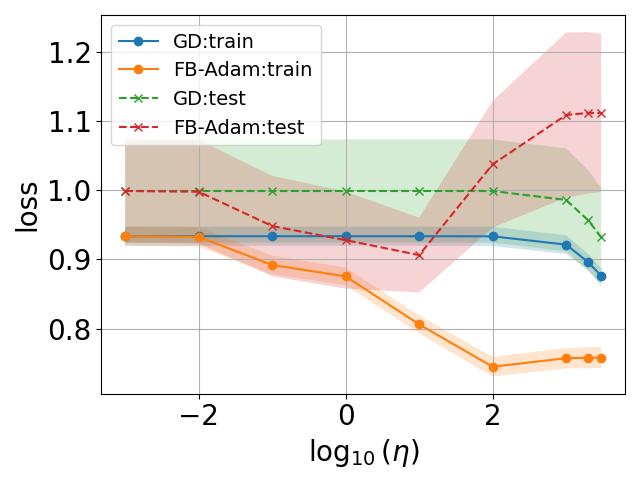}
         \caption{loss}
         \label{fig:app:gd_fb_adam_bulk_loss_alignments_n_8000_loss}
     \end{subfigure}
     \hfill
     \begin{subfigure}[b]{0.32\textwidth}
         \centering
         \includegraphics[width=\textwidth]{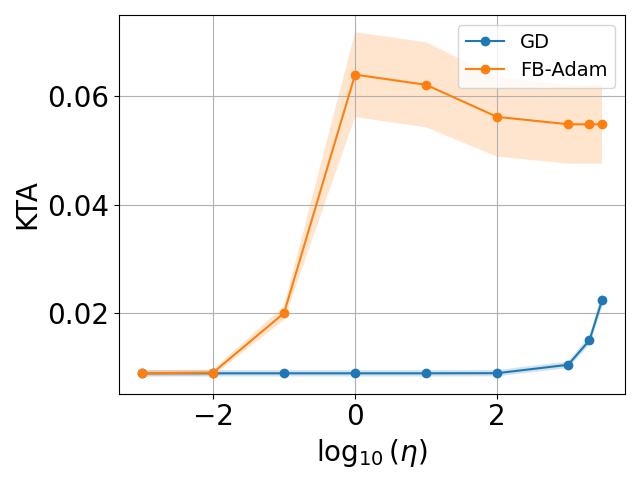}
         \caption{\texttt{KTA}}
         \label{fig:app:gd_fb_adam_bulk_loss_alignments_n_8000_kta}
     \end{subfigure}
     \hfill
     \begin{subfigure}[b]{0.32\textwidth}
         \centering
         \includegraphics[width=\textwidth]{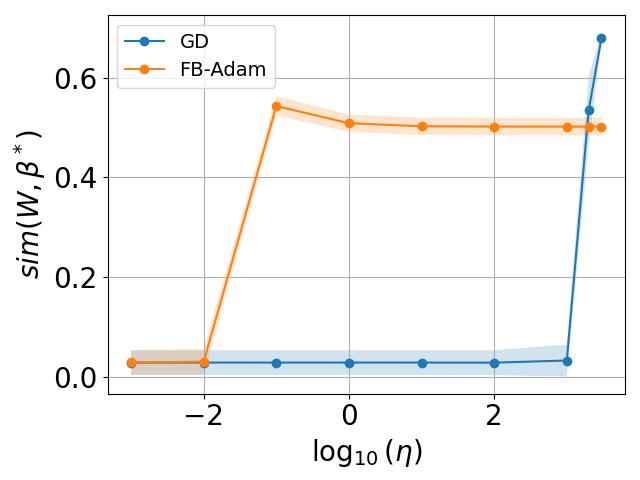}
         \caption{\texttt{sim}$(\mW, \vbeta^*)$
         }
         \label{fig:app:gd_fb_adam_bulk_loss_alignments_n_8000_sim}
     \end{subfigure}
        \caption{Train/test losses, \texttt{KTA}, \texttt{sim}$(\mW, \vbeta^*)$ for $f(\cdot)$ trained with one-step of  \texttt{GD}, \texttt{FB-Adam}. Here $n=8000$, $d=1000$, $h=1500$, $\sigma_* = \texttt{softplus}, \sigma = \texttt{tanh}, \rho_e = 0.3$.}
        \label{fig:app:gd_fb_adam_bulk_loss_alignments_n_8000}
\end{figure}

In this section, we present additional experiments for one-step optimizer updates. Figure \ref{fig:app:gd_fb_adam_W_esd_n_2000} leverages the same experimental setup as Section \ref{sec:one_step_fb_adam_update} and illustrates the ESD of $\mW^\top\mW$ after the first step of \texttt{GD} and \texttt{FB-Adam}. Based on the results obtained in the main text, $\eta=2000$ is an extremely large learning rate for \texttt{FB-Adam}, which results in a clear bimodal distribution. Note that the tendency towards such a distribution was already observed with $\eta=10$ in Figure \ref{fig:main:gd_fb_adam_W_esd} (Section \ref{sec:one_step_fb_adam_update}).  Based on the same setup as Section \ref{subsec:singular_vector_alignments}, Figures \ref{fig:app:overlap_gd_lr0.1_n_2000}, \ref{fig:app:overlap_fb_adam_lr0.1_n_2000}, \ref{fig:app:overlap_gd_lr2000_n_2000} represent the overlaps of singular vectors after one-step update and showcase the presence of outliers for sufficiently large $\eta$. Finally, we illustrate the role of sample sizes on the losses, \texttt{KTA} and \texttt{sim$(\mW, \vbeta^*)$} in Figure \ref{fig:app:gd_fb_adam_bulk_loss_alignments_n_4000} (for $n=4000$)  and in Figure \ref{fig:app:gd_fb_adam_bulk_loss_alignments_n_8000} (for $n=8000$).

\clearpage
\subsection{$10$ Step Optimizer Updates}
\label{app:subsec:10_steps}

\begin{figure}[h!]
     \centering
     \begin{subfigure}[b]{0.23\textwidth}
         \centering
         \includegraphics[width=\textwidth]{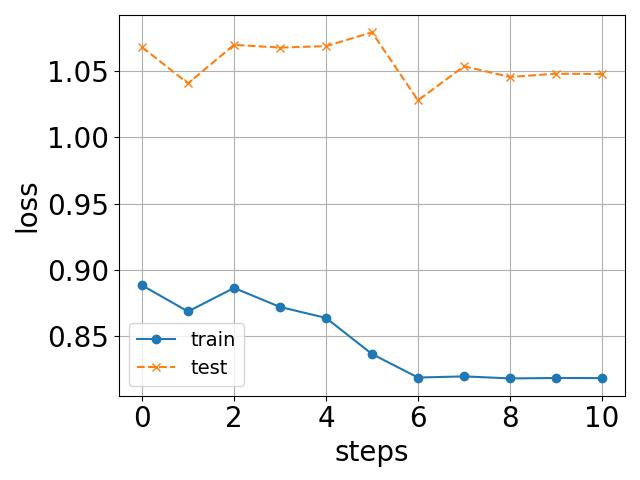}
         \caption{losses}
         \label{fig:app:10_step_gd_lr2000_W_losses}
     \end{subfigure}
     \hfill
     \begin{subfigure}[b]{0.23\textwidth}
         \centering
         \includegraphics[width=\textwidth]{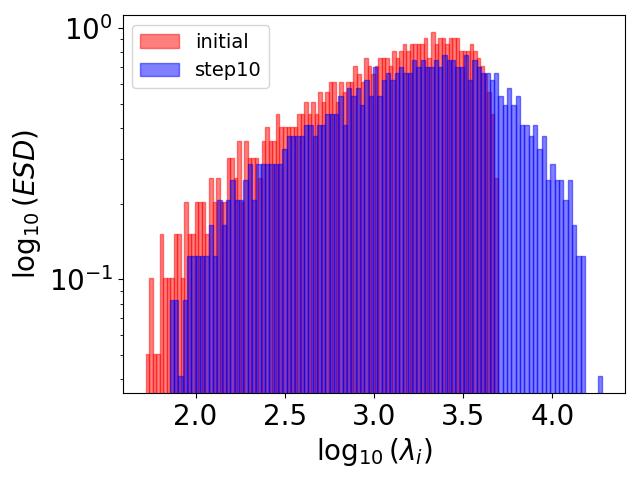}
         \caption{ESD of $\mW^\top\mW$}
         \label{fig:app:10_step_gd_lr2000_W_esd}
     \end{subfigure}
     \hfill
     \begin{subfigure}[b]{0.25\textwidth}
         \centering
         \includegraphics[width=\textwidth]{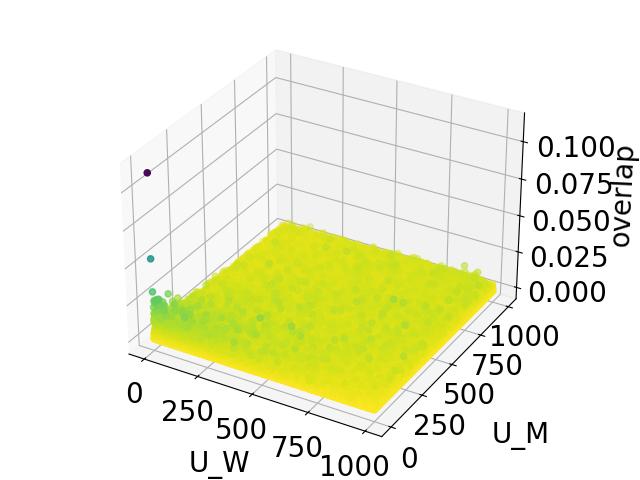}
         \caption{ $\gO(\mU_{W_{10}}, \mU_{M_9})$ } 
         \label{fig:app:overlap_gd_lr2000_u_o10}
     \end{subfigure}
     \hfill
     \begin{subfigure}[b]{0.25\textwidth}
         \centering
         \includegraphics[width=\textwidth]{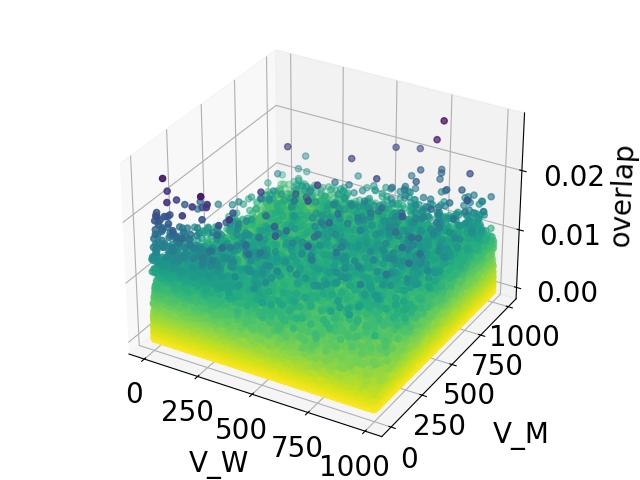}
         \caption{ $\gO(\mV_{W_{10}}, \mV_{M_9})$ } 
         \label{fig:app:overlap_gd_lr2000_v_o10}
     \end{subfigure}
        \caption{Losses, ESD, and Overlaps between singular vectors after 10 steps of \texttt{GD}($\eta=2000$). Here $n=2000$, $d=1000$, $h=1500$, $\sigma_* = \texttt{softplus}, \sigma = \texttt{tanh}, \rho_e = 0.3$. }
        \label{fig:app:overlap_gd_lr2000_step10}
\end{figure}

\begin{figure}[h!]
     \centering
     \begin{subfigure}[b]{0.24\textwidth}
         \centering
         \includegraphics[width=\textwidth]{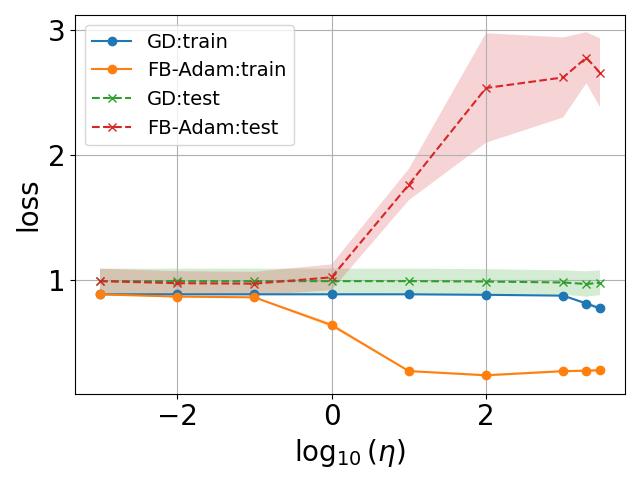}
         \caption{loss}
         \label{fig:app:gd_fb_adam_bulk_loss_alignments_10_steps_n_2000_loss}
     \end{subfigure}
     \hfill
     \begin{subfigure}[b]{0.24\textwidth}
         \centering
         \includegraphics[width=\textwidth]{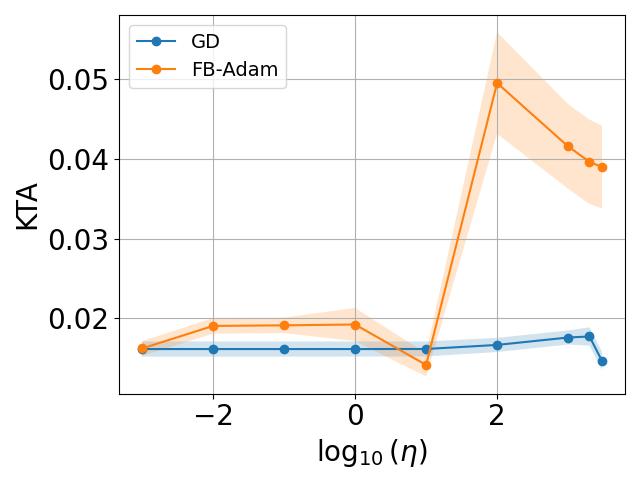}
         \caption{\texttt{KTA}}
         \label{fig:app:gd_fb_adam_bulk_loss_alignments_10_steps_n_2000_kta}
     \end{subfigure}
     \hfill
     \begin{subfigure}[b]{0.24\textwidth}
         \centering
         \includegraphics[width=\textwidth]{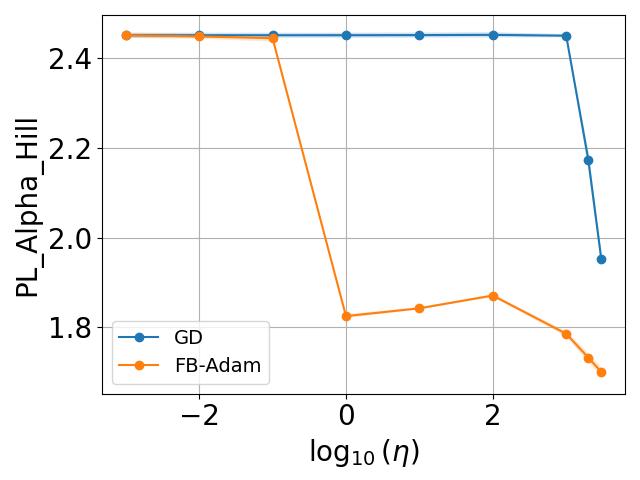}
         \caption{\texttt{PL\_Alpha\_Hill}}
         \label{fig:app:gd_fb_adam_bulk_loss_alignments_10_steps_n_2000_alpha}
     \end{subfigure}
     \hfill
     \begin{subfigure}[b]{0.24\textwidth}
         \centering
         \includegraphics[width=\textwidth]{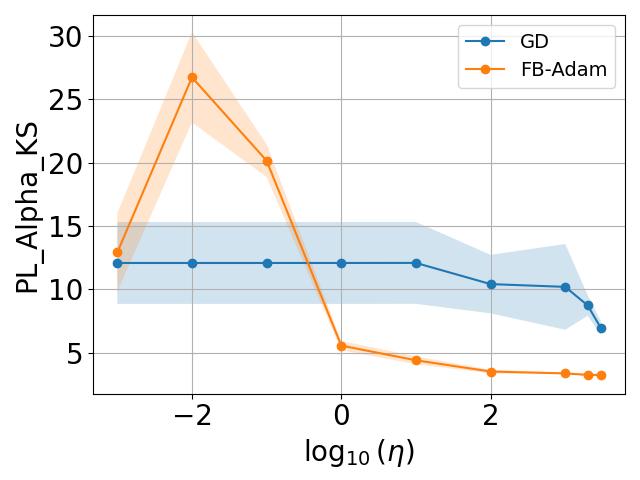}
         \caption{\texttt{PL\_Alpha\_KS}}
         \label{fig:app:gd_fb_adam_bulk_loss_alignments_10_steps_n_2000_alpha_ks}
     \end{subfigure}
        \caption{Train/test losses, \texttt{KTA}, \texttt{PL\_Alpha\_Hill}, \texttt{PL\_Alpha\_KS} for $f(\cdot)$ trained with $10$ steps of  \texttt{GD}, \texttt{FB-Adam}. Here $n=2000$, $d=1000$, $h=1500$, $\sigma_* = \texttt{softplus}, \sigma = \texttt{tanh}, \rho_e = 0.3$.}
        \label{fig:app:gd_fb_adam_bulk_loss_alignments_10_steps_n_2000}
\end{figure}

\begin{figure}[h!]
     \centering
     \begin{subfigure}[b]{0.24\textwidth}
         \centering
         \includegraphics[width=\textwidth]{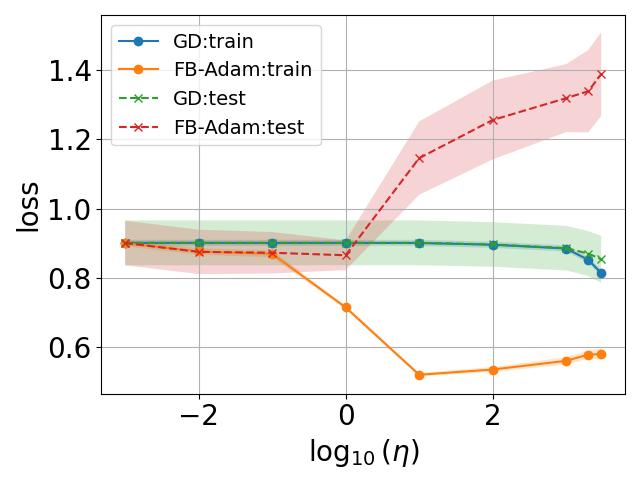}
         \caption{loss}
         \label{fig:app:gd_fb_adam_bulk_loss_alignments_10_steps_n_4000_loss}
     \end{subfigure}
     \hfill
     \begin{subfigure}[b]{0.24\textwidth}
         \centering
         \includegraphics[width=\textwidth]{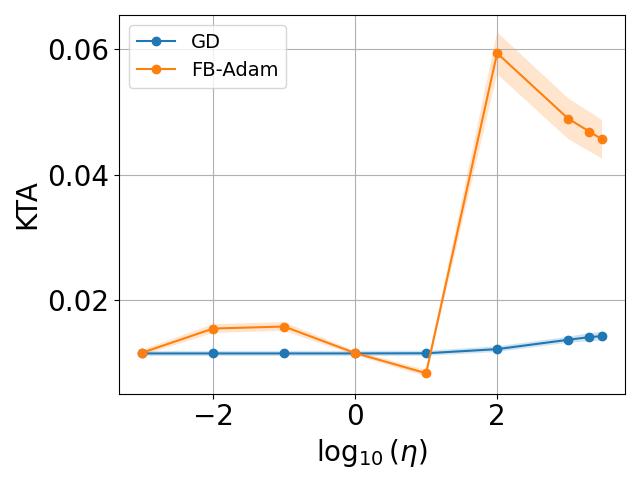}
         \caption{\texttt{KTA}}
         \label{fig:app:gd_fb_adam_bulk_loss_alignments_10_steps_n_4000_kta}
     \end{subfigure}
     \hfill
     \begin{subfigure}[b]{0.24\textwidth}
         \centering
         \includegraphics[width=\textwidth]{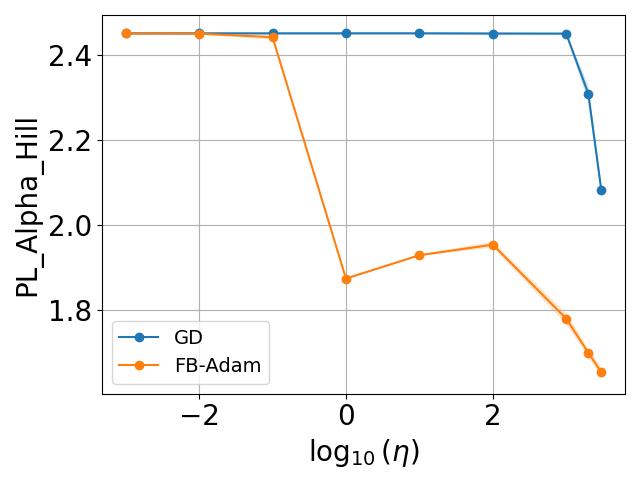}
         \caption{\texttt{PL\_Alpha\_Hill}}
         \label{fig:app:gd_fb_adam_bulk_loss_alignments_10_steps_n_4000_alpha}
     \end{subfigure}
     \hfill
     \begin{subfigure}[b]{0.24\textwidth}
         \centering
         \includegraphics[width=\textwidth]{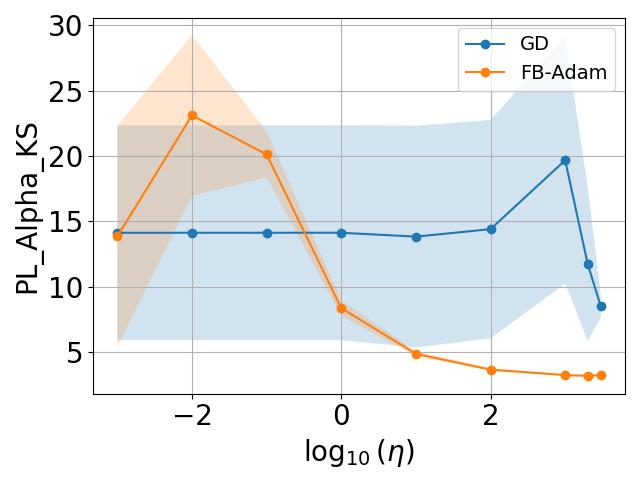}
         \caption{\texttt{PL\_Alpha\_KS}}
         \label{fig:app:gd_fb_adam_bulk_loss_alignments_10_steps_n_4000_alpha_ks}
     \end{subfigure}
        \caption{ Train/test losses, \texttt{KTA}, \texttt{PL\_Alpha\_Hill}, \texttt{PL\_Alpha\_KS} for $f(\cdot)$ trained with $10$ steps of  \texttt{GD}, \texttt{FB-Adam}. Here $n=4000$, $d=1000$, $h=1500$, $\sigma_* = \texttt{softplus}, \sigma = \texttt{tanh}, \rho_e = 0.3$.}
        \label{fig:app:gd_fb_adam_bulk_loss_alignments_10_steps_n_4000}
\end{figure}

\begin{figure}[h!]
     \centering
     \begin{subfigure}[b]{0.24\textwidth}
         \centering
         \includegraphics[width=\textwidth]{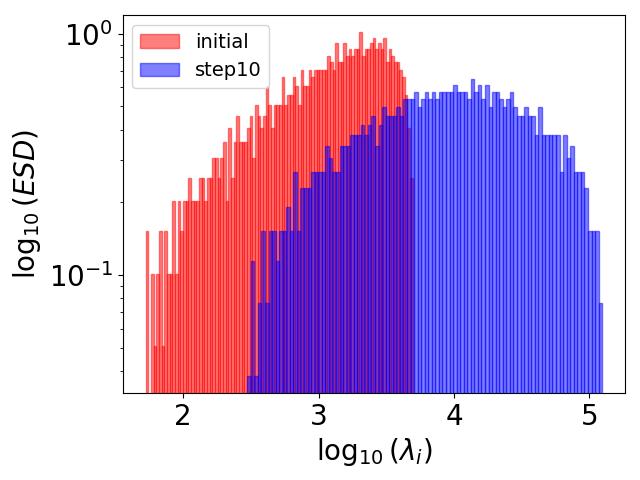}
         \caption{$\eta=1$}
     \end{subfigure}
     \hfill
     \begin{subfigure}[b]{0.24\textwidth}
         \centering
         \includegraphics[width=\textwidth]{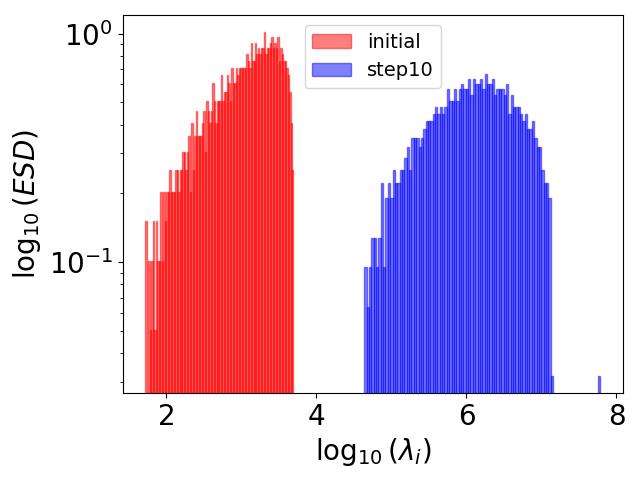}
         \caption{$\eta=10$}
     \end{subfigure}
     \hfill
     \begin{subfigure}[b]{0.24\textwidth}
         \centering
         \includegraphics[width=\textwidth]{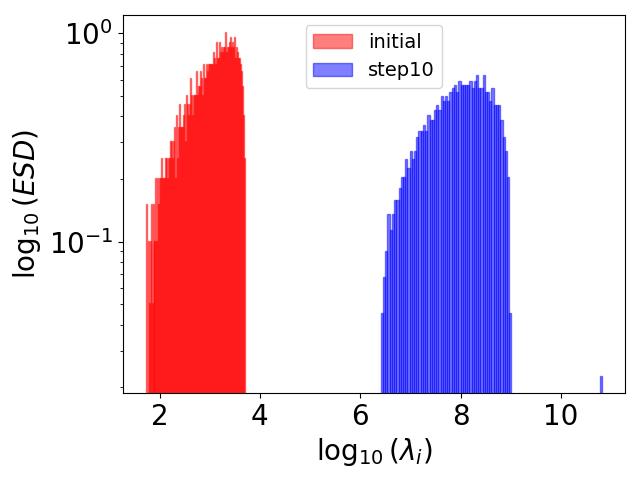}
         \caption{$\eta=100$}
     \end{subfigure}
     \hfill
     \begin{subfigure}[b]{0.24\textwidth}
         \centering
         \includegraphics[width=\textwidth]{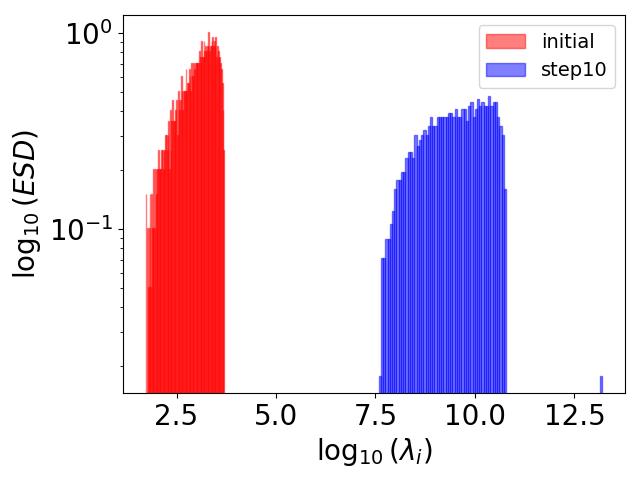}
         \caption{$\eta=1000$}
     \end{subfigure}
     \hfill
     \begin{subfigure}[b]{0.24\textwidth}
         \centering
         \includegraphics[width=\textwidth]{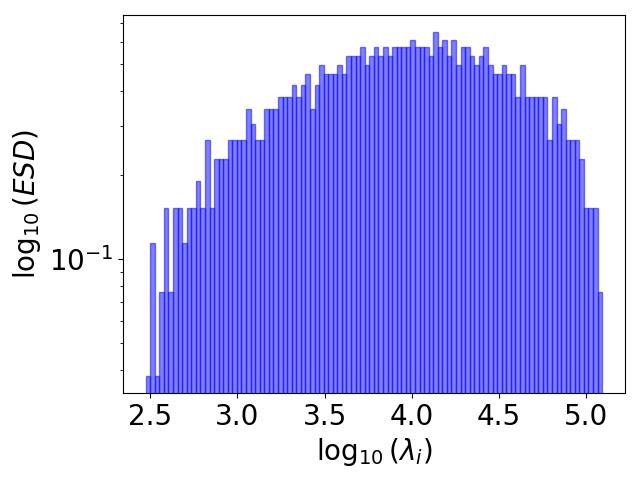}
         \caption{$\eta=1$}
     \end{subfigure}
     \hfill
     \begin{subfigure}[b]{0.24\textwidth}
         \centering
         \includegraphics[width=\textwidth]{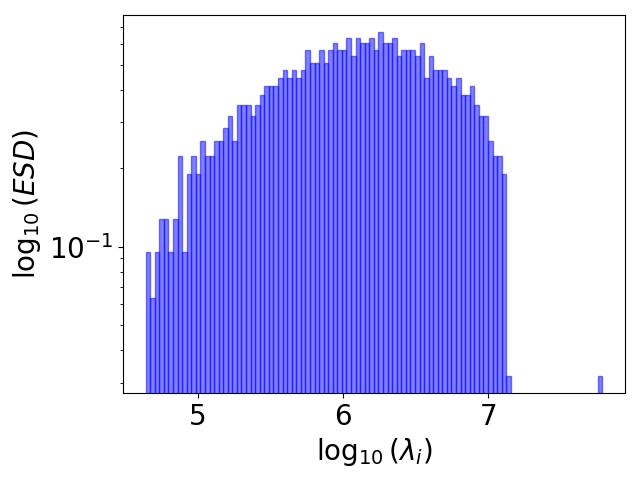}
         \caption{$\eta=10$}
     \end{subfigure}
     \hfill
     \begin{subfigure}[b]{0.24\textwidth}
         \centering
         \includegraphics[width=\textwidth]{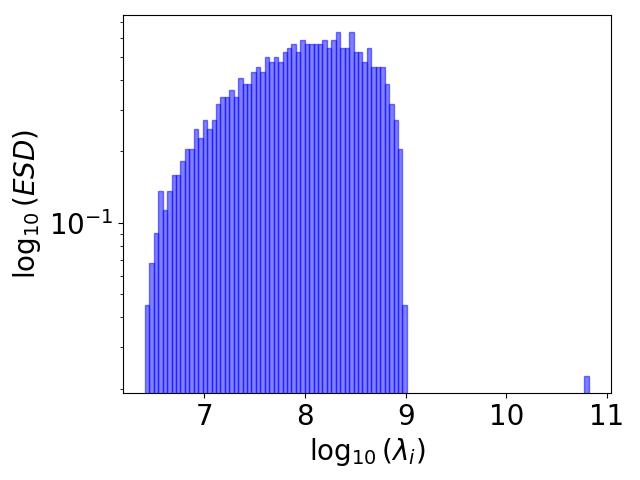}
         \caption{$\eta=100$}
     \end{subfigure}
     \hfill
     \begin{subfigure}[b]{0.24\textwidth}
         \centering
         \includegraphics[width=\textwidth]{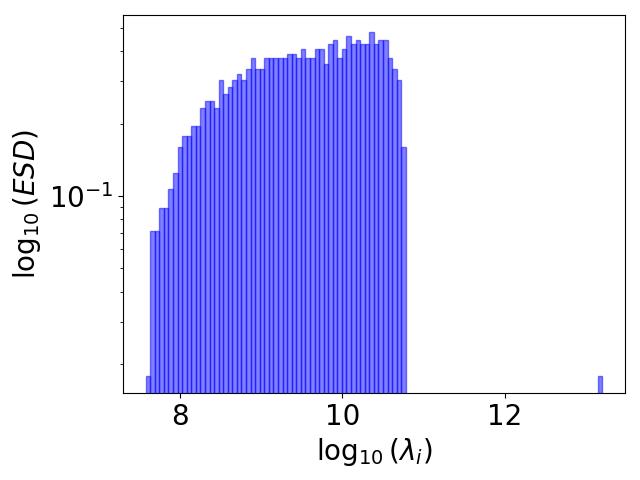}
         \caption{$\eta=1000$}
     \end{subfigure}
        \caption{Evolution of ESD of $\mW^\top\mW$ after $10$ steps of \texttt{FB-Adam} updates with $n=8000, d=1000, h=1500, \lambda=0.01, \rho_e=0.3$. The first row compares the initial and final ESDs. The second row illustrates solely the final ESD of $\mW^\top\mW$ (i.e. $\mW_{10}^\top\mW_{10}$) for better visualizations of the shape. }
        \label{fig:app:10_step_fb_adam_W_esd_n_8000_eta_1_to_1000}
\end{figure}

\begin{figure}[h!]
     \centering
     \begin{subfigure}[b]{0.23\textwidth}
         \centering
         \includegraphics[width=\textwidth]{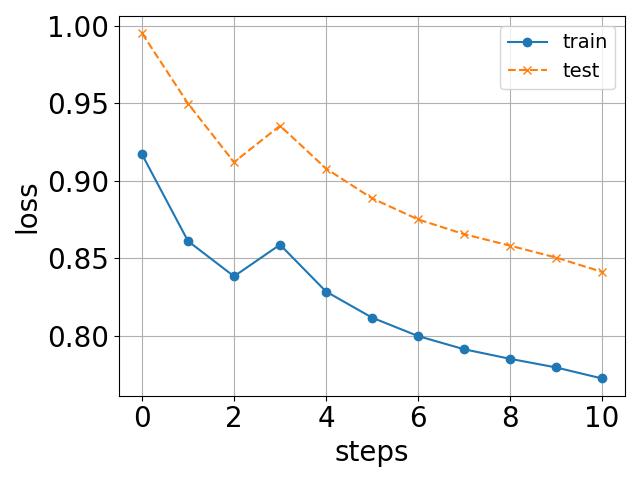}
         \caption{losses}
         \label{fig:app:10_step_fb_adam_lr1_n_8000_W_losses}
     \end{subfigure}
     \hfill
     \begin{subfigure}[b]{0.23\textwidth}
         \centering
         \includegraphics[width=\textwidth]{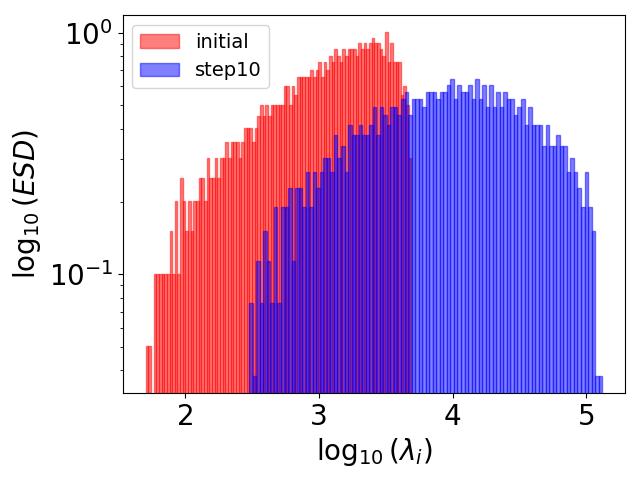}
         \caption{ESD of $\mW^\top\mW$}
         \label{fig:app:10_step_fb_adam_lr1_n_8000_W_esd}
     \end{subfigure}
     \hfill
     \begin{subfigure}[b]{0.25\textwidth}
         \centering
         \includegraphics[width=\textwidth]{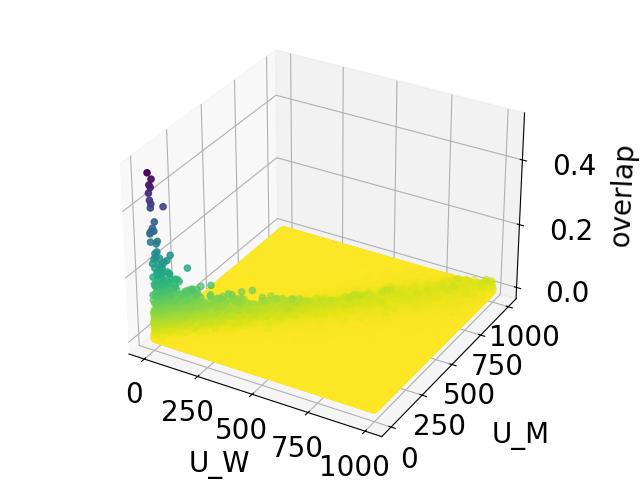}
         \caption{$\gO(\mU_{W_{10}}, \mU_{M_9})$ } 
         \label{fig:app:overlap_fb_adam_lr1_n_8000_u_o10}
     \end{subfigure}
     \begin{subfigure}[b]{0.25\textwidth}
         \centering
         \includegraphics[width=\textwidth]{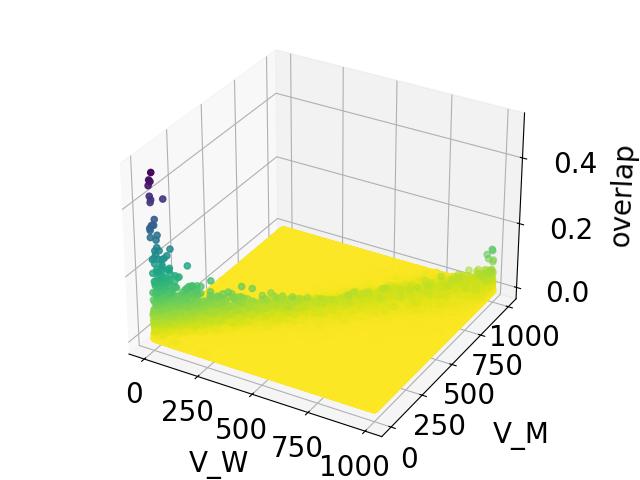}
         \caption{ $\gO(\mV_{W_{10}}, \mV_{M_9})$ } 
         \label{fig:app:overlap_fb_adam_lr1_n_8000_v_o10}
     \end{subfigure}
        \caption{ Losses, ESD, and Overlaps between singular vectors after $10$ \texttt{FB-Adam}($\eta=1$) steps for $n=8000, d=1000, h=1500, \lambda=0.01, \rho_e=0.3$. }
        \label{fig:app:overlap_fb_adam_lr1_n_8000_step10}
\end{figure}

\begin{figure}[h!]
     \centering
     \begin{subfigure}[b]{0.23\textwidth}
         \centering
         \includegraphics[width=\textwidth]{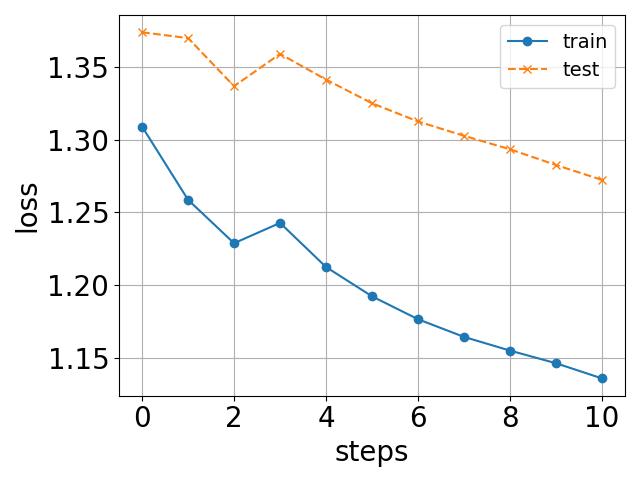}
         \caption{losses}
         \label{fig:app:10_step_fb_adam_lr1_n_8000_rho_0.7_W_losses}
     \end{subfigure}
     \hfill
     \begin{subfigure}[b]{0.23\textwidth}
         \centering
         \includegraphics[width=\textwidth]{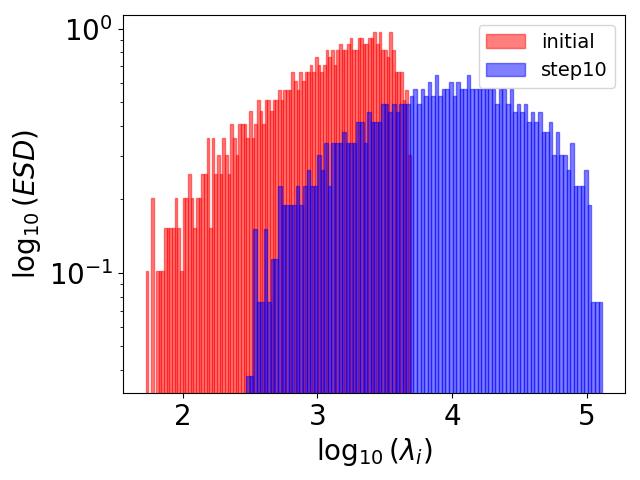}
         \caption{ESD of $\mW^\top\mW$}
         \label{fig:app:10_step_fb_adam_lr1_n_rho_0.7_8000_W_esd}
     \end{subfigure}
     \hfill
     \begin{subfigure}[b]{0.25\textwidth}
         \centering
         \includegraphics[width=\textwidth]{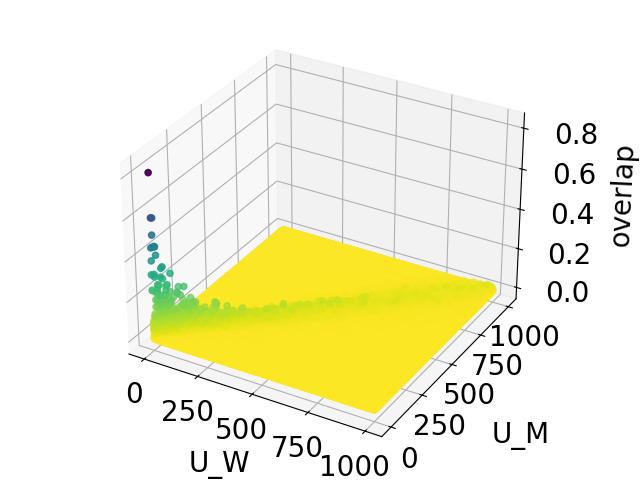}
         \caption{$\gO(\mU_{W_{10}}, \mU_{M_9})$ } 
         \label{fig:app:overlap_fb_adam_lr1_n_8000_rho_0.7_u_o10}
     \end{subfigure}
     \begin{subfigure}[b]{0.25\textwidth}
         \centering
         \includegraphics[width=\textwidth]{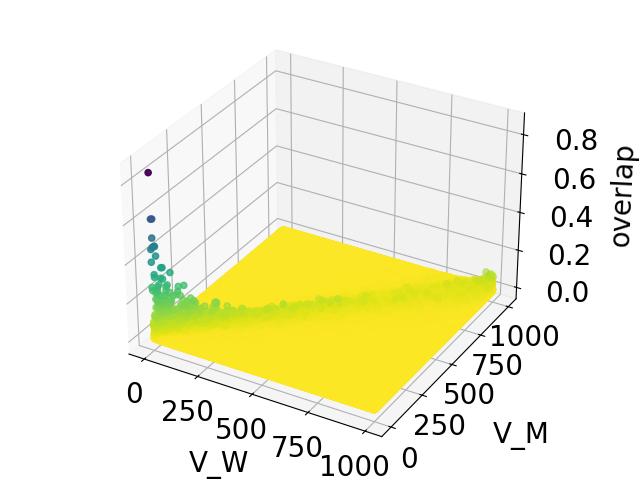}
         \caption{ $\gO(\mV_{W_{10}}, \mV_{M_9})$ } 
         \label{fig:app:overlap_fb_adam_lr1_n_8000_rho_0.7_v_o10}
     \end{subfigure}
        \caption{ Losses, ESD, and Overlaps between singular vectors after $10$ \texttt{FB-Adam}($\eta=1$) steps for $n=8000, d=1000, h=1500, \lambda=0.01, \rho_e=0.7$. }
        \label{fig:app:overlap_fb_adam_lr1_n_8000_rho_0.7_step10}
\end{figure}

In this section, we present additional experiments for $10$ optimizer updates. Figure \ref{fig:app:overlap_gd_lr2000_step10} presents the losses, ESD of $\mW^\top\mW$ and the overlaps of singular vectors after $10$ steps with \texttt{GD}($\eta=2000$). Notice that the prominent outlier values in Figures \ref{fig:app:overlap_gd_lr2000_n_2000_u_w1_m0}, \ref{fig:app:overlap_gd_lr2000_n_2000_v_w1_m0} after the one-step update have now reduced significantly. Thus illustrating the varying spread of values even for the left and right singular vector overlaps. The role of $\eta$ on the losses, \texttt{KTA} and the ESD metric \texttt{PL\_Alpha\_Hill} are illustrated in Figures \ref{fig:app:gd_fb_adam_bulk_loss_alignments_10_steps_n_2000}, \ref{fig:app:gd_fb_adam_bulk_loss_alignments_10_steps_n_4000}. We illustrate the ESD of $\mW^\top\mW$ after $10$ steps with $n=8000$ and learning rates $\eta$ chosen from $\{1, 10, 100, 1000\}$ in Figure \ref{fig:app:10_step_fb_adam_W_esd_n_8000_eta_1_to_1000}. Observe that as $\eta$ increases, the spike tends to move far away from the bulk and significantly distorts the shape of the bulk only for $\eta=1000$. Finally, in Figures \ref{fig:app:overlap_fb_adam_lr1_n_8000_step10}, \ref{fig:app:overlap_fb_adam_lr1_n_8000_rho_0.7_step10}
we illustrate the role of label noise increasing from $\rho_e=0.3$ (Figure \ref{fig:app:overlap_fb_adam_lr1_n_8000_step10}) to $\rho_e=0.7$ (Figure \ref{fig:app:overlap_fb_adam_lr1_n_8000_rho_0.7_step10}). Although the ESDs look the same in both cases, note that the outlier (max) values of the overlap matrices for $\rho_e=0.7$ have relatively larger values compared to the $\rho_e=0.3$ case.

\clearpage

\subsection{Spike movement with \texttt{GD}} 
\label{spike movement}

\begin{figure}[h!]
     \centering
     \begin{subfigure}[b]{0.24\textwidth}
         \centering
         \includegraphics[width=\textwidth]{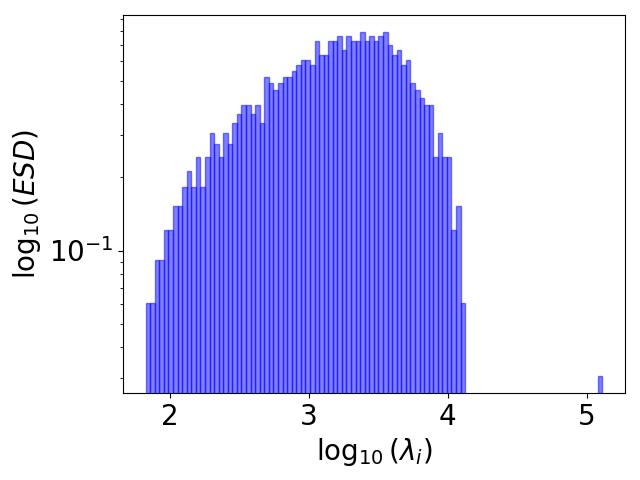}
         \caption{ ESD $\mW_5^\top\mW_5$ } 
         \label{fig:app:gd_lr2000_phase_transition_esd_w5}
     \end{subfigure}
     \hfill
     \begin{subfigure}[b]{0.24\textwidth}
         \centering
         \includegraphics[width=\textwidth]{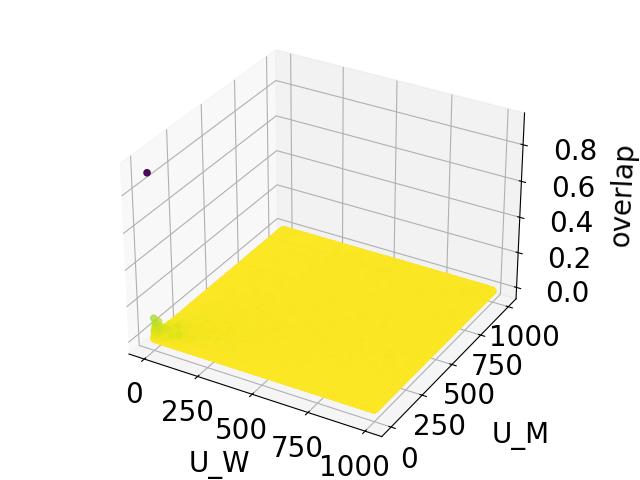}
         \caption{ $\gO(\mU_{W_{5}}, \mU_{M_5})$ } 
         \label{fig:app:gd_lr2000_phase_transition_o_w5}
     \end{subfigure}
     \hfill
     \begin{subfigure}[b]{0.24\textwidth}
         \centering
         \includegraphics[width=\textwidth]{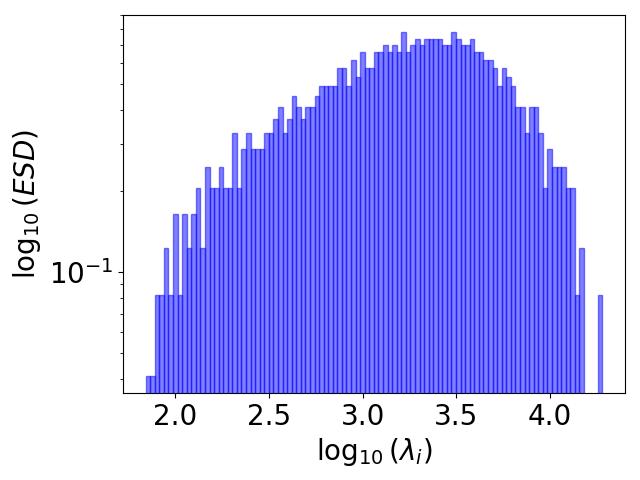}
         \caption{ ESD $\mW_6^\top\mW_6$ } 
         \label{fig:app:gd_lr2000_phase_transition_esd_w6}
     \end{subfigure}
     \hfill
     \begin{subfigure}[b]{0.24\textwidth}
         \centering
         \includegraphics[width=\textwidth]{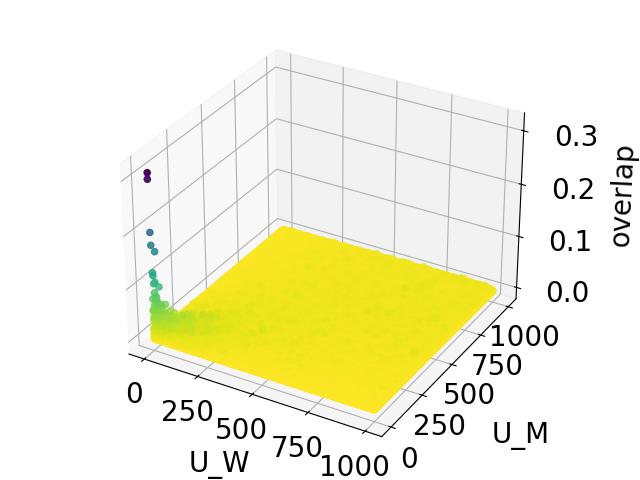}
         \caption{ $\gO(\mU_{W_{6}}, \mU_{M_5})$ } 
         \label{fig:app:gd_lr2000_phase_transition_o_w6}
     \end{subfigure}
        \caption{ Phase transition in ESD of $\mW^\top\mW$ between steps $5, 6$ when updated using \texttt{GD} ($\eta=2000$). Here $n=2000$, $d=1000$, $h=1500$, $\sigma_* = \texttt{softplus}, \sigma = \texttt{tanh}, \rho_e = 0.3$. }
        \label{fig:app:gd_lr2000_phase_transition}
\end{figure}

During our experiments with \texttt{GD}($\eta=2000$), we observed a surprising transition in the position of the spike relative to the bulk of the ESD. Particularly between steps $5$ and $6$, the spike in the ESD of $\mW^\top\mW$ which represents the largest singular value, reduces in value by an order of magnitude. Additionally, this reduction seems to be correlated with the reduction in maximum overlap values from $\texttt{max}(\gO(\mU_{W_{5}}, \mU_{M_5}))$ to $\texttt{max}(\gO(\mU_{W_{6}}, \mU_{M_5}))$ (see Figure \ref{fig:app:gd_lr2000_phase_transition}). To understand this behavior, we emphasize the spike in the ESD of $\mW_5^\top\mW_5 $ in Figure \ref{fig:app:gd_lr2000_phase_transition_esd_w5} and the large overlap value (black dot) in  Figure \ref{fig:app:gd_lr2000_phase_transition_o_w5}. Let $\hat{u}_{W_5}, \hat{u}_{M_5}, \hat{u}_{W_6} \in \sR^h$ represent the left singular vectors corresponding to the largest singular values in $\mW_5, \mM_5, \mW_6$ respectively. The large overlap value (black dot) in  Figure \ref{fig:app:gd_lr2000_phase_transition_o_w5} intuitively represents a high degree of overlap/alignment between $\hat{u}_{W_5}, \hat{u}_{M_5}$. As a result of obtaining $\mW_6$  by $\mW_6 = \mW_5 + \mM_5$, the singular vector $\hat{u}_{W_6}$ seems to be rotated from $\hat{u}_{W_5}$ in such a way that its alignment with $\hat{u}_{M_5}$ is reduced (see Figure \ref{fig:app:gd_lr2000_phase_transition_o_w6}). One can intuitively think of this process as the `diffusion'/`spread' of the overlap between $\hat{u}_{M_5}$ and all left singular vectors of $\mW_6$.

\subsection{On Generalization, Learning Rate Schedules and Weight Normalization}
\label{app:sec:wn_lr}

\begin{figure}[t!]
     \centering
     \begin{subfigure}[b]{0.24\textwidth}
         \centering
         \includegraphics[width=\textwidth]{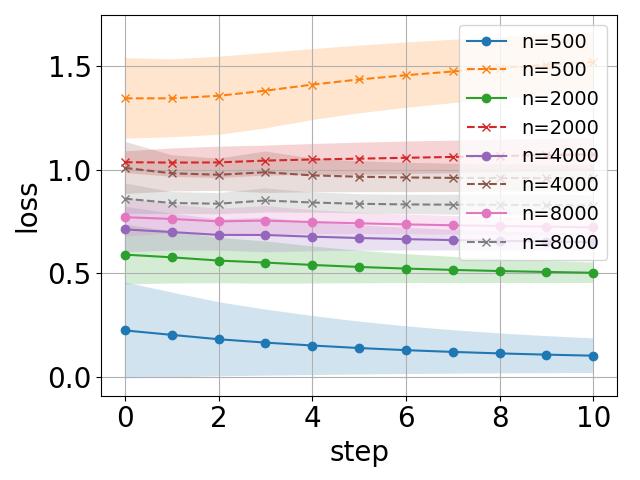}
         \caption{vary $n$}
         \label{fig:main:10_step_fb_adam_varying_loss_plots_vary_n}
     \end{subfigure}
     \hfill
     \begin{subfigure}[b]{0.24\textwidth}
         \centering
         \includegraphics[width=\textwidth]{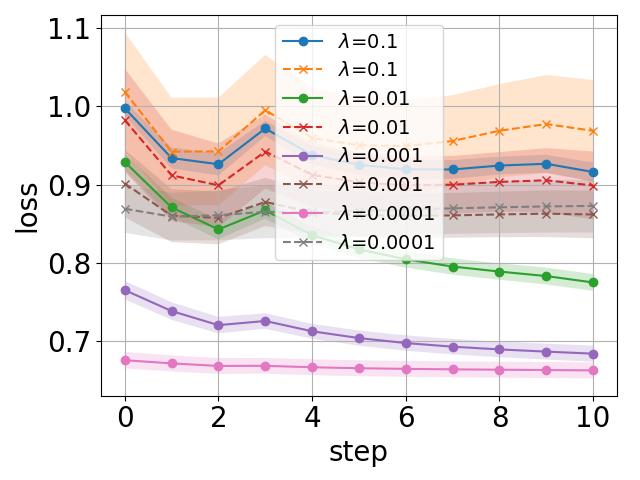}
         \caption{vary $\lambda$}
         \label{fig:main:10_step_fb_adam_varying_loss_plots_vary_lambda}
     \end{subfigure}
     \hfill
     \begin{subfigure}[b]{0.24\textwidth}
         \centering
         \includegraphics[width=\textwidth]{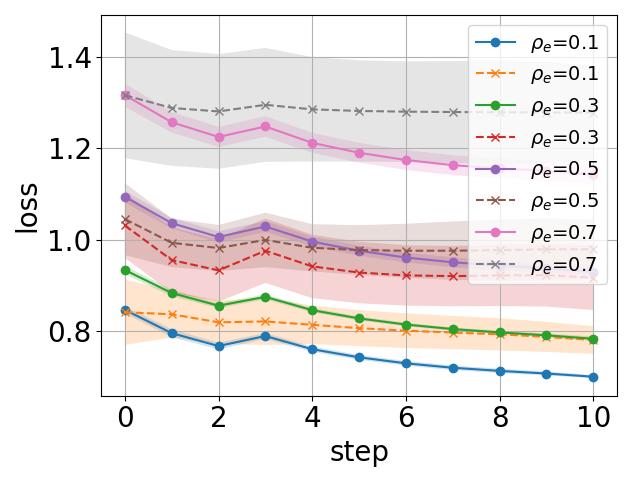}
         \caption{vary $\rho_e$}
         \label{fig:main:10_step_fb_adam_varying_loss_plots_vary_rho}
     \end{subfigure}
     \hfill
     \begin{subfigure}[b]{0.24\textwidth}
         \centering
         \includegraphics[width=\textwidth]{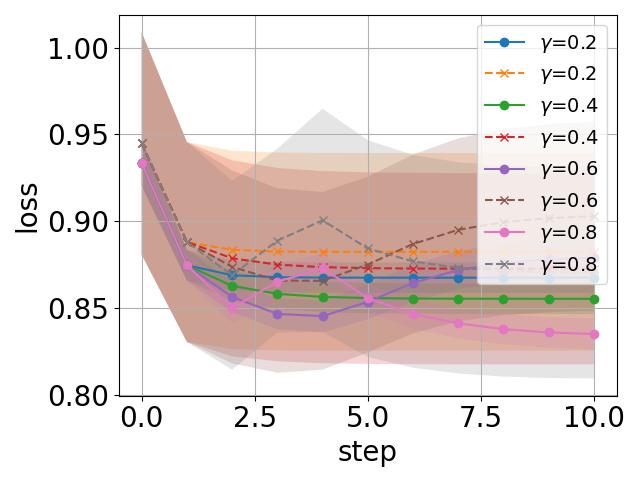}
         \caption{vary \texttt{StepLR} ($\gamma$)}
         \label{fig:main:10_step_fb_adam_varying_loss_plots_vary_gamma}
     \end{subfigure}
        \caption{\texttt{FB-Adam} train/test loss across $10$ steps with varying $n, \lambda, \rho_e$ and \texttt{StepLR}($\gamma$). (a) Varying dataset size $n$ (b) Varying regularization parameter $\lambda$ for obtaining the second-layer weights, (c) Varying the std.dev ($\rho_e$) of the additive Gaussian label noise, (d) Varying the $\gamma$ parameter which controls the decay of $\eta$. The bold lines indicate train loss and the dashed lines indicate test loss. }
        \label{fig:main:10_step_fb_adam_varying_loss_plots}
\end{figure}

\paragraph{Role of sample size ($n$).} By fixing $d=1000, h=1500$ and $\eta=1$, we vary the size of the training dataset $n$ as per the set $\{500, 2000, 4000, 8000\}$. In this setting, observe from Figure \ref{fig:main:10_step_fb_adam_varying_loss_plots_vary_n} that the train loss and test loss improve significantly for $n=8000$, while the network overfits for smaller $n$. Furthermore, for $n=2000$, we observed that $\mW_{10}^\top\mW_{10}$ exhibits a heavy-tailed ESD with \texttt{PL\_Alpha\_Hill}$=1.8$ (Figure \ref{fig:app:gd_fb_adam_bulk_loss_alignments_10_steps_n_2000_alpha}), which is less than the $n=8000$ case of $\approx 1.9$ (see Figure \ref{fig:gd_fb_adam_bulk_loss_alignments_10_steps_n_8000_alpha_hill}). The key difference in the latter case is that the spike in the ESD is consumed by the bulk, unlike the former where the outlier singular value is almost an order of magnitude away from the bulk.

\paragraph{Role of regularization constant ($\lambda$).} 

By considering a sample size of $n=8000$, and fixing $d=1000, h=1500, \eta=1$ as before, we can observe from Figure \ref{fig:main:10_step_fb_adam_varying_loss_plots_vary_lambda} that a lower training and test loss is achieved by $\lambda=10^{-3}$ and $\lambda=10^{-4}$, but with a large generalization gap (i.e. the difference between train and test loss). Additionally, observe that $\lambda=10^{-2}$ reasonably balances the test loss and generalization gap. Since the regression procedure does not modify the first layer weights, the choice of $\lambda$ can affect the interpretation of heavy tails leading to good/bad generalization.

\paragraph{Role of label noise ($\rho_e$).} Figure \ref{fig:main:10_step_fb_adam_varying_loss_plots_vary_rho} illustrates a consistent increase in losses for an increase in the additive Gaussian label noise $\rho_e$ from $\{0.1, 0.3, 0.5, 0.7\}$. However, the ESD of $\mW_{10}^\top\mW_{10}$ alone does not provide the complete picture to reflect this difficulty in learning. Instead, we observe noticeable differences in the distribution of values (especially outliers) along the diagonal of overlap matrices $\gO(\mU_{W_{10}}, \mU_{M_9})$, $\gO(\mV_{W_{10}}, \mV_{M_9})$ for $\rho_e=0.3$ (Figure \ref{fig:app:overlap_fb_adam_lr1_n_8000_u_o10}, \ref{fig:app:overlap_fb_adam_lr1_n_8000_v_o10}) and $\rho_e=0.7$ (Figure \ref{fig:app:overlap_fb_adam_lr1_n_8000_rho_0.7_u_o10}, \ref{fig:app:overlap_fb_adam_lr1_n_8000_rho_0.7_v_o10}). Especially, the outliers in the former case had smaller values (almost $0.5\times$) than the latter.

\paragraph{Role of learning rate scheduling.} As a natural extension of selecting a large learning rate at initialization, we analyze the role of employing learning rate schedules \citep{ge2019step} on the losses and ESD of $\mW^\top\mW$. We consider the simple \texttt{torch.optim.StepLR} scheduler with varying decay factors ($\gamma$) per step. A smaller $\gamma$ indicates a faster decay in the learning rate $\eta$ per step. We observe from Figure \ref{fig:main:10_step_fb_adam_varying_loss_plots_vary_gamma} that such fast decays (with $\gamma=0.2$ and $\gamma=0.4$) quickly turns a large learning rate to a smaller one and lead to stable loss curves. However, the trends are relatively unstable for $\gamma=0.6$ and $\gamma=0.8$ in the early steps. 

\begin{figure}[t!]
     \centering
     \begin{subfigure}[b]{0.24\textwidth}
         \centering
         \includegraphics[width=\textwidth]{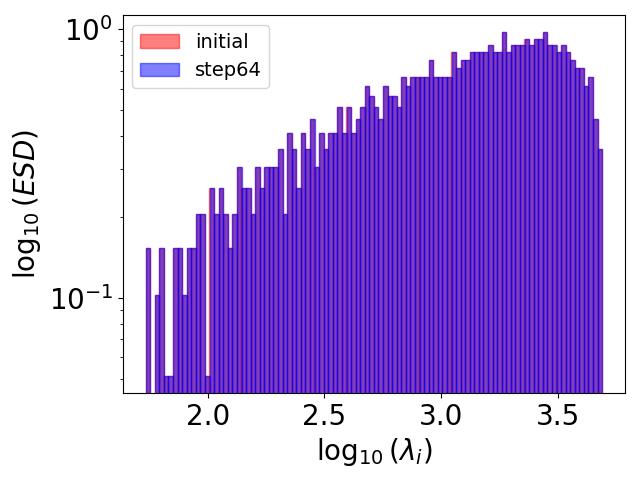}
         \caption{\texttt{GD} $\eta=0.1$}
     \end{subfigure}
     \hfill
     \begin{subfigure}[b]{0.24\textwidth}
         \centering
         \includegraphics[width=\textwidth]{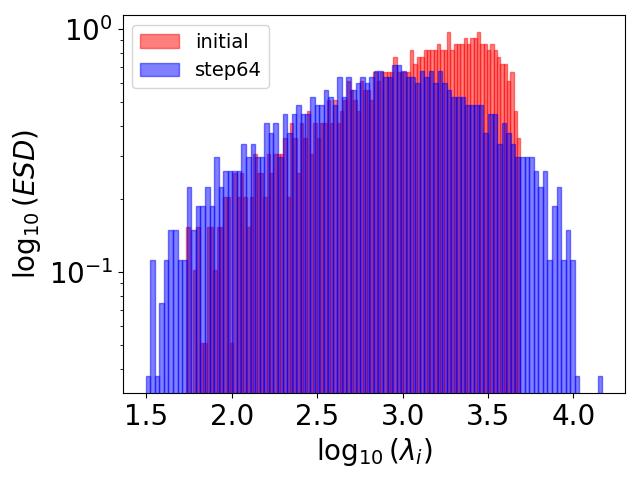}
         \caption{\texttt{GD} $\eta=2000$}
     \end{subfigure}
     \hfill
     \begin{subfigure}[b]{0.24\textwidth}
         \centering
         \includegraphics[width=\textwidth]{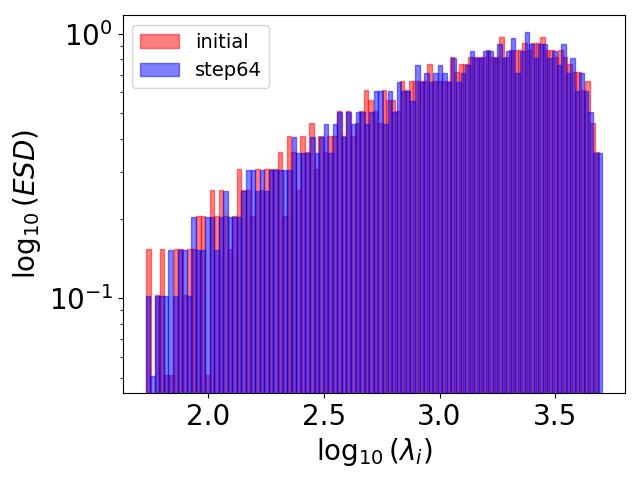}
         \caption{\texttt{FB-Adam} $\eta=0.1$}
     \end{subfigure}
     \hfill
     \begin{subfigure}[b]{0.24\textwidth}
         \centering
         \includegraphics[width=\textwidth]{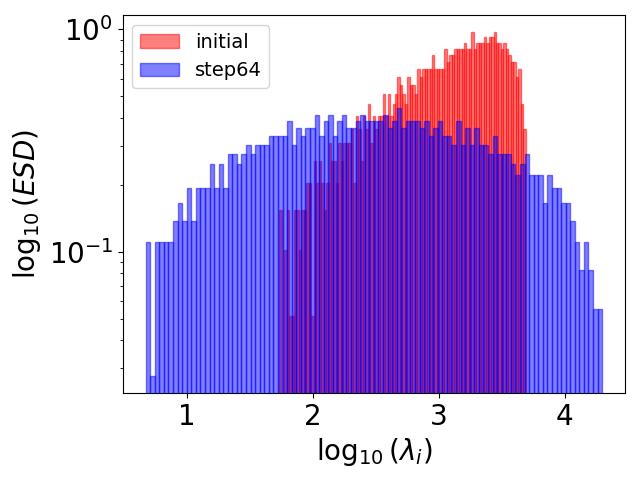}
         \caption{\texttt{FB-Adam} $\eta=1$}
     \end{subfigure}
        \caption{Evolution of ESD after 64 steps of \texttt{GD}, \texttt{FB-Adam} updates with varying $\eta$ and weight normalization. Here $n=2000$, $d=1000$, $h=1500$, $\sigma_* = \texttt{softplus}, \sigma = \texttt{tanh}, \rho_e = 0.3$. }
        \label{fig:app:64_step_gd_fb_adam_W_esd_wn}
        \vspace{-2mm}
\end{figure}

\begin{figure}[t!]
     \centering
     \begin{subfigure}[b]{0.24\textwidth}
         \centering
         \includegraphics[width=\textwidth]{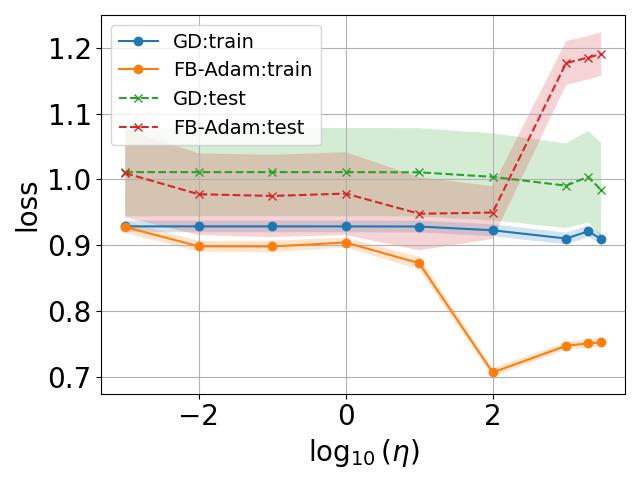}
         \caption{loss}
         \label{fig:gd_fb_adam_bulk_loss_alignments_10_steps_n_8000_wn_loss}
     \end{subfigure}
     \hfill
     \begin{subfigure}[b]{0.24\textwidth}
         \centering
         \includegraphics[width=\textwidth]{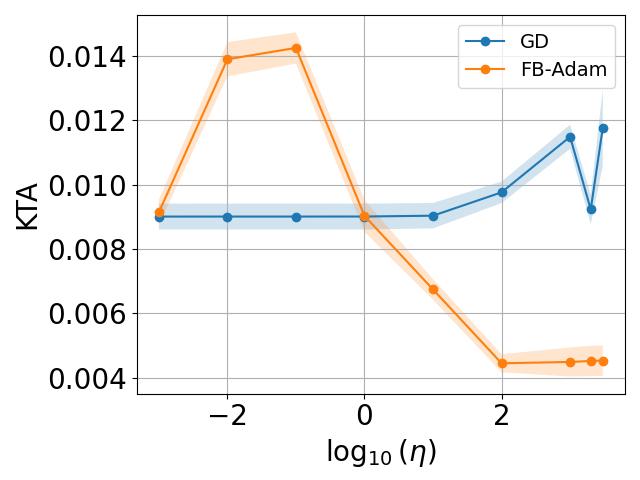}
         \caption{\texttt{KTA}}
         \label{fig:gd_fb_adam_bulk_loss_alignments_10_steps_n_8000_wn_kta}
     \end{subfigure}
     \hfill
     \begin{subfigure}[b]{0.24\textwidth}
         \centering
         \includegraphics[width=\textwidth]{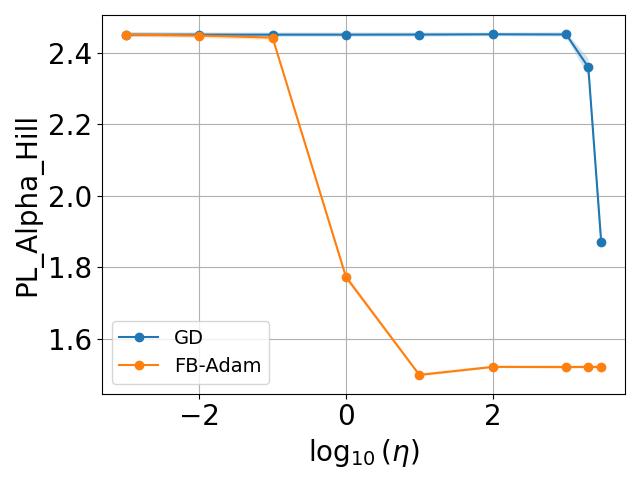}
         \caption{\texttt{PL\_Alpha\_Hill}}
         \label{fig:gd_fb_adam_bulk_loss_alignments_10_steps_n_8000_wn_alpha}
     \end{subfigure}
     \hfill
     \begin{subfigure}[b]{0.24\textwidth}
         \centering
         \includegraphics[width=\textwidth]{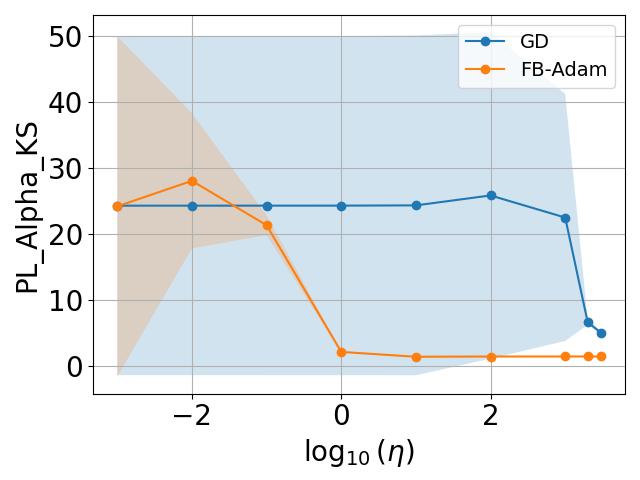}
         \caption{\texttt{PL\_Alpha\_KS}}
         \label{fig:gd_fb_adam_bulk_loss_alignments_10_steps_n_8000_wn_alpha_ks}
     \end{subfigure}
        \caption{ Train/test losses, \texttt{KTA}, \texttt{PL\_Alpha\_Hill}, \texttt{PL\_Alpha\_KS}  for $f(\cdot)$ trained with $10$ steps of  \texttt{GD}, \texttt{FB-Adam} with weight normalization. Here $n=8000$, $d=1000$, $h=1500$, $\sigma_* = \texttt{softplus}, \sigma = \texttt{tanh}, \rho_e = 0.3$.}
        \label{fig:gd_fb_adam_bulk_loss_alignments_10_steps_n_8000_wn}
\end{figure}

\begin{figure}[t!]
     \centering
     \begin{subfigure}[b]{0.24\textwidth}
         \centering
         \includegraphics[width=\textwidth]{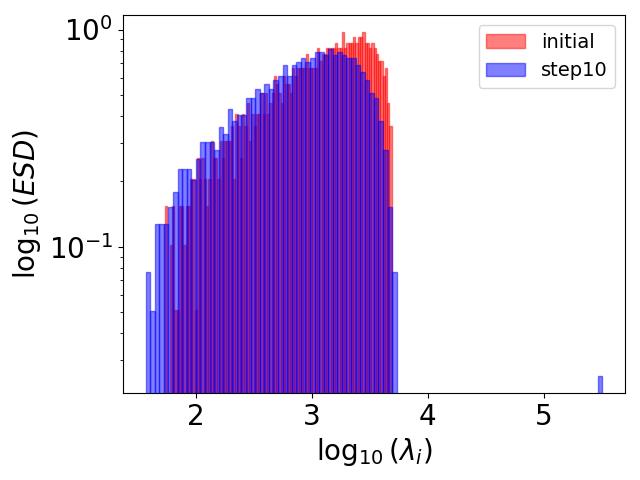}
         \caption{\texttt{StepLR($\gamma=0.2$)}}
     \end{subfigure}
     \hfill
     \begin{subfigure}[b]{0.24\textwidth}
         \centering
         \includegraphics[width=\textwidth]{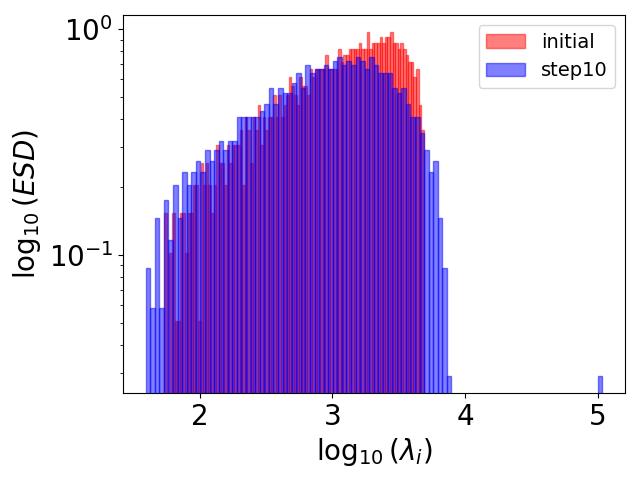}
         \caption{\texttt{StepLR($\gamma=0.4$)}}
     \end{subfigure}
     \hfill
     \begin{subfigure}[b]{0.24\textwidth}
         \centering
         \includegraphics[width=\textwidth]{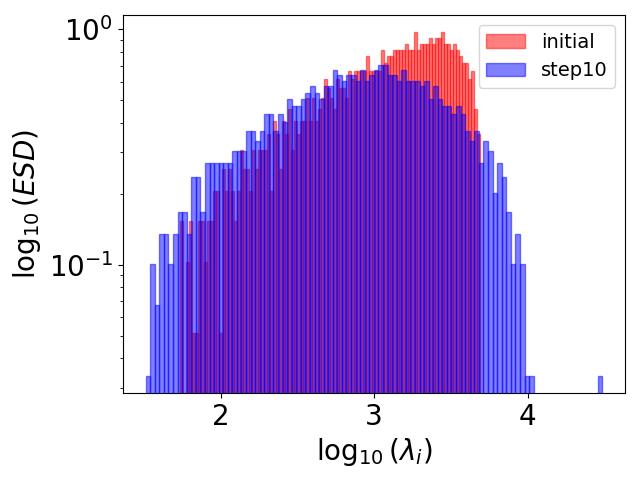}
         \caption{\texttt{StepLR($\gamma=0.6$)}}
     \end{subfigure}
     \hfill
     \begin{subfigure}[b]{0.24\textwidth}
         \centering
         \includegraphics[width=\textwidth]{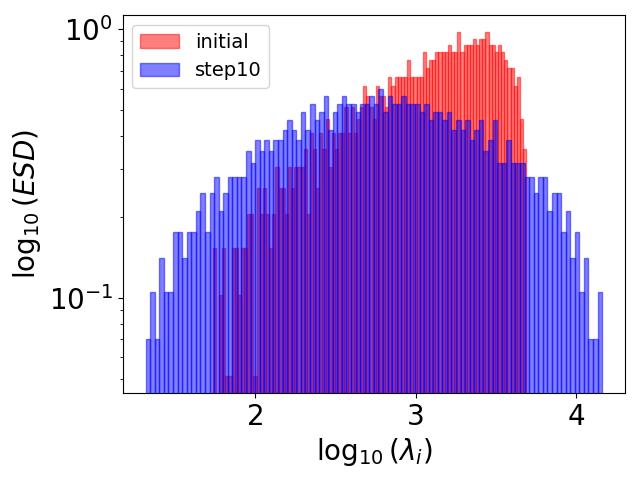}
         \caption{\texttt{StepLR($\gamma=0.8$)}}
     \end{subfigure}
        \caption{Evolution of ESD of $\mW^\top\mW$ after 10 steps of \texttt{FB-Adam}($\eta=1$) with weight normalization and varying decay rates for \texttt{StepLR} schedule. The decay factor ($\gamma$) is applied after every step. Here $n=2000$, $d=1000$, $h=1500$, $\sigma_* = \texttt{softplus}, \sigma = \texttt{tanh}, \rho_e = 0.3$. }
        \label{fig:app:10_step_gd_fb_adam_W_esd_steplr}
\end{figure}

\paragraph{Role of weight normalization (WN).} As mentioned in the main text, we employ a weight normalization technique \citep{huang2023normalization} after each optimizer update to $\mW$ as follows: $\mW_{t+1} = \frac{\sqrt{hd}\mW_{t+1}'}{\norm{\mW_{t+1}'}_F},\mW_{t+1}' = \mW_t + \mM_t$ to ensure that $\norm{\mW_{t+1}}_F$ is always $\sqrt{hd}$, before the forward pass. Observe from Figure~\ref{fig:app:64_step_gd_fb_adam_W_esd_wn} that after $64$ steps of \texttt{GD/FB-Adam} updates with varying $\eta$, the large $\eta$ cases lead to HT ESDs which spread over a wider range of singular values than the non-WN cases (see Table~\ref{tab:ADAM_transition}). Furthermore, Figure~\ref{fig:gd_fb_adam_bulk_loss_alignments_10_steps_n_8000_wn} conveys that in the case of \texttt{FB-Adam} for $0.01 \le \eta \le 100$, the \texttt{PL\_Alpha\_Hill} metric and the mean estimates of \texttt{PL\_Alpha\_KS} tend to exhibit HT ESDs and generalize well. Notice that this range is much broader than the non-WN case in the main text (Figure~\ref{fig:gd_fb_adam_bulk_loss_alignments_10_steps_n_8000}). Additionally, note that the \texttt{PL\_Alpha\_Hill} estimates for $10 \le \eta \le 100$ are relatively lower than the non-WN case. Finally, by employing learning rate schedules, we can control the presence of the spike while nudging the bulk of the ESD toward an HT distribution (Figure~\ref{fig:app:10_step_gd_fb_adam_W_esd_steplr}).

\subsection{Effect of using the  same activation function  for teacher and student }
\label{app:sec:same_act}


In our main paper, we adopt different activation functions for the teacher and student models: 
$\sigma_{*}=\texttt{softplus}$ for teacher and $\sigma=\texttt{tanh}$ for student,
Although this distinction introduces some learning challenges for the two-layer NN — making it harder to minimize both training and test loss — it does not affect the key theoretical results of this work, such as the scaling of $\eta$ for a single-step \texttt{FB-Adam} update, provided that both activation functions meet the fundamental assumptions. In this section, we use \texttt{tanh} activation for both models to examine whether a more consistent teacher-student setup yields any notable differences. We use the same setup as Section~\ref{Impact on Generalization}.




\paragraph{Correlations between ESD and losses.} Similar to the observations in the main text (see Figure~\ref{fig:gd_fb_adam_bulk_loss_alignments_10_steps_n_8000}) for different activation functions, note that for \texttt{GD} with $\eta \ge 1000$, the \texttt{KTA} increases and \texttt{PL\_Alpha\_Hill}, mean estimates of \texttt{PL\_Alpha\_KS} reduces. More importantly, note that the test loss values are relatively smaller (i.e $\approx 0.85$ in Figure~\ref{fig:gd_fb_adam_bulk_loss_alignments_10_steps_n_8000}, and $\approx 0.2$ in this setup). In the case of \texttt{FB-Adam}, we observe a surprising shift in the trend of loss values for $\eta \ge 0.1$. In essence, the range of $\eta$ for which HT ESDs emerge and lead to good generalization has now shrunk to a much smaller range. Especially, the $\texttt{PL\_Alpha\_Hill}$ value of $\approx 2$ which resulted in good generalization in Figure~\ref{fig:gd_fb_adam_bulk_loss_alignments_10_steps_n_8000} (relative to losses of the full range of $\eta$), correlates with poorer generalization in this setup. Overall, there seem to be non-trivial dependencies on the choice of activation functions to determine the correlations between HT ESDs and generalization.

\begin{figure}[h!]
     \centering
     \begin{subfigure}[b]{0.24\textwidth}
         \centering
         \includegraphics[width=\textwidth]{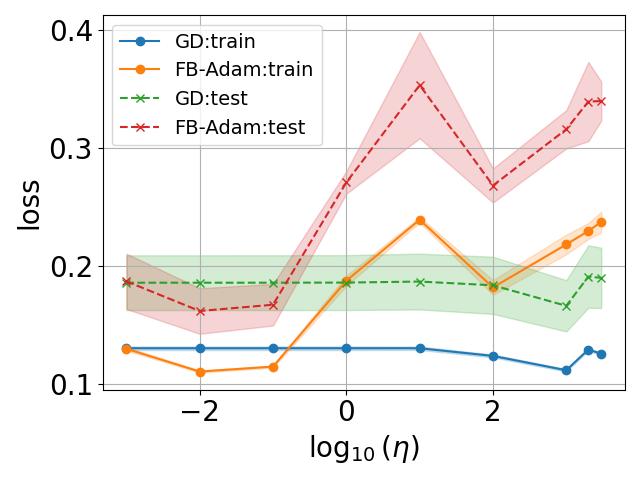}
         \caption{loss}
         \label{fig:gd_fb_adam_bulk_loss_alignments_10_steps_n_8000_loss_same_activation}
     \end{subfigure}
     \hfill
     \begin{subfigure}[b]{0.24\textwidth}
         \centering
         \includegraphics[width=\textwidth]{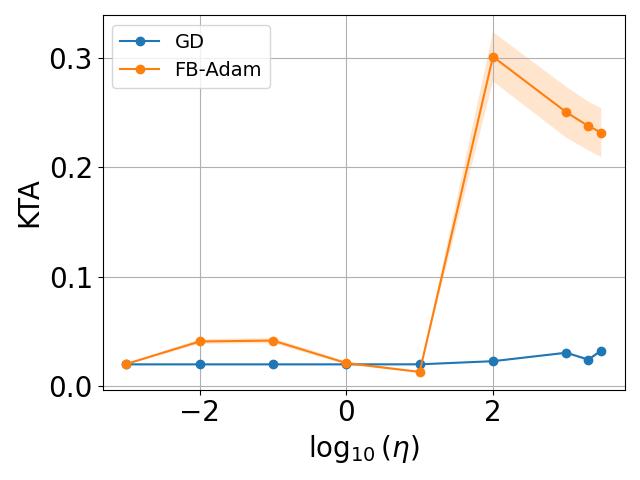}
         \caption{\texttt{KTA}}
         \label{fig:gd_fb_adam_bulk_loss_alignments_10_steps_n_8000_kta_same_activation}
     \end{subfigure}
     \hfill
     \begin{subfigure}[b]{0.24\textwidth}
         \centering
         \includegraphics[width=\textwidth]{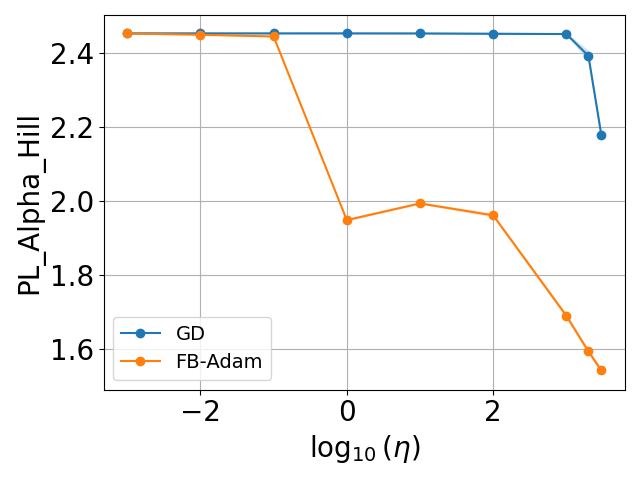}
         \caption{\texttt{PL\_Alpha\_Hill}}
         
         \label{fig:gd_fb_adam_bulk_loss_alignments_10_steps_n_8000_alpha_hill_same_activation}
     \end{subfigure}
     \begin{subfigure}[b]{0.24\textwidth}
         \centering
         \includegraphics[width=\textwidth]{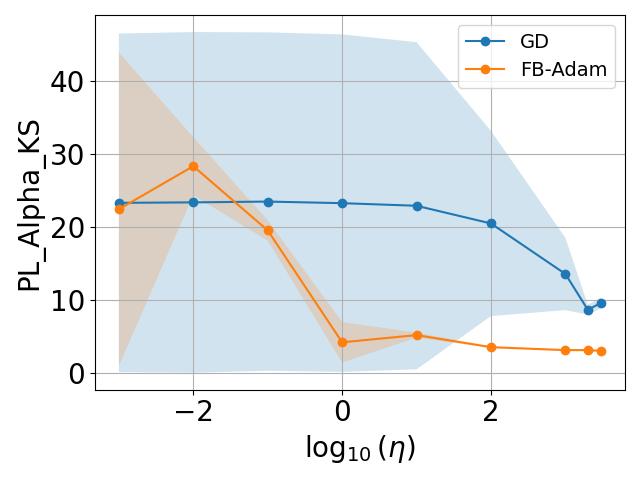}
         \caption{\texttt{PL\_Alpha\_KS}}
         
         \label{fig:gd_fb_adam_bulk_loss_alignments_10_steps_n_8000_alpha_ks_same_activation}
     \end{subfigure}
        \caption{Losses, \texttt{KTA}, \texttt{PL\_Alpha\_Hill}, \texttt{PL\_Alpha\_KS} for $f(\cdot)$ trained with $10$ steps of  \texttt{GD}, \texttt{FB-Adam}, with $n=8000$, $d=1000$, $h=1500$, $\sigma_* = \texttt{tanh}, \sigma = \texttt{tanh}, \rho_e = 0.3, \lambda=0.01$.}
        \label{fig:gd_fb_adam_bulk_loss_alignments_10_steps_n_2000_same_activation}
\vspace{-2mm}
\end{figure}

\subsection{Applicability of our results on deeper NNs}
\label{app:sec:vgg_mnist}

\begin{figure}[h!]
     \centering
     \begin{subfigure}[b]{0.32\textwidth}
         \centering
         \includegraphics[width=\textwidth]{final_images/10_step_VGG_ResNet/lr0.01_cifar10-ResNet18-Layer5.jpg}
         \caption{ ResNet18-Layer5 }  
     \end{subfigure}
     \hfill
     \begin{subfigure}[b]{0.32\textwidth}
         \centering
         \includegraphics[width=\textwidth]{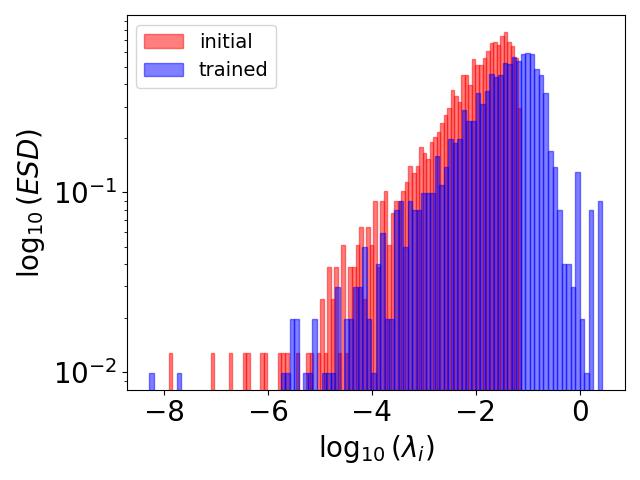}
         \caption{ ResNet18-Layer8 } 
     \end{subfigure}
     \hfill
     \begin{subfigure}[b]{0.32\textwidth}
         \centering
         \includegraphics[width=\textwidth]{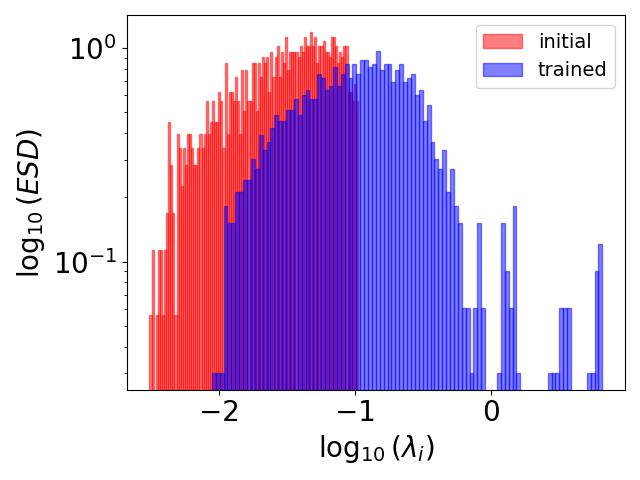}
         \caption{ResNet18-Layer10 } 
     \end{subfigure}
     \caption{layerwise ESDs of \texttt{ResNet18} after $10$ steps with \texttt{FB-Adam} and $\eta=0.01$ on \texttt{CIFAR10}.}
     \label{fig:app:ResNet:CIFAR10}
\end{figure}

\begin{figure}[h!]
     \centering
     \begin{subfigure}[b]{0.32\textwidth}
         \centering
         \includegraphics[width=\textwidth]{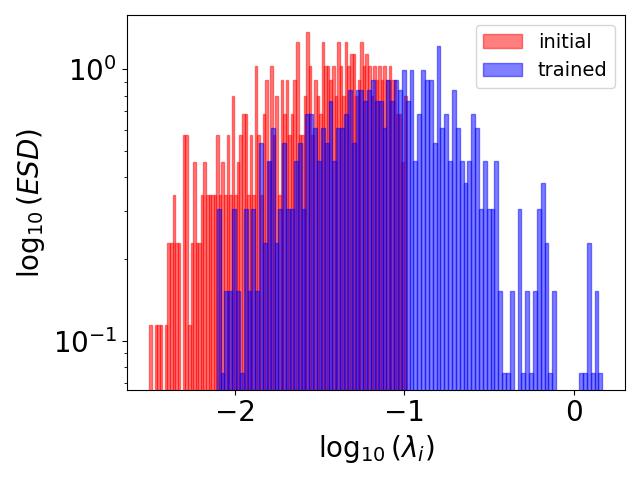}
         \caption{ ResNet18-Layer5 }  
     \end{subfigure}
     \hfill
     \begin{subfigure}[b]{0.32\textwidth}
         \centering
         \includegraphics[width=\textwidth]{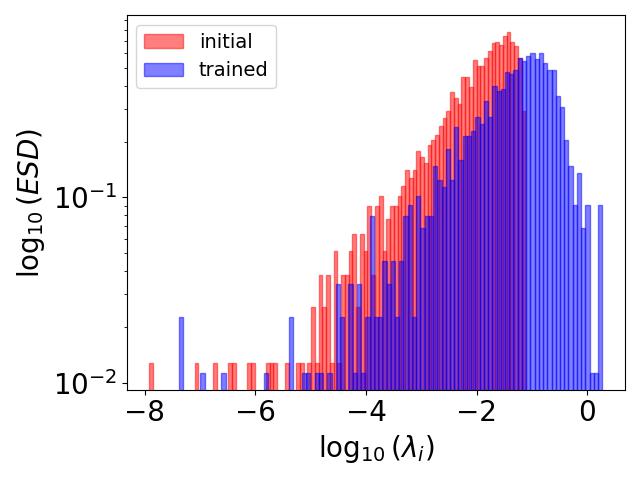}
         \caption{ ResNet18-Layer8 } 
     \end{subfigure}
     \hfill
     \begin{subfigure}[b]{0.32\textwidth}
         \centering
         \includegraphics[width=\textwidth]{final_images/10_step_VGG_ResNet/lr0.01_svhn-ResNet18-Layer10.jpg}
         \caption{ResNet18-Layer10 } 
     \end{subfigure}
     \caption{layerwise ESDs of \texttt{ResNet18} after $10$ steps with \texttt{FB-Adam} and $\eta=0.01$ on \texttt{SVHN}.}
     \label{fig:app:ResNet:SVHN}
\end{figure}

Beyond two-layer NNs trained on synthetic data, we train \texttt{VGG11, ResNet18} with \texttt{FB-Adam} on \texttt{MNIST, CIFAR10, SVHN} with a constant $\eta=0.01$ for 10 steps (epochs) to validate our claims. In Figure~\ref{fig:app:VGG:MNIST}, Figure~\ref{fig:app:VGG:CIFAR10}, Figure~\ref{fig:app:VGG:SVHN} below, we illustrate that the layers of \texttt{VGG11} can indeed exhibit HT ESDs even without stochastic gradient noise during the early phases of training. See Figure~\ref{fig:app:ResNet:CIFAR10} and Figure~\ref{fig:app:ResNet:SVHN} for \texttt{ResNet18} plots with \texttt{CIFAR10, SVHN} datasets.

\begin{figure}[h!]
     \centering
     \begin{subfigure}[b]{0.32\textwidth}
         \centering
         \includegraphics[width=\textwidth]{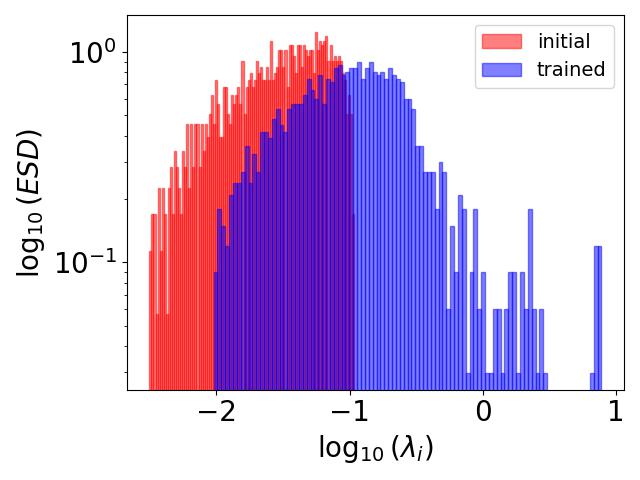}
         \caption{ VGG11-Layer2} 
     \end{subfigure}
     \hfill
     \begin{subfigure}[b]{0.32\textwidth}
         \centering
         \includegraphics[width=\textwidth]{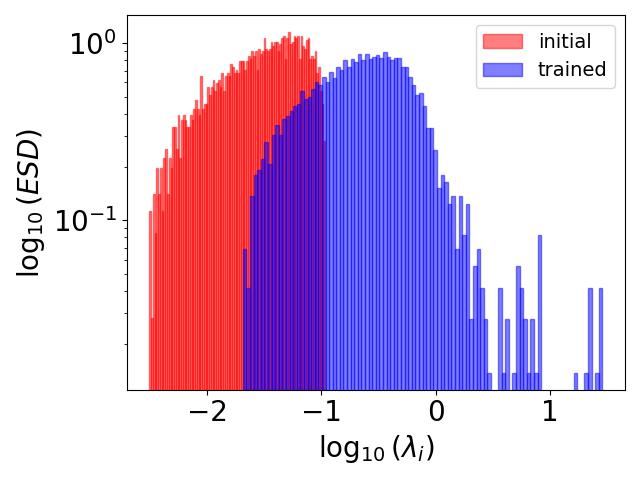}
         \caption{ VGG11-Layer4 } 
     \end{subfigure}
     \hfill
     \begin{subfigure}[b]{0.32\textwidth}
         \centering
         \includegraphics[width=\textwidth]{final_images/10_step_VGG_ResNet/lr0.01_mnist-VGG11-Layer7.jpg}
         \caption{VGG11-Layer7 } 
     \end{subfigure}
     \caption{layerwise ESDs of \texttt{VGG11} after $10$ steps with \texttt{FB-Adam} and $\eta=0.01$ on \texttt{MNIST}.}
     \label{fig:app:VGG:MNIST}

\end{figure}

\begin{figure}[h!]
     \centering
     \begin{subfigure}[b]{0.32\textwidth}
         \centering
         \includegraphics[width=\textwidth]{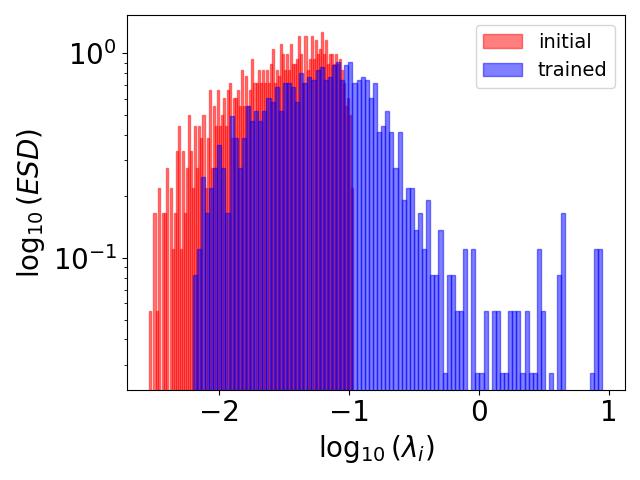}
         \caption{ VGG11-Layer2} 
     \end{subfigure}
     \hfill
     \begin{subfigure}[b]{0.32\textwidth}
         \centering
         \includegraphics[width=\textwidth]{final_images/10_step_VGG_ResNet/lr0.01_cifar10-VGG11-Layer4.jpg}
         \caption{ VGG11-Layer4 } 
     \end{subfigure}
     \hfill
     \begin{subfigure}[b]{0.32\textwidth}
         \centering
         \includegraphics[width=\textwidth]{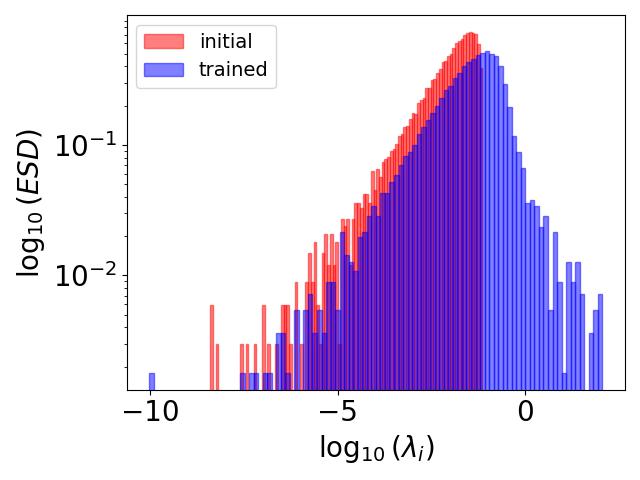}
         \caption{VGG11-Layer7 } 
     \end{subfigure}
     \caption{layerwise ESDs of \texttt{VGG11} after $10$ steps with \texttt{FB-Adam} and $\eta=0.01$ on \texttt{CIFAR10}.}
     \label{fig:app:VGG:CIFAR10}
\end{figure}

\begin{figure}[h!]
     \centering
     \begin{subfigure}[b]{0.32\textwidth}
         \centering
         \includegraphics[width=\textwidth]{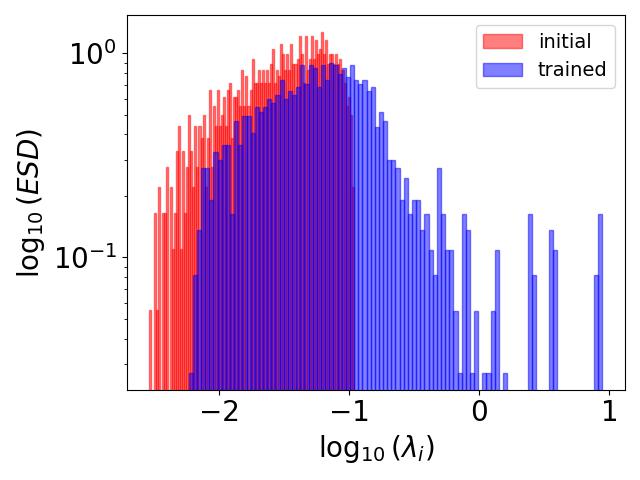}
         \caption{ VGG11-Layer2} 
     \end{subfigure}
     \hfill
     \begin{subfigure}[b]{0.32\textwidth}
         \centering
         \includegraphics[width=\textwidth]{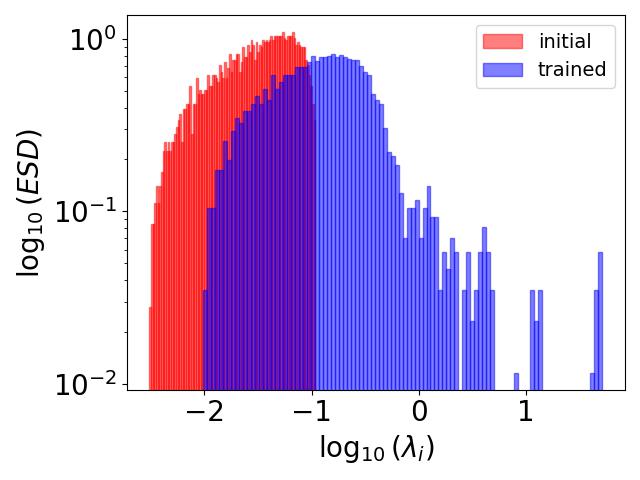}
         \caption{ VGG11-Layer4 } 
     \end{subfigure}
     \hfill
     \begin{subfigure}[b]{0.32\textwidth}
         \centering
         \includegraphics[width=\textwidth]{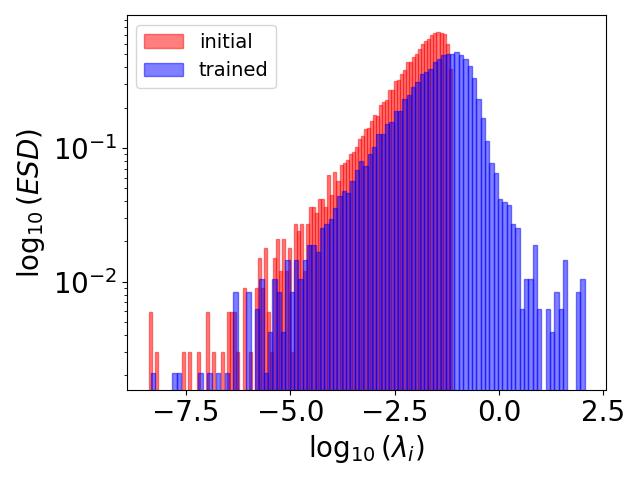}
         \caption{VGG11-Layer7 } 
     \end{subfigure}
     \caption{layerwise ESDs of \texttt{VGG11} after $10$ steps with \texttt{FB-Adam} and $\eta=0.01$ on \texttt{SVHN}.}
     \label{fig:app:VGG:SVHN}
\end{figure}

\section{A note on estimators of power-law exponents}
\label{app:sec:pl_estimator_note}

We found that the estimation of power-law (PL) exponents, such as \texttt{PL\_Alpha\_Hill} and \texttt{PL\_Alpha\_KS}, can be sensitive to the scale of singular values in the weight matrix.
However, Figure \ref{fig:gd_fb_adam_bulk_loss_alignments_10_steps_n_8000_alpha_hill} , \ref{fig:gd_fb_adam_bulk_loss_alignments_10_steps_n_8000_alpha_ks} illustrates a similar trend for both estimates. We believe this does not affect our qualitative interpretation of the relationship between HT, learning rates, and generalization.

Additionally, we highlight a particularly interesting observation. Considering \texttt{FB-Adam} based updates with $\eta = 1$ after $t=10$ steps and after $t=100$ steps in Table~\ref{tab:ADAM_transition}, notice that the spike tends to get closer to the bulk in the latter (i.e as training progresses). We have observed a similar behavior for GD (see Appendix~\ref{spike movement}) where the spike tends to merge with the bulk. Thus, the estimation of PL exponents should be relatively less affected by such outlier spikes as training progresses. 
Furthermore, in Figure~\ref{fig:gd_fb_adam_bulk_loss_alignments_10_steps_n_8000_alpha_ks}, the \texttt{PL\_Alpha\_KS} is affected by the relative position of the spike from the bulk, especially when the gap is negligible or there is no gap (see Table~\ref{tab:ADAM_transition} for an ESD visualization with $\eta=0.01$). Therefore, the trends with PL alpha estimates being the same apply to the $\eta \ge 0.1$ values, where the spike is sufficiently far from the bulk and the estimates tend to exhibit a non-increasing trend.

On a related note, some recent papers have adopted both fitting methods to explore the relationship between HT and generalization and leveraged it to improve model training. For example, \citet{martin2021implicit} used  \texttt{PL\_Alpha\_KS} to study the relationship between HT and generalization to propose the HT-SR theory, whereas \citet{zhou2023temperature} proposed a layer-wise $\eta$ selection technique based on layer-wise \texttt{PL\_Alpha\_Hill} estimates of the ESD’s during training.

\section{Limitations and Future work} 
\label{app:sec:fw}

Currently, our analysis on \texttt{FB-Adam} updates does not theoretically analyze the role of large learning rates on the training and test losses after one or more steps. such a rigorous characterization requires new techniques to analyze the regression loss beyond the Gaussian equivalence assumption \citep{ba2022high}. Furthermore, the techniques used to calculate the \texttt{PL\_Alpha} of ESDs are subject to bias (in the case of the hill estimator) and to relatively larger variance in the case of the KS variant. Similar issues have been discussed in previous works \citep{yang2023test, zhou2023temperature}, and analyzing multiple approaches can provide a complete picture of the heaviness of the tails.

To this end, we discuss the following potential future efforts.

\paragraph{Generalization with \texttt{FB-Adam}.} Recent papers employing a similar setup have analyzed the feature matrix $\overline{\mZ}$ to rigorously characterize the training and test errors after a one-step \texttt{GD} update \citep{moniri2023theory,cui2024asymptotics}. In a similar spirit, the results from our work can be utilized to theoretically explore the spectral properties of $\overline{\mZ}$ with \texttt{FB-Adam}. Additionally, since the spectral properties of $\overline{\mZ}$ are tightly linked to the ESD of $\mW_t^\top\mW_t$, the fundamental question on the necessity of HT ESDs for generalization can be studied.

\paragraph{Spectral gap and step complexity.} In Section~\ref{sec:HT_phenomenon} we observed that the number of steps required for HT ESD emergence depends on the spectral gap (i.e the distance between the spike and the bulk in our setup) after the first step. While our work establishes the necessity of the spike, further analysis of the relationship between this spectral gap and the step complexity for HT ESD emergence can lead to novel insights. 

\paragraph{HT phenomenon and singular vector overlaps.} By presenting qualitative results that indicate HT-like distributions along the diagonals of overlap matrices, we aim to bring singular vectors into the picture for future HT phenomenon studies. Particularly, the qualitative differences in the overlap matrices of left and right singular vectors remain to be explored. Furthermore, metrics to quantify the `spread' of the on/off-diagonal overlap values can present a holistic picture of the interactions between the singular spaces of the weights and optimizer updates.

\paragraph{Towards analysis with deeper NNs and the `$5+1$' phase model.} Our work showcased how the `$5+1$' phase model of the ESDs can be studied under simpler two-layer NN settings and reasonably explain the practical NN ESD dynamics. A valuable direction of research is to extend this analysis to deeper NNs \citep{nichani2023provable} with multi-index models as teacher networks.

\paragraph{Designing novel techniques to improve generalization.} Recently, \citet{zhou2023temperature} employed the layer-wise \texttt{PL\_Alpha\_Hill} metric to design a layer-wise learning rate scheduler based on \texttt{SGD}. Based on our observations on the varying effects of $\eta$ for $\texttt{GD/FB-Adam}$ and the dynamics of ESD evolution, there is immense potential to further improve such schedulers. Furthermore, new regularization techniques to balance the \texttt{PL\_Alpha\_Hill} metrics across layers during training can lead to NNs that satisfy the ESD shape metric criteria to be considered as `well-trained' by model selection approaches \citep{martin2021predicting, yang2023test}. The intriguing consequences of such a technique on the convergence rates and sample complexities can lead to new directions of research.

\end{document}

%% file: math_commands.tex
\usepackage{amsmath,amsfonts,bm}

\def\eqref#1{equation~\ref{#1}}

\def\1{\bm{1}}

\newcommand{\norm}[1]{\left\lVert#1\right\rVert}
\newcommand\gb{ \rowcolor{gray!15}}

\def\vzero{{\bm{0}}}
\def\vone{{\bm{1}}}

\def\vbeta{{\boldsymbol{\beta}}}

\def\va{{\bm{a}}}

\def\vu{{\bm{u}}}
\def\vv{{\bm{v}}}

\def\vx{{\bm{x}}}
\def\vy{{\bm{y}}}

\def\mA{{\bm{A}}}
\def\mB{{\bm{B}}}

\def\mG{{\bm{G}}}

\def\mI{{\bm{I}}}
\def\mJ{{\bm{J}}}
\def\mK{{\bm{K}}}

\def\mM{{\bm{M}}}

\def\mP{{\bm{P}}}
\def\mQ{{\bm{Q}}}

\def\mS{{\bm{S}}}

\def\mU{{\bm{U}}}
\def\mV{{\bm{V}}}
\def\mW{{\bm{W}}}
\def\mX{{\bm{X}}}

\def\mZ{{\bm{Z}}}

\DeclareMathAlphabet{\mathsfit}{\encodingdefault}{\sfdefault}{m}{sl}
\SetMathAlphabet{\mathsfit}{bold}{\encodingdefault}{\sfdefault}{bx}{n}

\def\gN{{\mathcal{N}}}
\def\gO{{\mathcal{O}}}

\def\sN{{\mathbb{N}}}

\def\sP{{\mathbb{P}}}

\def\sR{{\mathbb{R}}}
\def\sS{{\mathbb{S}}}

\newtheorem{theorem}{Theorem}[section]

\newtheorem{proposition}[theorem]{Proposition}
\newtheorem{lemma}[theorem]{Lemma}
\newtheorem{corollary}[theorem]{Corollary}
\newtheorem{definition}[theorem]{Definition}
\newtheorem{assumption}[theorem]{Assumption}